\documentclass[DIV15, a4paper]{scrartcl}

\usepackage{authblk}
\usepackage{prettyref}
\usepackage{graphicx}
\usepackage{subfig}
\usepackage{booktabs}
\usepackage{colortbl}
\usepackage{amssymb}
\usepackage{amsfonts}
\usepackage{amsthm}
\usepackage{multirow}
\usepackage{tabularx}
\usepackage{footnote}
\usepackage{hyperref}
\usepackage{amsopn}
\usepackage{hhline}
\usepackage{cite}
\usepackage{setspace}
\usepackage{algorithmic}
\usepackage{algorithm}
\usepackage{amsmath}

\usepackage{makecell}
\usepackage{bm}
\usepackage{amstext} 
\usepackage{array}   
\newcolumntype{L}{>{$}l<{$}} 

\usepackage{xcolor}  
\definecolor{myblue}{RGB}{34,31,217}

\definecolor{mycyan}{gray}{.7}

\newtheorem{definition}{Definition}

\newrefformat{fig}{Figure~\ref{#1}}
\newrefformat{tab}{Table~\ref{#1}}
\newrefformat{sec}{Section~\ref{#1}}
\newrefformat{app}{Appendix~\ref{#1}}
\newrefformat{alg}{Algorithm~\ref{#1}}
\newrefformat{property}{Property~\ref{#1}}
\newrefformat{theorem}{Theorem~\ref{#1}}
\newrefformat{corollary}{Corollary~\ref{#1}}
\newrefformat{proposition}{Proposition~\ref{#1}}
\newrefformat{def}{Definition~\ref{#1}}
\newrefformat{lemma}{Lemma~\ref{#1}}
\newrefformat{eq}{equation~(\ref{#1})}

\usepackage{graphicx}
\definecolor{Gray}{gray}{0.9}
\usepackage{scrlayer-scrpage}

\usepackage{lscape}

\begin{document}

\title{\textbf\LARGE\fontfamily{cmss}\selectfont An Effective and Efficient Evolutionary Algorithm for Many-Objective Optimization}

\author[1]{\normalsize\fontfamily{lmss}\selectfont Yani Xue}
\affil[1]{\normalsize\fontfamily{lmss}\selectfont Department of Computer Science, Brunel University London, Uxbridge, Middlesex UB8 3PH, U.~K.}

\author[2]{\normalsize\fontfamily{lmss}\selectfont Miqing Li}
\affil[2]{\normalsize\fontfamily{lmss}\selectfont CERCIA, School of Computer Science, University of Birmingham, Birmingham B15 2TT, U.~K.}

\author[1]{\normalsize\fontfamily{lmss}\selectfont Xiaohui Liu}

\date{}

\maketitle

{\normalsize\fontfamily{lmss}\selectfont\textbf{Abstract:} In evolutionary multiobjective optimization, effectiveness refers to how an evolutionary algorithm performs in terms of converging its solutions into the Pareto front and also diversifying them over the front. This is not an easy job, particularly for optimization problems with more than three objectives, dubbed many-objective optimization problems. In such problems, classic Pareto-based algorithms fail to provide sufficient selection pressure towards the Pareto front, whilst recently developed algorithms, such as decomposition-based ones, may struggle to maintain a set of well-distributed solutions on certain problems (e.g., those with irregular Pareto fronts). Another issue in some many-objective optimizers is rapidly increasing computational requirement with the number of objectives, such as hypervolume-based algorithms and shift-based density estimation (SDE) methods. In this paper, we aim to address this problem and develop an effective and efficient evolutionary algorithm (E3A) that can handle various many-objective problems. In E3A, inspired by SDE, a novel population maintenance method is proposed to select high-quality solutions in the environmental selection procedure. We conduct extensive experiments and show that E3A performs better than $11$ state-of-the-art many-objective evolutionary algorithms in quickly finding a set of well-converged and well-diversified solutions.}
	
{\normalsize\fontfamily{lmss}\selectfont\textbf{Keywords:} evolutionary algorithms, many-objective optimization, effectiveness, efficiency}

{\normalsize\fontfamily{lmss}\selectfont\textbf{To appear in}: \textit{Information Sciences}}

\section{Introduction}
Many-objective optimization problems (MaOPs) refer to the optimization scenarios having four or more objectives to be considered simultaneously~\cite{Xiang2020many}. MaOPs abound in real-world applications in many fields, such as software engineering, manufacturing, and logistics. In the last decade, there is an increasing interest in the use of evolutionary algorithms for MaOPs, resulting in a variety of many-objective evolutionary algorithms (MaOEAs).

In the context of evolutionary multi-/many-objective optimization, effectiveness refers to how a search algorithm performs in terms of converging its solutions into the Pareto front and also diversifying them over the front. Efficiency refers to how quickly a search algorithm is executed. Achieving both high effectiveness and efficiency is not easy, particularly on practical applications where the problem's Pareto front may be highly complex and unpredictable.

Convergence is especially challenging in many-objective optimization. Pareto-based algorithms in evolutionary multiobjective optimization (EMO), e.g., the non-dominated sorting genetic algorithm II (NSGA-II)~\cite{Deb2002nsga_ii} and the strength Pareto evolutionary algorithm 2 (SPEA2)~\cite{Zitzler2001spea2}, often fail to scale up well in objective dimensionality. Moreover, recent studies~\cite{Li2017multiline} suggest that well-established decomposition-based~\cite{Zhang2007moead, Deb2014nsga_iii}, and indicator-based~\cite{Zitzler2004indicator, Beume2007sms} algorithms may also struggle to converge their solutions even when the objective dimension is as low as four.

Maintaining a well-distributed solution set is another important issue in many-objective optimization. Improving convergence may come at a cost of compromising the diversity. For example, new dominance relations (such as the $\epsilon$-\textit{dominance}~\cite{Laumanns2002combin} and the \textit{fuzzy Pareto dominance}~\cite{He2014fuzzy}), which are designed for promoting the convergence, may lead the population into one (or several) sub-area(s) of the Pareto front~\cite{Corne2007techniques,Lopez2009study}. Decomposition-based algorithms, which perform very well in terms of convergence, typically face challenges of diversifying their solutions over the Pareto front for problems with irregular Pareto front shapes~\cite{Ishibuchi2016PF}. In addition, indicator-based algorithms tend to favor a certain region of the Pareto front, such as the indicator-based evolutionary elgorithm (IBEA)~\cite{Zitzler2004indicator} for the boundary solutions of the Pareto front~\cite{Li2014diversity} and the S metric Selection based evolutionary multiobjective optimization algorithm (SMS-EMOA)~\cite{Beume2007sms} for the knee point(s)~\cite{Li2015performance}. In contrast, some recent many-objective algorithms may miss a certain region of the Pareto front. For example, the SPEA2 based on shift-based density estimation (SPEA2+SDE)~\cite{Li2014sde} has difficulty in maintaining boundary solutions in some problems~\cite{Li2016stochastic}.

On the other hand, the efficiency of some EMO algorithms decreases dramatically with the increase of the number of objectives. For example, the time requirements of algorithms based on the hypervolume indicator~\cite{Zitzler1999}, such as SMS-EMOA~\cite{Beume2007sms}, increases exponentially with the increase of the objective dimensionality. For another example, SPEA2+SDE, which has been found to perform well on many-objective optimization problems~\cite{Li2017multiline,Li2018evolutionary}, also suffers from poor efficiency as the computational complexity is $O(mn^3)$, where $n$ denotes the size of the population.

Lastly, many multi-/many-objective evolutionary algorithms need some extra parameters which should be set properly and individually for different problems. For example, for algorithms based on the modified Pareto dominance relation (e.g., the $\epsilon$-domination based multiobjective evolutionary algorithm, or simply $\epsilon$-MOEA~\cite{Deb2005eval}), it is crucial to specify the relax degree of the Pareto dominance (i.e., the area that a solution dominates) for a problem with a large number of objectives~\cite{Laumanns2011st, Li2013comparative}. For another example, in some region-based MaOEAs, it is important to find right parameters to determine the size of the region, such as the grid division parameter in the grid-based evolutionary algorithm (GrEA)~\cite{Yang2013grid} and the neighborhood size in the knee point driven evolutionary algorithm (KnEA)~\cite{Zhang2014knee}. However, the \textit{sweet-spot} of such parameters may significantly shrink with the number of objectives~\cite{Purshouse2007on}. It is difficult or even impossible to find their best settings in the high-dimensional objective space~\cite{Li2013comparative}.

Given the above, this paper aims to develop a many-objective evolutionary algorithm (called E3A) that is able to achieve both high effectiveness and efficiency. The main contributions of this paper can be summarized as follows:
\begin{enumerate}
  \item[1)] The proposed algorithm is of the following desirable features: a) effectiveness in the sense of converging its solutions into the Pareto front and also diversifying them on the front; b) efficiency in the sense of a reasonable amount of execution time; c) suitability for MaOPs with various Pareto front shapes; and d) no additional parameter except those associated with an evolutionary algorithm (e.g., population size and crossover rate.
  \item[2)] In E3A, a novel population maintenance method is proposed to preserve promising solutions during the evolutionary process. The proposed population maintenance method consists of two key operations -- selecting boundary solutions and selecting non-boundary solutions. The former is to determine the range of the estimated Pareto front, from which the latter aims to maintain a set of well-distributed and well-converged solutions in the high-dimensional space. 
  \item[3)] Extensive experiments are conducted to evaluate E3A. We consider $60$ well-established problem instances with up to $15$ objectives and $11$ state-of-the-art MaOEAs. The experimental results demonstrate that E3A generally outperforms its peers in achieving high effectiveness and efficiency.
\end{enumerate}

The remainder of this paper is organized as follows. Section \ref{sec:E3A_Background} gives basic definitions in MaOPs and reviews the related work on MaOEAs. Section \ref{sec:E3A_Proposed_Algorithm} describes the framework and details the proposed E3A algorithm. The experimental results are presented in Section \ref{sec:E3A_Experimental_Results}. Finally, the conclusion and future work are given in Section \ref{sec:E3A_Conclusion}.

\section{Background}\label{sec:E3A_Background}
\subsection{Concepts and Terminology}\label{sec:E3A_concepts}
Mathematically, a minimization MaOP with $m$ objectives and $d$ decision variables can be defined as follows:
\begin{equation}
	\label{equ:MaOP_definition}
	\begin{split}
		& \textrm{minimize} \quad\quad\;\; F(\bm{x}) = (f_1(\bm{x}), f_2(\bm{x}), \dots, f_m(\bm{x})) \\
		& \textrm{subject to} \quad\quad\; \bm{x} = (x_1, x_2, \dots, x_d), \quad \bm{x} \in \Omega
	\end{split}
\end{equation}
where $\bm{x}$ represents a solution ($d$-dimensional decision variable vector) in the decision space $\Omega$, $F: \Omega \rightarrow \Theta $ defines an objective vector consisting of $m \geq 4$ objective functions, which maps the decision space $\Omega$ to the objective space $\Theta$.

In many-objective optimization, since these objectives often conflict with each other, there is no single optimal solution for an MaOP, but rather a set of Pareto-optimal solutions, which is defined on the basis of the Pareto dominance relation.

\begin{definition}[Pareto Dominance]
	\label{ParetoDominance}
	For a given MaOP, a decision vector $\bm{x} = (x_1$, $x_2$, $\dots$, $x_d)$ is said to (Pareto) \textit{dominate} another decision vector $ \bm{y}= (y_1$, $y_2$, $\dots$, $y_d) $ (denoted as $\bm{x} \prec \bm{y}$), or equivalently $\bm{y}$ is dominated by $\bm{x}$, if and only if
	\begin{equation}
		\label{equ:PD}
		\begin{split}
			\forall \, i \in (1, 2, \dots, m):  f_i (\bm{x}) \leq f_i (\bm{y}) \\
			\wedge \, \exists \, i \in (1, 2, \dots, m):  f_i (\bm{x}) < f_i (\bm{y}).
		\end{split}
	\end{equation}
\end{definition}

\begin{definition}[Pareto Optimality]
	\label{ParetoOptimity}
	For a solution to a given MaOP in (\ref{equ:MaOP_definition}), $\bm{x}^{\ast} \in \Omega $, if there is no solution $\bm{z} \in \Omega $ that dominates solution $\bm{x}^{\ast}$, then $\bm{x}^{\ast}$ is said to be Pareto optimal. All such solutions are called Pareto-optimal (or non-dominated) solutions.
\end{definition}

\begin{definition}[Pareto Set]
	\label{PS}
	The Pareto set of a given MaOP is defined as all Pareto-optimal (or non-dominated) solutions in the decision space.
\end{definition}

\begin{definition}[Pareto Front]
	\label{Pareto front}
	The Pareto front of a given MaOP is defined as corresponding objective vectors to the Pareto set.
\end{definition}

\subsection{Related Work} \label{sec:E3A_related_work}
Over the last decade, there is a growing interest in developing many-objective evolutionary algorithms (MaOEAs) for various MaOPs. Roughly, these MaOEAs can be classified into five categories. 

The first category is concerned with modifying the conventional Pareto dominance relation for many-objective optimization. Along this line, some algorithms relax the dominance condition, such as the $\epsilon$-dominance \cite{Laumanns2002combin} and the strengthened dominance relation \cite{Tian2019strengthened}. Some other algorithms develop new dominance relations, such as the fuzzy Pareto dominance~\cite{He2014fuzzy}, the grid-based dominance~\cite{Yang2013grid}, and the angle dominance \cite{Liu2020angle}. Overall, compared with the Pareto dominance relation, these dominance relations allow a solution to be easily dominated by other solutions in the high-dimensional space, thus enhancing the selection pressure towards the Pareto front.

The second category is concerned with modifying density estimation of the conventional Pareto-based algorithms since maintaining diverse nondominated solutions may harm the convergence of the population evolving towards the Pareto front in the high-dimensional space~\cite{Purshouse2007on}. Some algorithms in this category weaken the diversity maintenance mechanism, such as~\cite{Wagner2007pareto, Adra2010diversity}, whereas some other algorithms incorporate convergence information into density estimation, such as SPEA2+SDE~\cite{Li2014sde}, which proposes a shift-based density estimation (SDE) strategy to enable poorly converged solutions to be penalized by high density values. 

The third category refers to decomposition-based algorithms. Such algorithms decompose an MaOP into a set of subproblems (single-objective optimization subproblems~\cite{Zhang2007moead} or simple multiobjective subproblems~\cite{Liu2014decomposition}), and optimize these subproblems simultaneously. Specifically, the comparison between solutions is based on their scalar values on a weight vector, thus being unaffected by the increase of the objective dimensionality. Since the emergence of the multiobjective evolutionary algorithm based on decomposition (MOEA/D)~\cite{Zhang2007moead}, decomposition-based algorithms have enjoyed popularity in the area. Some representative algorithms include the reference-point-based many-objective evolutionary algorithm following NSGA-II framework (NSGA-III)~\cite{Deb2014nsga_iii}, the reference vector-guided evolutionary algorithm (RVEA)~\cite{Cheng2016reference}, the evolutionary many-objective optimization algorithm based on dominance and decomposition (MOEA/DD)~\cite{Li2015MOEADD}, and the decomposition-based multiobjective evolutionary algorithm with weights updated adaptively (DMEA-WUA)~\cite{Liu2021decomposition}.

The fourth category of MaOEAs consists of algorithms based on indicators. Such algorithms adopt performance indicators as selection criteria to guide the search of a population towards the Pareto front. Some algorithms in this category tend to employ one specific indicator, such as IBEA using the $I_{\epsilon+}$ indicator~\cite{Zitzler2004indicator}, the hypervolume estimation algorithm for multiobjective optimization (HypE) using the hypervolume (HV) indicator~\cite{Bader2011hype}, and the inverted generational distance (IGD) based many-objective evolutionary algorithm (MaOEA/IGD) using the IGD indicator~\cite{Sun2019igd}. In contrast, some others adopt multiple indicators, with the aim of obtaining a balance between the indicators, such as the stochastic ranking-based multi-indicator algorithm (SRA) using two indicators $I_{\epsilon+}$ and $I_{SDE}$~\cite{Li2016stochastic}.

The last category is the aggregation-based approaches. They aggregate the objectives of solutions into one or multiple criteria to make solutions comparable. Some algorithms in this category develop novel selection methods through estimating solutions’ performance regarding convergence and diversity, such as the many-objective evolutionary algorithm using a one-by-one selection strategy (1by1EA)~\cite{Liu20171by1EA} and the many-objective evolutionary algorithms based on coordinated selection strategy (MaOEA-CSS)~\cite{He2017MaOEACSS}. Some other algorithms integrate (assumed) preference information into the fitness assignment, such as KnEA for the knee points~\cite{Zhang2014knee} and the preference-inspired coevolutionary algorithm (PICEA-g) for the goals (i.e., target vectors)~\cite{Wang2013preference}.

Despite the advances above, there are few algorithms that are able to work well on a variety of MaOPs effectively and efficiently.
For example, it is difficult for the aggregation-based algorithms to strike a balance between convergence and diversity. This also applies to the algorithms modifying Pareto dominance or diversity maintenance. The decomposition-based algorithms often fail to maintain diversity on problems with irregular Pareto front shapes. The computational complexity of hypervolume-based algorithms can dramatically increase with the increase of the number of objectives. In addition, many algorithms need some extra parameters, particularly for the algorithms modifying the Pareto dominance relation; unfortunately, the \textit{sweet-spot} of such parameters is difficult to find in many-objective optimization. Given the above, we propose a many-objective evolutionary algorithm (called E3A), with the aim of achieving both high effectiveness and efficiency, and without any extra parameter.

\section{The Proposed Algorithm} \label{sec:E3A_Proposed_Algorithm}
\subsection{Framework of E3A}
The framework of the proposed E3A is described in Algorithm~\ref{alg:E3AFramework}. The basic procedure follows many MaOEAs. First, a population $\bm{P}$ containing $n$ solutions is randomly generated. Next, each solution in $\bm{P}$ is assigned a fitness value according to its nondominated rank. Then, the mating selection is applied to select promising solutions based on their nondominated ranks, followed by the variation operation to generate an offspring population. Finally, the environmental selection procedure is performed to preserve the $n$ best solutions for the next-generation evolution. Like most of the existing MaOEAs (e.g., NSGA-III~\cite{Deb2014nsga_iii}), 
the proposed algorithm focuses on the design of environmental selection, an operation which plays a key role in algorithm performance in many-objective optimization.


\begin{algorithm}[h]
	\caption{Framework of E3A}
	\label{alg:E3AFramework}
	\begin{algorithmic}[1]
		\begin{small}
			\REQUIRE{$\bm{P}$ (population), $n$ (population size)}
			\STATE $\bm{P} \gets Initialize\_population(n) $
			\STATE $Front \gets Pareto\_nondominated\_sort(\bm{P})$;
			\WHILE{the termination criterion is not met}
			\STATE $\bm{P}' \gets Mating\_selection(\bm{P}, Front)$
			\STATE $\bm{P}'' \gets Variation(\bm{P}')$
			\STATE $\bm{P} \gets Environmental\_selection(\bm{P} \cup \bm{P}'' )$
			\ENDWHILE
			\RETURN $\bm{P}$
		\end{small}
	\end{algorithmic}
\end{algorithm}

\subsection{Environmental Selection}
\subsubsection{Main Procedure}
Environmental selection is an evolutionary operation to determine the survival of solutions (i.e., next-generation population) from the current population and their offspring. Algorithm~\ref{alg:E3AEnvironmental_selection} shows the main procedure of our method. First, the combined set of the current population and their offspring are sorted into different fronts ($\bm{F}_{1}, \bm{F}_{2},\dots$) using the nondominated sorting method \cite{Deb2002nsga_ii} (line $4$). Then, fronts are selected one by one to construct a new population $\bm{P}$, starting from $\bm{F}_1$, until the critical front $\bm{F}_{i}$ is found, where $|\bm{F}_{1} \cup \bm{F}_{2} \cup \dots \cup \bm{F}_{i-1}| < n$ and $|\bm{F}_{1} \cup \bm{F}_{2} \cup \dots \cup \bm{F}_{i-1} \cup \bm{F}_{i}| \geq n$ ($n$ denotes the population size) (lines $5$--$8$). In fact, for MaOPs, the solutions are typically Pareto nondominated to each other and consequently the critical front is usually the first front, namely $i$ is equal to $1$. When the size of the population exceeds a predefined capacity, a population maintenance operation (i.e., the function $Population\_maintenance$ in line $15$) will be implemented to select promising solutions from the critical front. For simplicity, we use ``candidate set'' to represent all the solutions in the critical front. The population maintenance method can break down into two operations:
selecting boundary solutions and selecting non-boundary solutions as shown in Algorithm~\ref{alg:E3A}. In the next two sections, we will introduce them respectively.

\renewcommand{\algorithmiccomment}[1]{$/^{*}$ #1 $^{*}/$ }
{\renewcommand\baselinestretch{1.0}\selectfont
	\begin{algorithm}[h]
		\caption{$Environmental\_selection(\bm{P})$}
		\label{alg:E3AEnvironmental_selection}
		\begin{algorithmic}[1]
			\begin{small}
			    \REQUIRE {$\bm{P}$ (combination of the current population and the newly produced solutions)}
				\STATE $\bm{Q} \gets \emptyset $
				\STATE $\bm{S} \gets \emptyset $
				\STATE $i \gets 1 $
				\STATE $(\bm{F}_{1}, \bm{F}_{2}, \dots) \gets Pareto\_nondominated\_sort(\bm{P}) $
				\WHILE{ $|\bm{S}| + |\bm{F}_{i}| < n $}
    				\STATE $\bm{S} \gets  \bm{S} \cup \bm{F}_{i} $
    				\STATE $i \gets i + 1 $
				\ENDWHILE
				\STATE $\bm{F}_{l} = \bm{F}_{i}$ \,  /* Find the critical front  */
				\IF {$|\bm{S}| = n$}
				    \STATE $ \bm{Q} \gets \bm{S}$
				\ELSE
				    \STATE $ \bm{Q} \gets \cup_{j=1}^{l-1}F_{j} $
				    \STATE $ \bm{Q} \gets Population\_maintenance(\bm{F}_{l}, n-|\bm{Q}|) $ \; /* Select $n-|\bm{Q}|$ promising solutions from $\bm{F}_{l}$ for the next-generation evolution */
				\ENDIF
				\RETURN {$\bm{Q}$}
			\end{small}
		\end{algorithmic}
	\end{algorithm}
	\par}

\renewcommand{\algorithmiccomment}[1]{$/^{*}$ #1 $^{*}/$ }
{\renewcommand\baselinestretch{1.0}\selectfont
	\begin{algorithm}[h]
		\caption{$Population\_maintenance(\bm{P},k)$}
		\label{alg:E3A}
		\begin{algorithmic}[1]
			\begin{small}
			    \REQUIRE {$\bm{P}$ (candidate set), $k$ (number of solutions to be selected from the candidate set)}
				\STATE $\bm{Q} \gets Boundary\_solution\_selection (\bm{P})$	\\  /* Select boundary solutions from $\bm{P}$ */
			    \STATE $\bm{Q} \gets Nonboundary\_solution\_selection(\bm{P},\bm{Q},k)$  \\  /*Select high-quality non-boundary solutions from $\bm{P}$ */
			    \RETURN {$\bm{Q}$}
			\end{small}
		\end{algorithmic}
	\end{algorithm}
	\par}

\subsubsection{Selecting Boundary Solutions}\label{sec:E3AboundarySolution}
Given a candidate set $\bm{P}$, we first select $m$ ($m$ denotes the number of objectives) boundary solutions from $\bm{P}$ and place them into a set $\bm{Q}$ (which stores the solutions for the next-generation evolution). Specifically, boundary solution $\bm{b}_{j}$ corresponds to the one in $\bm{P}$ that minimizes a modified version of the Tchebycheff function $agg(\bm{x},\bm{w}_j)$ \cite{Deb2014nsga_iii}:

\begin{equation}\label{eqn_boundary}
\bm{b}_{j} = {\text{arg}} \:\min_{\bm{x}\in P}agg(\bm{x},\bm{w}_j)
\end{equation}

\begin{equation}\label{eqn_ModifiedTch}
agg(\bm{x},\bm{w}_j) = \max_{i=1}^{m}\bigg\{\frac{1}{\bm{w}_{j,i}}|\hat{f_i}(\bm{x})-z_i^{min}| \bigg\}
\end{equation}
where $\hat{f_i}(\bm{x})$ is the $i$th normalized objective value of individual $\bm{x}$, $z_i^{min}$ is the minimum value of the $i$th objective for all solutions in the candidate set, and $\bm{w}_j=(w_{j,1}$, $w_{j,2}$, $\dots$, $w_{j,m})^T$ is a weight vector very close to the $j$th objective axis direction:
\begin{equation}\label{eqn_boundaryWeightVector}
\bm{w}_{j,i} = \left \{ \begin{array}{@{}ll}
~1, & i = j\\
~10^{-6}, & i \neq j
\end{array}\right.
\end{equation}
In the case that multiple weight vectors share the same boundary solutions, we delete duplicate boundary solutions. 

Selecting boundary solutions determines the range of the solution set approximating the Pareto front for the preprocessing of selecting non-boundary solutions as described in the next section. Algorithm~\ref{alg:E3A_boundary_solutions} gives the procedure of selecting boundary solutions.

\renewcommand{\algorithmiccomment}[1]{$/^{*}$ #1 $^{*}/$ }
{\renewcommand\baselinestretch{1.0}\selectfont
	\begin{algorithm}[h]
		\caption{$Boundary\_solution\_selection(\bm{P})$}
		\label{alg:E3A_boundary_solutions}
		\begin{algorithmic}[1]
			\begin{small}
			    \REQUIRE {$\bm{P}$ (candidate set), $m$ (number of objectives)}
				\STATE $\bm{Q} \gets \emptyset $
			    \FOR{$j \gets 1$ to $m$}
			    \STATE $\bm{b}_{j} \gets {\text{arg}}\:\min_{\bm{x}\in P}agg(\bm{x},\bm{w}_j)$
			    \ENDFOR
			    \STATE $\bm{Q} \gets \{\bm{b}_{1}, \bm{b}_{2}, \dots, \bm{b}_{m}\}$
			    \STATE $\bm{Q} \gets Unique(\bm{Q})$; \; \;\;  /* Delete duplicate boundary solutions */
			    \STATE $Normalization(\bm{P}, \bm{Q})$ \; \;\; /* Normalize the objectives */
			    \RETURN {$\bm{Q}$}
			\end{small}
		\end{algorithmic}
	\end{algorithm}
	\par}

In Algorithm~\ref{alg:E3A_boundary_solutions}, the function $Normalization$ in line $7$ is utilized to normalize objectives of solutions in the candidate set. In particular, we adopt the normalization method proposed in NSGA-III~\cite{Deb2014nsga_iii}.

\subsubsection{Selecting Non-Boundary Solutions}
To maintain a representative solution set that approximate the Pareto front, we need to further distinguish between the remaining solutions (i.e., non-boundary solutions). Inspired by SDE~\cite{Li2014sde}, we consider shifted positions of solutions to reflect their convergence and diversity. The main difference of SDE and our algorithms will be described in the end of this section. The process of selecting non-boundary solutions is described below.

After the boundary solutions are selected, now we want to select solutions far away from them (thus good diversity), but at the same time take into account how these solutions perform in terms of their closeness to the Pareto front, relative to other solutions. That is, we want to select solutions which have a good balance between diversity and convergence, in which the diversity is measured based on the solutions we have already selected whereas the convergence is concerned with the comparison between the unselected solutions. To do so, we propose a step-by-step selection approach.

First, we define a metric ($sd$) to reflect the quality of the unselected solutions in terms of both convergence and diversity, relative to the selected solutions:
\begin{equation}
\label{eqn_indicatorE3A}
sd(\bm{x}, \bm{Q}) = \min_{\bm{y}\in \bm{Q}}\; dist(\bm{x}, \bm{y}')
\end{equation}
where $\bm{Q}$ denotes the set of solutions which are already selected for the next-generation evolution, $dist(\bm{x}, \bm{y}')$ represents the normalized Euclidean distance between solution $\bm{x}$ and $\bm{y}'$, $\bm{x} = (\hat{f_1}(\bm{x})$, $\hat{f_2}(\bm{x})$, $\dots$, $\hat{f}_m(\bm{x}))$ is an unselected solution in the candidate set, and $\bm{y}' = (\hat{f_1}(\bm{y}')$, $\hat{f_2}(\bm{y}')$, $\dots$, $\hat{f_m}(\bm{y}'))$ is the shifted version of selected solution $\bm{y} = (\hat{f_1}(\bm{y})$, $\hat{f_2}(\bm{y})$, $\dots$, $\hat{f_m}(\bm{y}))$, which is defined as follows~\cite{Li2014sde}:
\begin{equation}
\hat{f_i}(\bm{y}')  = \left \{ \begin{array}{ll} \hat{f_i}(\bm{x}),& \textrm{if\,\,\,} \hat{f_i}(\bm{y}) < \hat{f_i}(\bm{x}) \\
\hat{f_i}(\bm{y}),& \textrm{otherwise}
\end{array}\right.
\label{eq:individualComparison}
\end{equation}
where $\hat{f_i}(\bm{x})$, $\hat{f_i}(\bm{y})$, and $\hat{f_i}(\bm{y}') $ stand for the $i$th normalized objective value of individuals $\bm{x}$, $\bm{y}$, and $\bm{y}'$, respectively. Based on $\bm{y}'$, we calculate the normalized Euclidean distance between $\bm{x}$ and $\bm{y}'$, namely $dist$, as follows:
\begin{equation}
\label{eqn_SDE}
dist(\bm{x}, \bm{y}')=\sqrt{\sum_{i=1}^{m}(\hat{f_i}(\bm{x})- \hat{f_i}(\bm{y}'))^2}
\end{equation}
where $m$ denotes the number of objectives.

Note that the $sd$ value is calculated based on Euclidean distance instead of an angle-based metric. The main difference is that an angle-based metric is often used to maintain the distribution of the solution set (such as the maximum-vector-angle-first principle in the vector angle-based evolutionary algorithm (VaEA)~\cite{Xiang2017VaEA}), while $sd$ value based on Euclidean distance could measure both diversity and convergence of a solution relative to those already selected solutions.

Second, we select high-quality non-boundary solutions based on the $sd$ metric. Specifically, first, we select the solution with the maximum $sd$ value among the unselected solutions and place it into $\bm{Q}$. Then, we update the $sd$ value of each remaining solution in the candidate set, and, again, the current best solution that has the highest $sd$ value is found and added into $\bm{Q}$. The above steps are repeated until the size of $\bm{Q}$ reaches a predefined population size.
Algorithm~\ref{alg:E3A_nonboundary_solutions} gives the procedure of selecting non-boundary solutions.

\renewcommand{\algorithmiccomment}[1]{$/^{*}$ #1 $^{*}/$ }
{\renewcommand\baselinestretch{1.0}\selectfont
	\begin{algorithm}[h]
		\caption{$Nonboundary\_solution\_selection(\bm{P},\bm{Q},k)$}
		\label{alg:E3A_nonboundary_solutions}
		\begin{algorithmic}[1]
			\begin{small}
			    \REQUIRE {$\bm{P}$ (candidate set), $\bm{Q}$ (selected solutions), $k$ (number of solutions to be selected from the candidate set)}
			    \STATE $\bm{P} \gets \bm{P} \backslash \bm{Q}$
			    \FOR{$\bm{x} \in \bm{P}$}
			    \STATE $sd(\bm{x}, \bm{Q}) \gets \mathop{\min}\limits_{\bm{y}\in \bm{Q}}\; dist(\bm{x}, \bm{y})$ \; /* Initialize the $sd$ values of all unselected solutions in the candidate set according to (\ref{eqn_SDE}) */
			    \ENDFOR
			    \WHILE{$|\bm{Q}| < k $}
			    \STATE $ \bm{s} \gets {\text{arg}} \mathop{\max}\limits_{\bm{x}\in \bm{P}}\; sd(\bm{x}, \bm{Q})$  \; /* Select non-boundary solutions from the remaining solutions in the candidate set according to (\ref{eqn_indicatorE3A}) */
			    \STATE $\bm{Q} \gets \bm{Q} \cup \{\bm{s}\}$
			    \STATE $\bm{P} \gets \bm{P} \backslash \{\bm{s}\}$
			    \FOR{$\bm{x} \in \bm{P} $}
			    \STATE $sd(\bm{x}, \bm{Q}) \gets \min \{dist(\bm{x}, \bm{s}'), sd(\bm{x}, \bm{Q})\}$  \; /* According to (\ref{eqn_indicatorE3A}) and (\ref{eqn_SDE}) */
			    \ENDFOR
			    \ENDWHILE
			    \RETURN {$\bm{Q}$}

			\end{small}
		\end{algorithmic}
	\end{algorithm}
	\par}

\subsubsection{An Example of the Population Maintenance Operation}
\begin{figure}[h]
	\begin{center}
		\footnotesize
		\begin{tabular}{@{}c@{}c@{}c@{}c@{}}
			\includegraphics[scale=0.14]{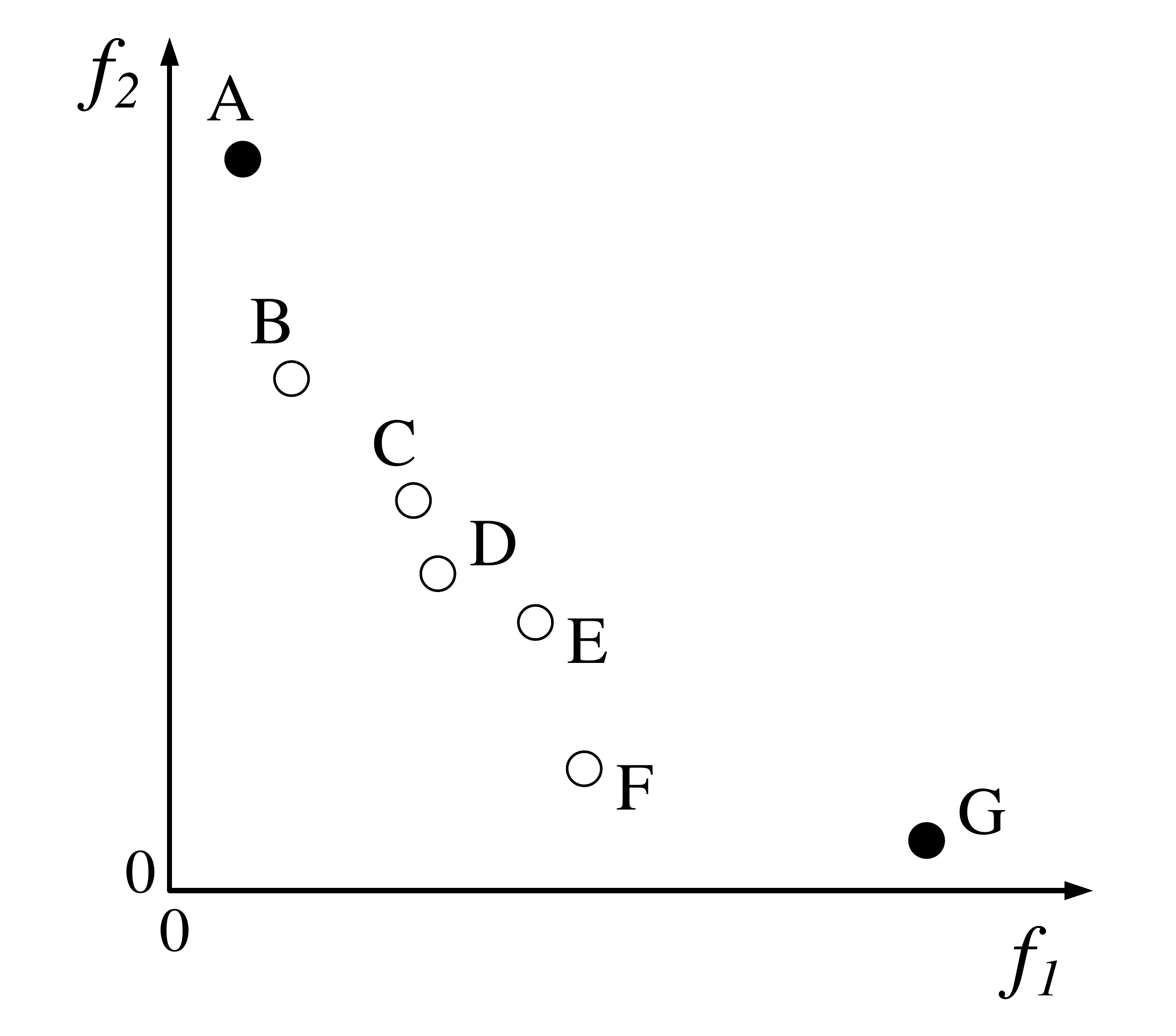}&
			\includegraphics[scale=0.14]{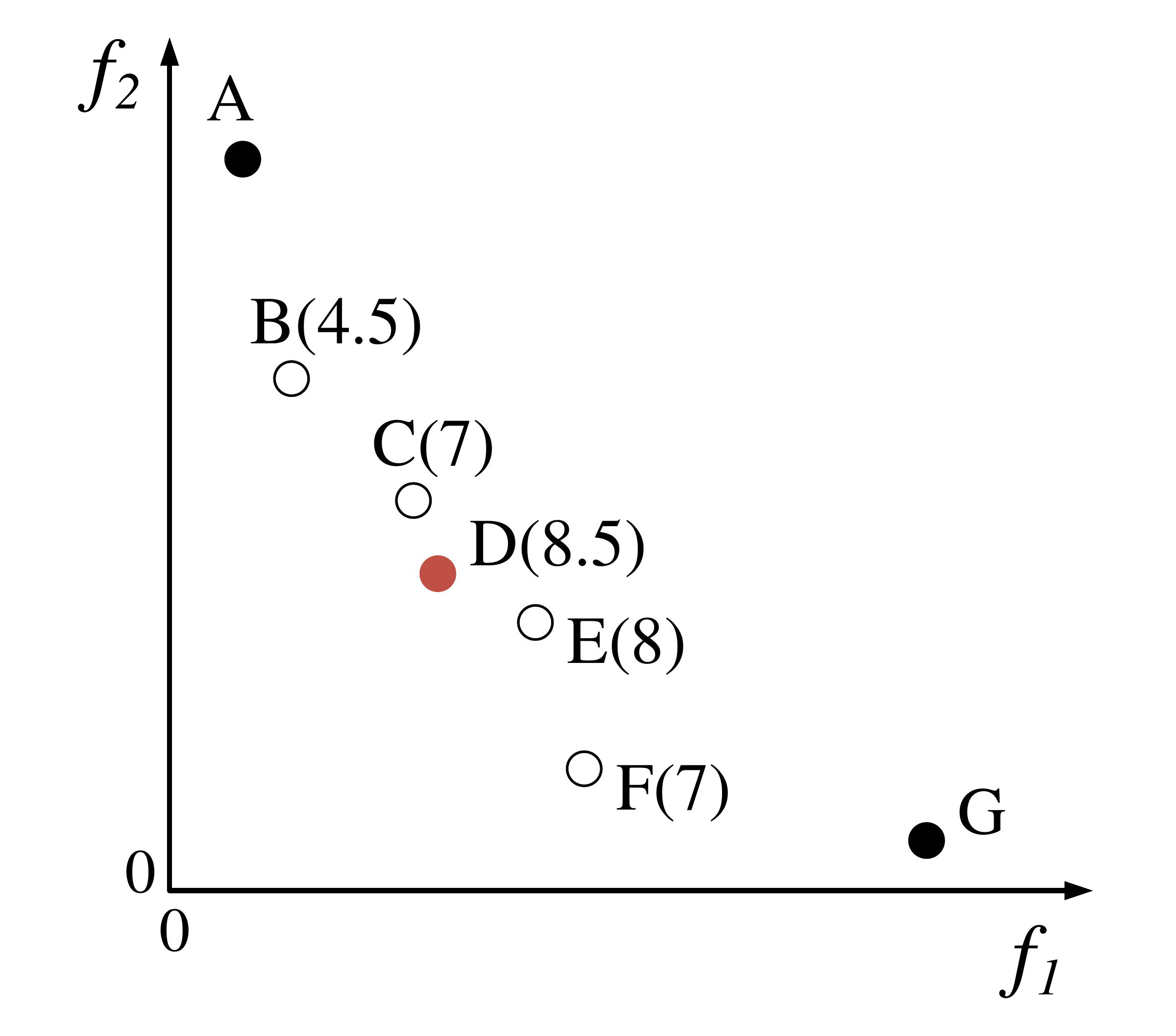}&
			\includegraphics[scale=0.14]{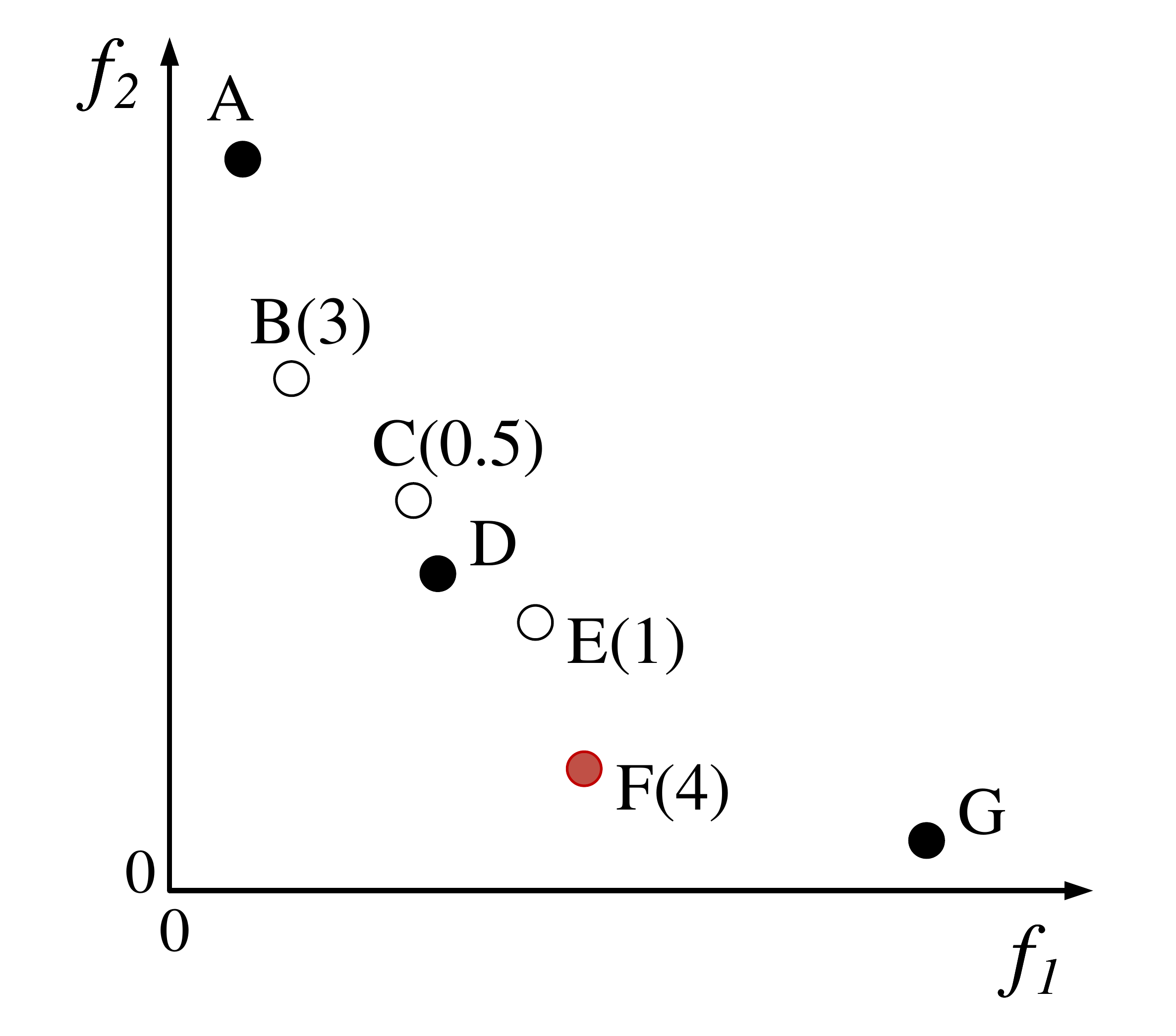}&
			\includegraphics[scale=0.14]{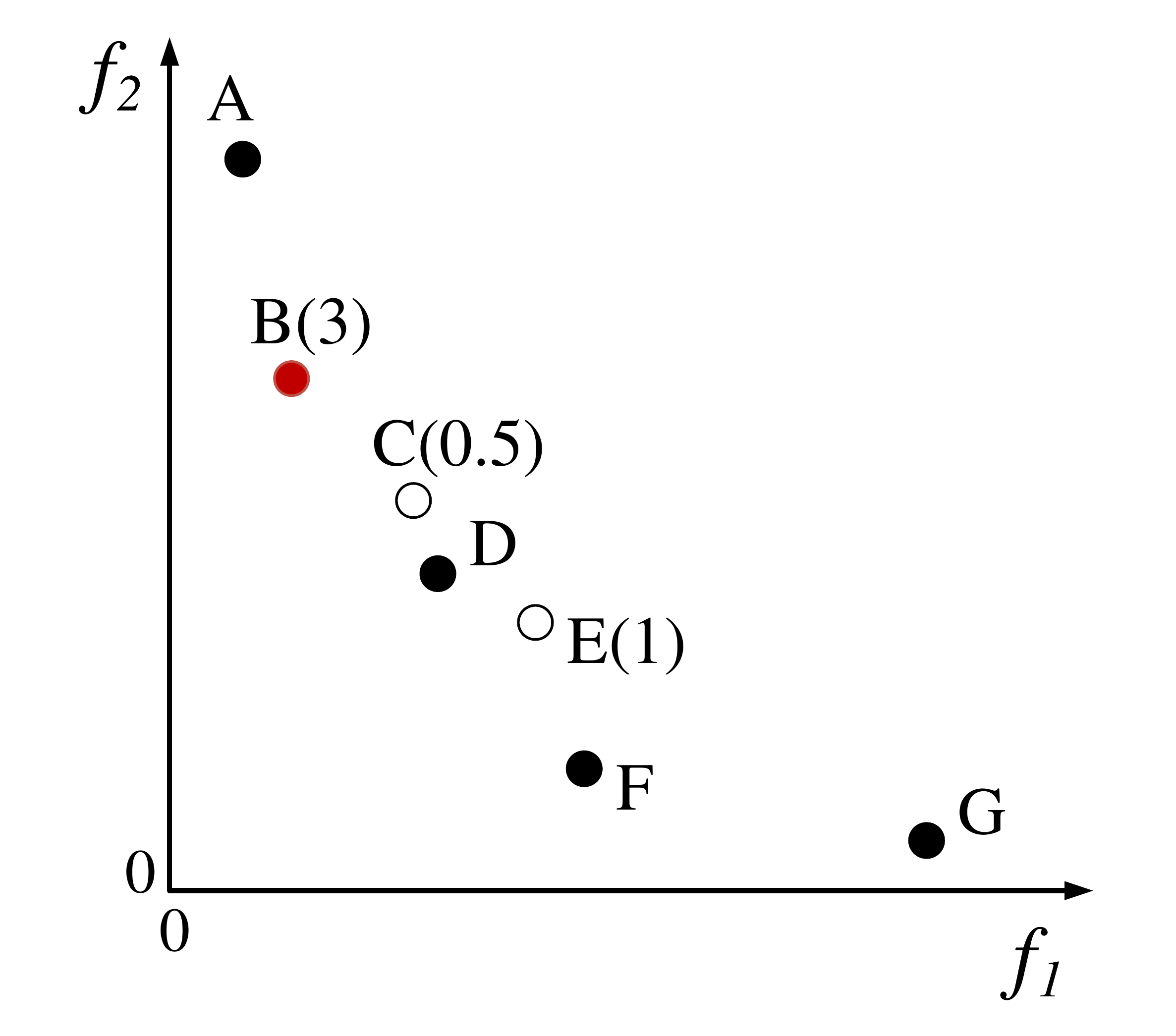}\\
			(a)  &(b)  &(c) &(d)  \\
		\end{tabular}
	\end{center}
	\caption{A bi-objective example of the population maintenance procedure, where the population size is set to five and the candidate set consists of seven solutions $\textbf{A}$--$\textbf{G}$. The number in parentheses associated with a non-boundary solution represents the $sd$ value of that solution. (a) Selecting $\textbf{A}$ and $\textbf{G}$ (since they are two boundary solutions). (b) Selecting $\textbf{D}$ (since it has the maximum $sd$ value of $8.5$ among non-boundary solutions $\textbf{B}$--$\textbf{F}$). (c) Selecting $\textbf{F}$ (since it has the maximum $sd$ value of $4$ among non-boundary solutions $\textbf{B}$, $\textbf{C}$, $\textbf{E}$, $\textbf{F}$). (d) Selecting $\textbf{B}$ (since it has the maximum $sd$ value of $3$ among non-boundary solutions $\textbf{B}$, $\textbf{C}$, $\textbf{E}$).}
	\label{fig:threeIterations}
\end{figure}

\begin{figure}[h]
	\begin{center}
		\includegraphics[scale=0.95]{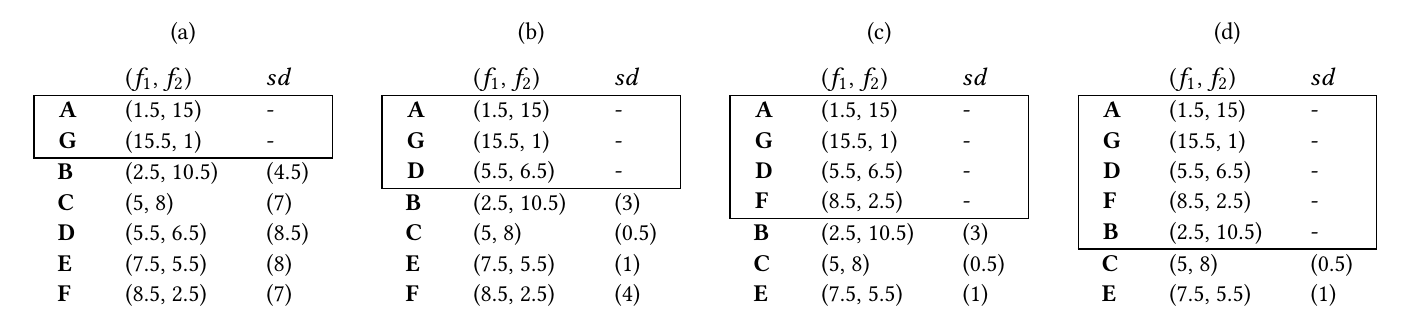}
	\end{center}
	\caption{An illustration of how the $sd$ value of each solution changes during selecting non-boundary solutions. The framed solutions mean that they have been selected. The population size is set to five.}
	\label{fig:sd_changes}
\end{figure}

Consider a bi-objective minimization problem where a set of seven candidate solutions $\textbf{A}$--$\textbf{G}$ to be selected. Figure~\ref{fig:threeIterations} shows the procedure of population maintenance in E3A. The actual objective values ($f_1$, $f_2$) and $sd$ values of the non-boundary solutions in the candidate set are given in Figure~\ref{fig:sd_changes}.

First, $\textbf{A}$ and $\textbf{G}$ are selected (i.e., $\bm{Q} = \{\textbf{A}$, $\textbf{G}\}$) since both $\textbf{A}$ and $\textbf{G}$ are boundary solutions identified by~(\ref{eqn_boundary}) and (\ref{eqn_ModifiedTch}), as shown in Figure~\ref{fig:threeIterations} (a). For the non-boundary solutions $\textbf{B}$--$\textbf{F}$, we calculate their $sd$ values, as shown in Figure~\ref{fig:sd_changes} (a).

Second, the non-boundary solution $\textbf{D}$ is selected (i.e., $\bm{Q} = \{\textbf{A}$, $\textbf{G}$, $\textbf{D}\}$) since it has the maximum $sd$ value among non-boundary solutions $\textbf{B}$--$\textbf{F}$ (Figure~\ref{fig:threeIterations} (b)). Accordingly, the $sd$ values of the unselected solutions $\textbf{B}$, $\textbf{C}$, $\textbf{E}$, $\textbf{F}$ are updated (Figure~\ref{fig:sd_changes} (b)) according to lines $9$--$11$ in Algorithm~\ref{alg:E3A_nonboundary_solutions}.

Third, the non-boundary solution $\textbf{F}$ is chosen (i.e., $\bm{Q} = \{\textbf{A}$, $\textbf{G}$, $\textbf{D}$, $\textbf{F}\}$) since it has the maximum $sd$ value among the non-boundary solutions $\textbf{B}$, $\textbf{C}$, $\textbf{E}$, $\textbf{F}$ (Figure~\ref{fig:threeIterations} (c)). The $sd$ values of the unselected solutions $\textbf{B}$, $\textbf{C}$, $\textbf{E}$ remain unchanged (Figure~\ref{fig:sd_changes} (c)) since the Euclidean distance between these solutions and the shifted version of $\textbf{F}$, namely $dist$ values, are no less than their original $sd$ values.

Finally, the non-boundary solution $\textbf{B}$ is selected since it has the maximum $sd$ value among the non-boundary solutions $\textbf{B}$, $\textbf{C}$, $\textbf{E}$ (Figure~\ref{fig:threeIterations} (d)). The $sd$ values of $\textbf{C}$ and $\textbf{E}$ remain unchanged (Figure~\ref{fig:sd_changes} (d)) since the Euclidean distance between these solutions and the shifted version of $\textbf{B}$, namely $dist$ values, are no less than their original $sd$ values. The final selected solutions are $\textbf{A}$, $\textbf{G}$, $\textbf{D}$, $\textbf{F}$, $\textbf{B}$.

Overall, two key operations of the population maintenance in E3A, i.e., selecting boundary solutions and selecting non-boundary solutions, work collaboratively. The former is to preserve solutions which perform best on at least one objective. On the basis of these selected boundary solutions, the latter is to preserve well-balanced solutions step by step. As can be seen in Figure~\ref{fig:threeIterations}, the best balanced solution, relative to the current selected solutions, is selected in turn; i.e., select $\textbf{D}$ (relative to $\textbf{A}$ and $\textbf{G}$), select $\textbf{F}$ (relative to $\textbf{A}$, $\textbf{G}$, and $\textbf{D}$), and select $\textbf{B}$ (relative to $\textbf{A}$, $\textbf{G}$, $\textbf{D}$ and $\textbf{F}$).

\subsection{Time Complexity}
Considering a situation having $m$ objectives and the evolutionary population consisting of $n$ individuals. Here, we assume $n >> m$ and consider the worst case where all solutions are nondominated with each other in the population. In one generation of E3A, nondominated sorting (line $4$ in Algorithm~\ref{alg:E3AEnvironmental_selection}) of the population requires $O(mn^2)$ comparisons.
In the environmental selection, selecting boundary solutions (lines $2$--$4$ in Algorithm~\ref{alg:E3A_boundary_solutions}) requires $O(m^2n)$ comparisons. The normalization of the candidate solutions (line $7$ in Algorithm~\ref{alg:E3A_boundary_solutions}) requires $O(n)$ comparisons. Initializing the $sd$ values of all the unselected solutions in the candidate set (lines $2$--$4$ in Algorithm~\ref{alg:E3A_nonboundary_solutions}) requires $O(m^2n)$ computations. Selecting high-quality non-boundary solutions from the unselected solutions (line $6$ in Algorithm~\ref{alg:E3A_nonboundary_solutions}) requires $O(n)$ computations. Updating the $sd$ values for the unselected solutions (lines $7$--$11$ in Algorithm~\ref{alg:E3A_nonboundary_solutions}) requires $O(mn)$ computations. In summary, by taking into account all the above operations, the overall time complexity of one generation of E3A in the worst case is $O(mn^2)$.

\subsection{Comparison with SPEA2+SDE}
Since both our algorithm and SPEA2+SDE~\cite{Li2014sde} use the shift-based comparison between solutions in the selection procedure, we would like to point out their differences. In SPEA2+SDE, the solutions are shifted for the density estimation, thus the performance depends heavily on the accuracy of the density estimation operator of the original SPEA2. In contrast, E3A is a standalone algorithm which focuses on how to select a set of well-converged and diverse solutions. Furthermore, E3A preserves the boundary solutions explicitly, but SPEA2+SDE does not and may lose the boundary solutions of the Pareto front in some problems, as shown in~\cite{Li2016stochastic}. In addition, the time complexity of E3A is $O(mn^2)$, lower than $O(mn^3)$ of SPEA2+SDE.

The above differences result in E3A outperforming SPEA2+SDE in terms of both performance and computational time, as we will see in the next section.

\section{Experimental Results} \label{sec:E3A_Experimental_Results}
\subsection{Experimental Design} \label{sec:E3A_Experimental_Design}
To assess the performance of E3A, we consider the following test suite, performance metric, state-of-the-art peer algorithms, and parameter settings, which are summarized in Table \ref{Table:allsettings}.

\begin{table}[h]
    \begin{center}
    \begin{scriptsize}
	\caption{Experimental setup}
	\label{Table:allsettings}
	\renewcommand{\arraystretch}{1.2} 
        \begin{tabular}{ll}
        \toprule
        \textbf{Factor} & \textbf{Details} \\
        \midrule
        Test problems & MaF1--15 \cite{Cheng2017MaF} \\ \hline
        Performance metrics & IGD \cite{Coello2004IGD}, HV \cite{Zitzler1999}  \\ \hline
        Number of objectives ($m$) & $\{3$, $5$, $10$, $15\}$ \\ \hline
        Population size ($n$) for $m =\{3$, $5$, $10$, $15\}$  & $\{105$, $126$, $230$, $240\}$ \\ \hline
        SBX \cite{Deb1995simulated} probability ($p_c $)  & 1.0 \\ \hline
        PM \cite{Deb1996combined} probability ($p_m$) &  $1/d$ \\ \hline
        Distribution index for SBX ($\eta_c$)&  $20$ \\ \hline
        Distribution index for PM ($\eta_m$ )&  $20$ \\ \hline
        Number of runs & $30$ \\ \hline
        Maximum of generations ($MaxGen$) & $300$ \\ \hline
        Parameter values of RVEA & $ \alpha = 2 $, $ f_r = 0.1 $ \\ \hline
        Parameter values of MOEA/DD & $ T = 0.1n $, $ \delta = 0.9$, $ \theta = 5 $ \\ \hline
        Parameter values of 1by1EA & $ k = 0.1n $, $ R = 1 $ \\ \hline
        Parameter values of MaOEA-CSS & $ t = 0 $ \\ \hline
        Parameter values of SRA & $ {p}_c = 0.6 $ \\
        \bottomrule
        \end{tabular}
    \end{scriptsize}
    \end{center}
\end{table}

\subsubsection{Test Problems}
The MaF test suite~\cite{Cheng2017MaF} is a continuous benchmark test suite proposed for the CEC'2017 competition on evolutionary many-objective optimization. In MaF, $15$ test functions with diverse properties, which are selected or modified from existing test problems, aim to provide a good representation of various real-world scenarios. 
The characteristics of all the MaF problems are summarized in Table \ref{Table:MaF Characteristics}. The number of objectives is set to $m =3$, $5$, $10$, $15$. Their parameters are set according to~\cite{Cheng2017MaF}.

\begin{table}[h]
    \begin{center}
    \begin{scriptsize}
	\caption{Characteristics of test problems in MaF. }
	\label{Table:MaF Characteristics}
	\renewcommand{\arraystretch}{1.2} 
        \begin{tabular}{@{}cc@{}}
        \toprule
        \textbf{Problem} & \textbf{Characteristics}  \\ \midrule
        MaF1 & \makecell{Linear, no single optimal solution in any subset of objectives} \\ \hline
        MaF2 & \makecell{Concave, no single optimal solution in any subset of objectives} \\ \hline
        MaF3 & Convex, multimodal  \\ \hline
        MaF4 & \makecell{Concave, multimodal, badly scaled and no single optimal \\solution in any subset of objectives}  \\ \hline
        MaF5 & Convex, biased, badly scaled \\ \hline
        MaF6 & Concave, degenerate  \\ \hline
        MaF7 & Mixed, disconnected, multimodal  \\ \hline
        MaF8 & Linear, degenerate   \\ \hline
        MaF9 & \makecell{Linear, degenerate, Pareto optimal solutions are similar to\\ their image in the objective space} \\ \hline
        MaF10 & Mixed, biased  \\ \hline
        MaF11 & Convex, disconnected, nonseparable  \\ \hline
        MaF12 & Concave, nonseparable, biased deceptive  \\ \hline
        MaF13 & \makecell{Concave, unimodal, nonseparable, degenerate, complex Pareto set} \\ \hline
        MaF14 & \makecell{Linear, partially separable, large scale, non-uniform correlations \\between decision variables and objective functions} \\ \hline
        MaF15 & \makecell{Convex, partially separable, large scale, non-uniform correlations \\between decision variables and objective functions} \\ \bottomrule
        \end{tabular}
    \end{scriptsize}
	\end{center}
\end{table}

\subsubsection{Performance Metrics}
Two commonly used performance metrics, IGD~\cite{Coello2004IGD} and HV \cite{Zitzler1999} are adopted to assess algorithm performance. The IGD and HV metrics can reflect the quality of a solution set in terms of convergence (i.e., the closeness to the true Pareto front) and diversity (i.e., the distribution over the whole Pareto front). 
Following the suggestions in~\cite{Li2019quality}, we test the Pareto dominance relation among the solution sets obtained by different algorithms. The results indicate that the compared solution sets in the experiments are nondominated to each other, and therefore, IGD is well-suited for the experiments.
Let $\bm{P}^{*}$ denotes a set of uniformly distributed reference points on the Pareto front and $\bm{P}$ represents a solution set, the IGD is the average distance from points in the set $\bm{P}^{*}$ to their nearest solution in the set $\bm{P}$, which is given by
\begin{equation}
\textrm{IGD} = \frac {1}{|\bm{P}^{*}|} \sum_{\bm{z} \in \bm{P}^{*}} d(\bm{z},\bm{P})
\end{equation}
where $d(\bm{z},\bm{P})$ stands for the minimum Euclidean distance between a reference point $\bm{z}$ and its nearest solution in $\bm{P}$, and $|\bm{P}^{*}|$ represents the size of $\bm{P}^{*}$. A smaller value of IGD is preferable, which indicates better quality of the set $\bm{P}$ for approximating the Pareto front.

The HV metric calculates the volume of the objective space between a solution set and a reference point, and a larger value indicates better overall quality of the solution set. The HV can be formulated as follows \cite{Zitzler1999}:
\begin{equation}
    \textrm{HV} = \Lambda(\bigcup_{F(\bm{x}) \in \bm{P}}\{[f_1(\bm{x}), r_1]\times [f_2(\bm{x}), r_2] \times \dots \times[f_n(\bm{x}), r_m]\})
\end{equation}
    where $\Lambda(\cdot)$ represents the Lebesgue measure, $\bm{P}$ stands for a solution set, $F(\bm{x}) = (f_1 (\bm{x})$, $f_2 (\bm{x})$, $\dots$, $f_m (\bm{x}))$ denotes an objective vector in $\bm{P}$, $\bm{r} = (r_1$, $r_2$, $\dots$, $r_m)^T$ represents a reference point in the objective space, and $m$ stands for the number of objectives. 
    In the calculation of HV, we set the reference point $\bm{r}$ to $1.1$ times of the nadir point. In addition, to enhance the robustness of the algorithm when the ranges of the objective values are different, in our experiments, each objective value of a solution is normalized according to the range of the problem's Pareto front. Moreover, to reduce the computational cost of HV with respect to both the exact computation, we utilized Monte Carlo sampling with $10^6$ sampling points to approximate the HV value \cite{Bader2011hype}.

\subsubsection{Peer Algorithms}
Eleven state-of-the-art MaOEAs are chosen to evaluate the proposed algorithm. They are
NSGA-III~\cite{Deb2014nsga_iii}, RVEA~\cite{Cheng2016reference}, MOEA/DD~\cite{Li2015MOEADD}, the reference point-based dominance based NSGA-II (RPD-NSGA-II)~\cite{Elarbi2018RDP_NSGA_II}, 1by1EA~\cite{Liu20171by1EA}, MaOEA-CSS~\cite{He2017MaOEACSS}, VaEA~\cite{Xiang2017VaEA}, NSGA-II based on the strengthened dominance relation (NSGA-II/SDR)~\cite{Tian2019strengthened}, SPEA2+SDE~\cite{Li2014sde}, SRA~\cite{Li2016stochastic}, and the IGD-NS indicator based multiobjective evolutionary algorithm (AR-MOEA)~\cite{Tian2018indicator}.
These algorithms span over the main categories (cf. Section ~\ref{sec:E3A_related_work}). Specifically, NSGA-III, RVEA, MOEA/DD, RPD-NSGA-II, and VaEA are decomposition-based algorithms.
1by1EA and MaOEA-CSS are aggregation-based algorithms.
NSGA-II/SDR modifies the conventional Pareto dominance relation.
SPEA2+SDE changes the density estimation of the conventional Pareto-based algorithms. Finally, SRA and AR-MOEA are two indicator-based algorithms.

All algorithms are implemented in a recently developed Matlab platform PlatEMO\footnote{PlatEMO can be downloaded from: \url{https://github.com/BIMK/PlatEMO.}}~\cite{Tian2017platemo}. PlatEMO includes more than $50$ EMO algorithms and more than $100$ test problems (e.g., MaF test problems), as well as some widely used performance metrics (e.g., IGD metric). All the experiments are conducted on an Intel(R) Core(TM) i$7$-$2600$ CPU @ 3.40GHz with $16$ GB RAM, running on Windows $10$. To make the comparison statistically significance, the statistical tests (i.e., Friedman test and posthoc Nemenyi test) is used throughout the experiments.

\subsubsection{Parameter Settings}
For a fair comparison, we adopt the following settings for all tested algorithms. 
\begin{itemize}
    \item Each algorithm is executed $30$ runs independently on each test problem to decrease the impact of their stochastic nature.
    \item The termination criterion is specified as the maximum number of generations being set to $300$.  
    \item The simulated binary crossover \cite{Deb1995simulated} and polynomial mutation \cite{Deb1996combined} operators are employed to perform variation, with both distribution indexes being set to $20$. The crossover and mutation probabilities are set to $1.0$ and $1/d$ ($d$ denotes the number of decision variables), respectively.
    \item The population size is consistent with the number of weight vectors in decomposition-based algorithms, NSGA-III, RVEA, MOEA/DD, and RPD-NSGA-II, being set to $105$, $126$, $230$, and $240$ for test problems with three, five, ten, and fifteen objectives, respectively. In this study, two approaches for weight vector generation are utilized, i.e., the Das and Dennis's~\cite{Das1998normal} systematic approach for test problems with $m \leq 5$ and the two-layer weight vectors generation approach~\cite{Deb2014nsga_iii} for test problems with $m > 5$, where $m$ represents the number of objectives. 
\end{itemize}

Specific parameters are required in certain peer algorithms. Here we set them according to their original papers. In RVEA, the parameter $\alpha$ to control the rate of change of the penalty function is set to $2$, and the frequency $f_r$ to employ reference vector adaptation is set to $0.1$. In MOEA/DD, the neighbourhood size $T$ is set to $10$ percent of the population size, the neighbourhood selection probability $ \delta $ is set to $0.9$, and the penalty parameter $\theta$ of the PBI function is set to $5$. In 1by1EA, the parameter $k$ to balance the computational cost and the accuracy in density estimation is set to $10$ percent of the population size, and the parameter $R$ of controlling distribution threshold is set to $1$. In MaOEA-CSS, the threshold value $t$ of determining the difference of two closest solutions' Euclidean distance in the environmental selection is set to $0$. For SRA, the parameter ${p}_c$ for the purpose of balancing the different indicators in the stochastic ranking strategy is set to $0.6$.

\subsection{Performance Comparison}
\subsubsection{Algorithm Performance on Tri-Objective Problems}
Tables \ref{Table:3obj_MaF_IGD} and \ref{Table:3obj_MaF_HV} show the mean and standard deviation (in parentheses) of the IGD and HV results obtained by E3A and the peer algorithms on the tri-objective MaF, respectively. From Table \ref{Table:3obj_MaF_IGD}, it can be seen that E3A obtains the best IGD results on seven test problems
(i.e., MaF$1$--$4$ and MaF$8$--$10$). Concerning pairwise comparison, the proportions of the test problems where E3A significantly outperforms NSGA-III, RVEA, MOEA/DD, RPD-NSGA-II, 1by1EA, MaOEA-CSS, VaEA, NSGA-II/SDR, SPEA2+SDE, SRA, and AR-MOEA are $7/15$, $12/15$, $8/15$, $11/15$, $8/15$, $7/15$, $5/15$, $9/15$, $4/15$, $9/15$, and $4/15$, respectively. Conversely, the proportions of the test problems where E3A performs significantly worse than these peer algorithms are $3/15$, $2/15$, $1/15$, $2/15$, $2/15$, $1/15$, $4/15$, $0/15$, $1/15$, $0/15$, and $3/15$, respectively. 

From Table \ref{Table:3obj_MaF_HV}, it can be seen that E3A obtains the best HV results on $11$ test problems (i.e., MaF$1$--$3$ and MaF$6$--$13$). Concerning pairwise comparison, 
the proportions of the test problems where E3A significantly outperforms NSGA-III, RVEA, MOEA/DD, RPD-NSGA-II, 1by1EA, MaOEA-CSS, VaEA, NSGA-II/SDR, SPEA2+SDE, SRA, and AR-MOEA are $11/15$, $13/15$, $10/15$, $11/15$, $12/15$, $11/15$, $9/15$, $9/15$, $1/15$, $11/15$, and $5/15$, respectively. Conversely, the proportions of the test problems where E3A performs significantly worse than these peer algorithms are $0/15$, $0/15$, $1/15$, $0/15$, $1/15$, $1/15$, $0/15$, $0/15$, $1/15$, $2/15$, and $0/15$, respectively.

\begin{table}[tbp]
	\scriptsize
    \begin{center}
    \caption{Mean and standard deviation of the IGD values obtained by the $12$ algorithms on the tri-objective MaF problems.}
    \label{Table:3obj_MaF_IGD}
    \resizebox{\textwidth}{!}{%
    \renewcommand{\arraystretch}{1.2} 
    \begin{tabular}{@{}cccccccccccccc@{}}
 \toprule
Problem                & Obj.               & NSGA-III                      & RVEA        & MOEA/DD      & RPD-NSGA-II   & 1by1EA      & MaOEA-CSS    & VaEA        & NSGA-II/SDR   & SPEA2+SDE    & SRA         & AR-MOEA      & E3A      \\
  \midrule
\multirow{2}{*}{MaF1}  & \multirow{2}{*}{3} & 5.587e-2  $+$ & 7.499e-2  $+$ & 6.176e-2  $+$ & 6.348e-2  $+$ & 4.123e-2  $\approx$ & 4.135e-2  $\approx$ & 4.114e-2  $\approx$ & 3.227e-1  $+$ & 4.099e-2  $\approx$ & 4.519e-2  $+$ & 4.166e-2  $+$ & 4.008e-2 \\
&                    & (1.33e-3) & (3.01e-3)    & (1.83e-3)    & (1.48e-3)    & (2.35e-3)    & (1.16e-3)    & (5.47e-4)    & (3.07e-1)    & (5.80e-4)    & (1.25e-3)    & (5.85e-4)    & (2.12e-4) \\\hline

\multirow{2}{*}{MaF2}  & \multirow{2}{*}{3} & 3.352e-2  $+$ & 3.839e-2  $+$ & 4.839e-2  $+$ & 3.486e-2  $+$ & 2.934e-2  $\approx$ & 2.866e-2  $\approx$ & 2.875e-2  $\approx$ & 3.114e-2  $+$ & 3.023e-2  $+$ & 3.501e-2  $+$ & 2.991e-2  $+$ & 2.816e-2 \\
&                    & (9.09e-4) & (1.10e-3)    & (2.36e-3)    & (2.87e-4)    & (1.06e-3)    & (2.69e-3)    & (3.06e-4)    & (6.88e-4)    & (5.61e-4)    & (1.78e-3)    & (8.44e-4)    & (2.70e-4) \\\hline

\multirow{2}{*}{MaF3}  & \multirow{2}{*}{3} & 1.246e+0  $+$                  & 2.054e+2  $+$ & 2.213e+1  $+$ & 1.176e+0  $+$ & 5.771e-1  $+$ & 1.590e+0  $+$ & 1.212e+0  $+$ & 1.203e+0  $+$ & 5.015e-1  $\approx$ & 1.824e+0  $+$ & 4.277e-1  $\approx$ & 1.536e-1 \\
&                    & (1.76e+0) & (6.29e+2)    & (2.15e+1)    & (1.89e+0)    & (1.23e+0)    & (1.85e+0)    & (1.88e+0)    & (2.46e+0)    & (8.76e-1)    & (2.25e+0)    & (1.30e+0)    & (4.06e-1) \\\hline

\multirow{2}{*}{MaF4}  & \multirow{2}{*}{3} & 1.070e+0  $+$ & 1.875e+0  $+$ & 1.331e+0  $+$ & 5.520e-1  $+$ & 1.732e+0  $+$ & 1.056e+0  $+$ & 1.073e+0  $+$ & 9.432e-1  $+$ & 6.157e-1  $+$ & 2.226e+0  $+$ & 7.428e-1  $\approx$ & 4.396e-1 \\
&                    & (1.19e+0) & (1.87e+0)    & (1.04e+0)    & (4.43e-1)    & (5.48e-1)    & (1.05e+0)    & (1.93e+0)    & (9.62e-1)    & (5.87e-1)    & (2.43e+0)    & (1.05e+0)    & (5.11e-1) \\\hline

\multirow{2}{*}{MaF5}  & \multirow{2}{*}{3} & 4.432e-1  $-$                  & 2.822e-1  $-$ & 3.189e-1  $\approx$ & 2.465e-1  $-$ & 6.565e-1  $\approx$ & 3.221e-1  $-$ & 2.875e-1  $-$ & 2.147e+0  $+$ & 1.069e+0  $\approx$ & 6.082e-1  $\approx$ & 1.247e+0  $\approx$ & 6.326e-1 \\
&                    & (5.36e-1) & (2.30e-1)    & (2.29e-1)    & (3.81e-3)    & (5.01e-1)    & (2.31e-2)    & (2.28e-1)    & (2.92e-1)    & (1.04e+0)    & (6.10e-1)    & (1.59e+0)    & (6.25e-1) \\\hline

\multirow{2}{*}{MaF6}  & \multirow{2}{*}{3} & 1.367e-2 $\approx$                  & 6.180e-2  $+$ & 2.881e-2  $+$ & 5.113e-2  $+$ & 4.334e-3  $-$ & 2.396e-2  $+$ & 4.427e-3  $-$ & 7.096e-3  $\approx$ & 9.777e-3  $\approx$ & 7.738e-3  $\approx$ & 4.371e-3  $-$ & 8.753e-3 \\
&                    & (1.41e-3) & (2.06e-2)    & (7.61e-4)    & (1.94e-2)    & (1.11e-4)    & (2.76e-3)    & (1.33e-4)    & (9.44e-4)    & (1.11e-3)    & (9.93e-4)    & (7.85e-5)    & (4.27e-4) \\\hline

\multirow{2}{*}{MaF7}  & \multirow{2}{*}{3} & 7.115e-2  $\approx$                  & 1.033e-1  $+$ & 3.711e-1  $+$ & 1.024e-1  $+$ & 9.632e-2  $+$ & 1.002e-1  $+$ & 5.954e-2  $\approx$ & 8.177e-2  $+$ & 6.756e-2  $\approx$ & 1.258e-1  $+$ & 2.248e-1  $+$ & 6.738e-2 \\
&                    & (2.69e-3) & (3.58e-3)    & (1.34e-1)    & (6.72e-2)    & (1.91e-2)    & (1.24e-2)    & (1.39e-3)    & (4.24e-3)    & (5.45e-2)    & (1.06e-1)    & (2.09e-1)    & (6.30e-2) \\\hline

\multirow{2}{*}{MaF8}  & \multirow{2}{*}{3} & 1.225e-1  $+$                  & 1.413e-1  $+$ & 1.317e-1  $+$ & 1.233e-1  $+$ & 3.415e-1  $+$ & 7.882e-2  $\approx$ & 8.251e-2  $\approx$ & 9.021e-2  $+$ & 7.713e-2  $\approx$ & 5.579e-1  $+$ & 7.950e-2  $\approx$ & 6.767e-2 \\
&                    & (3.18e-2) & (1.57e-2)    & (2.14e-2)    & (1.18e-2)    & (5.06e-2)    & (1.52e-2)    & (1.96e-2)    & (1.16e-2)    & (1.75e-2)    & (1.70e+0)    & (7.37e-3)    & (8.87e-3) \\\hline

\multirow{2}{*}{MaF9}  & \multirow{2}{*}{3} & 7.892e-2  $\approx$                  & 1.949e-1  $+$ & 6.868e-2  $\approx$ & 2.131e-1  $+$ & 2.187e-1  $+$ & 7.818e-2  $+$ & 4.420e-1  $+$ & 8.834e-2  $+$ & 6.813e-2  $\approx$ & 8.101e-2  $+$ & 7.540e-2  $\approx$ & 6.248e-2 \\
&                    & (1.72e-2) & (1.19e-1)    & (9.68e-3)    & (2.42e-1)    & (3.10e-2)    & (7.06e-3)    & (8.75e-2)    & (1.74e-2)    & (1.05e-2)    & (2.43e-2)    & (2.74e-2)    & (4.59e-3) \\\hline

\multirow{2}{*}{MaF10} & \multirow{2}{*}{3} & 4.125e-1  $+$                  & 5.276e-1  $+$ & 5.843e-1  $+$ & 2.704e-1  $\approx$ & 6.673e-1  $+$ & 4.085e-1  $+$ & 4.263e-1  $+$ & 2.573e-1  $\approx$ & 2.095e-1  $\approx$ & 6.575e-1  $+$ & 3.108e-1  $+$ & 1.880e-1 \\
&                    & (5.38e-2) & (7.61e-2)    & (1.38e-1)    & (3.83e-2)    & (7.80e-2)    & (3.92e-2)    & (5.16e-2)    & (3.79e-2)    & (2.27e-2)    & (8.64e-2)    & (3.56e-2)    & (1.97e-2) \\\hline

\multirow{2}{*}{MaF11} & \multirow{2}{*}{3} & 1.500e-1  $-$                  & 1.764e-1  $\approx$ & 1.690e-1  $\approx$ & 1.465e-1  $-$ & 2.497e-1  $+$ & 2.380e-1  $+$ & 1.607e-1  $-$ & 1.899e-1  $\approx$ & 2.023e-1  $\approx$ & 2.202e-1  $\approx$ & 1.492e-1  $-$ & 1.894e-1 \\
&                    & (1.12e-3) & (8.85e-3)    & (3.63e-3)    & (1.70e-3)    & (2.77e-2)    & (2.18e-2)    & (3.79e-3)    & (1.42e-2)    & (1.13e-2)    & (1.30e-2)    & (1.50e-3)    & (8.31e-3) \\\hline

\multirow{2}{*}{MaF12} & \multirow{2}{*}{3} & 2.095e-1  $-$                  & 2.170e-1  $-$ & 2.260e-1  $\approx$ & 2.389e-1  $\approx$ & 3.337e-1  $+$ & 2.786e-1  $\approx$ & 2.166e-1  $-$ & 2.370e-1  $\approx$ & 3.086e-1  $+$ & 2.827e-1  $\approx$ & 2.122e-1  $-$ & 2.542e-1 \\
&                    & (2.52e-3) & (4.89e-3)    & (3.49e-3)    & (4.08e-2)    & (4.85e-2)    & (1.78e-2)    & (5.12e-3)    & (7.39e-3)    & (2.43e-2)    & (1.25e-2)    & (2.41e-2)    & (5.92e-3) \\\hline

\multirow{2}{*}{MaF13} & \multirow{2}{*}{3} & 8.728e-2  $\approx$                  & 1.220e-1  $+$ & 6.954e-2  $-$ & 1.317e-1  $+$ & 7.837e-2  $-$ & 8.168e-2  $\approx$ & 9.545e-2  $\approx$ & 9.828e-2  $\approx$ & 1.182e-1  $+$ & 1.251e-1  $+$ & 8.649e-2  $\approx$ & 9.465e-2 \\
&                    & (9.63e-3) & (2.20e-2)    & (5.62e-3)    & (2.18e-2)    & (7.80e-3)    & (6.31e-3)    & (1.08e-2)    & (6.96e-3)    & (2.12e-2)    & (1.49e-2)    & (8.56e-3)    & (1.14e-2) \\\hline

\multirow{2}{*}{MaF14} & \multirow{2}{*}{3} & 1.320e+0  $\approx$                  & 2.945e+0  $+$ & 1.128e+0  $\approx$ & 1.597e+0  $+$ & 9.238e-1  $\approx$ & 8.823e-1  $\approx$ & 1.243e+0  $\approx$ & 1.398e+0  $\approx$ & 1.272e+0  $\approx$ & 1.369e+0  $\approx$ & 8.284e-1  $\approx$ & 1.017e+0 \\
&                    & (4.00e-1) & (1.07e+0)    & (2.22e-1)    & (5.24e-1)    & (2.82e-1)    & (1.24e-1)    & (4.18e-1)    & (4.72e-1)    & (4.59e-1)    & (4.55e-1)    & (1.97e-1)    & (2.54e-1) \\ \hline

\multirow{2}{*}{MaF15} & \multirow{2}{*}{3} & 7.568e-1  $+$                  & 8.138e-1  $+$ & 2.874e-1  $\approx$ & 9.375e-1  $+$ & 3.405e-1  $\approx$ & 3.426e-1  $\approx$ & 7.546e-1  $+$ & 5.902e-1  $+$ & 2.313e-1  $-$ & 2.404e-1  $\approx$ & 3.825e-1  $\approx$ & 3.585e-1 \\
&                    & (1.71e-1) & (2.10e-1)    & (4.29e-2)    & (3.93e-2)    & (6.36e-2)    & (4.41e-2)    & (7.73e-2)    & (4.34e-2)    & (2.79e-2)    & (2.56e-2)    & (6.37e-2)    & (1.18e-1) \\ \hline

\multicolumn{2}{c}{$+$/$\approx$/$-$}                   & 7/5/3  & 12/1/2  & 8/6/1  & 11/2/2    & 8/5/2     & 7/7/1   & 5/6/4   & 9/6/0    & 4/10/1   & 9/6/0  & 4/8/3      &         
\\\bottomrule 
    \end{tabular}%
    }
    \end{center}
    \vspace{6pt}
\footnotesize `$+$', `$\approx$', and `$-$' indicate that the result of E3A is significantly better than, equivalent to, and significantly worse than that of the peer algorithm at a $0.05$ level by the statistical tests (i.e., Friedman test and posthoc Nemenyi test), respectively.
\end{table}

\begin{table}[tbp]
	\scriptsize
    \begin{center}
    \caption{Mean and standard deviation of the HV values obtained by the $12$ algorithms on the tri-objective MaF problems.}
    \label{Table:3obj_MaF_HV}
    \resizebox{\textwidth}{!}{%
    \renewcommand{\arraystretch}{1.2} 
    \begin{tabular}{@{}cccccccccccccc@{}}
 \toprule
Problem                & Obj.               & NSGA-III                      & RVEA        & MOEA/DD      & RPD-NSGA-II   & 1by1EA      & MaOEA-CSS    & VaEA        & NSGA-II/SDR   & SPEA2+SDE    & SRA         & AR-MOEA      & E3A      \\ \midrule
\multirow{2}{*}{MaF1} & \multirow{2}{*}{3}  & 2.798e-01 $+$       & 2.349e-01 $+$       & 2.718e-01 $+$       & 2.663e-01 $+$       & 2.881e-01 $+$       & 2.867e-01 $+$       & 2.909e-01 $+$       & 1.723e-01 $+$       & 2.948e-01 $\approx$ & 2.872e-01 $+$       & 2.954e-01 $\approx$ & 2.981e-01  \\
        &      & (1.41e-03)          & (3.48e-02)          & (2.26e-03)          & (2.58e-03)          & (3.05e-03)          & (1.70e-03)          & (1.33e-03)          & (1.20e-01)          & (5.03e-04)          & (1.79e-03)          & (1.06e-03)          & (1.33e-04) \\\hline
\multirow{2}{*}{MaF2} & \multirow{2}{*}{3}  & 6.858e-01 $+$       & 6.632e-01 $+$       & 6.029e-01 $+$       & 6.842e-01 $+$       & 6.659e-01 $+$       & 6.638e-01 $+$       & 6.966e-01 $\approx$ & 7.006e-01 $\approx$ & 7.039e-01 $\approx$ & 6.941e-01 $+$       & 6.920e-01 $+$       & 7.048e-01  \\
        &      & (3.00e-03)          & (4.42e-03)          & (7.94e-03)          & (1.98e-03)          & (5.69e-03)          & (1.51e-02)          & (3.16e-03)          & (1.73e-03)          & (1.97e-03)          & (2.24e-03)          & (2.84e-03)          & (1.08e-03) \\\hline
\multirow{2}{*}{MaF3} & \multirow{2}{*}{3}  & 6.638e-01 $+$       & 1.619e-02 $+$       & 6.455e-03 $+$       & 8.054e-01 $+$       & 8.889e-01 $+$       & 5.088e-01 $+$       & 7.841e-01 $+$       & 9.227e-01 $\approx$ & 9.796e-01 $\approx$ & 6.181e-01 $+$       & 1.126e+00 $\approx$ & 1.171e+00  \\
        &      & (5.88e-01)          & (8.72e-02)          & (3.48e-02)          & (4.46e-01)          & (3.15e-01)          & (5.54e-01)          & (5.59e-01)          & (5.17e-01)          & (5.10e-01)          & (5.92e-01)          & (3.77e-01)          & (2.70e-01) \\\hline
\multirow{2}{*}{MaF4} & \multirow{2}{*}{3}  & 4.565e-01 $+$       & 3.416e-01 $+$       & 3.603e-01 $+$       & 6.013e-01 $+$       & 3.394e-01 $+$       & 5.133e-01 $+$       & 5.121e-01 $+$       & 5.302e-01 $+$       & 6.141e-01 $\approx$ & 2.872e-01 $+$       & 5.575e-01 $\approx$ & 6.359e-01  \\
        &      & (2.55e-01)          & (2.66e-01)          & (2.85e-01)          & (1.39e-01)          & (1.29e-01)          & (2.54e-01)          & (2.32e-01)          & (2.31e-01)          & (1.97e-01)          & (2.91e-01)          & (2.24e-01)          & (1.66e-01) \\\hline
\multirow{2}{*}{MaF5} & \multirow{2}{*}{3}  & 7.095e-01 $\approx$ & 7.388e-01 $\approx$ & 7.199e-01 $-$       & 7.447e-01 $\approx$ & 6.350e-01 $+$       & 7.031e-01 $-$       & 7.320e-01 $\approx$ & 3.608e-01 $+$       & 6.196e-01 $\approx$ & 6.897e-01 $-$       & 5.882e-01 $\approx$ & 6.816e-01  \\
        &      & (1.01e-01)          & (5.30e-02)          & (5.42e-02)          & (1.52e-03)          & (8.76e-02)          & (9.84e-03)          & (5.02e-02)          & (6.03e-02)          & (1.63e-01)          & (1.03e-01)          & (2.21e-01)          & (1.22e-01) \\\hline
\multirow{2}{*}{MaF6} & \multirow{2}{*}{3}  & 2.580e-01 $+$       & 2.174e-01 $+$       & 2.451e-01 $+$       & 2.407e-01 $+$       & 2.661e-01 $\approx$ & 2.441e-01 $+$       & 2.662e-01 $\approx$ & 2.645e-01 $+$       & 2.657e-01 $\approx$ & 2.632e-01 $+$       & 2.659e-01 $\approx$ & 2.661e-01  \\
        &      & (1.21e-03)          & (1.27e-02)          & (8.47e-04)          & (7.19e-03)          & (1.27e-04)          & (5.29e-03)          & (1.03e-04)          & (4.54e-04)          & (2.69e-04)          & (6.42e-04)          & (7.60e-05)          & (1.09e-04) \\\hline
\multirow{2}{*}{MaF7} & \multirow{2}{*}{3}   & 5.449e-01 $+$       & 5.187e-01 $+$       & 4.349e-01 $+$       & 5.433e-01 $+$       & 5.193e-01 $+$       & 5.107e-01 $+$       & 5.597e-01 $\approx$ & 5.473e-01 $+$       & 5.597e-01 $\approx$ & 5.419e-01 $+$       & 5.165e-01 $+$       & 5.633e-01  \\
        &      & (3.24e-03)          & (7.54e-03)          & (2.86e-02)          & (2.03e-02)          & (1.61e-02)          & (1.53e-02)          & (1.66e-03)          & (3.24e-03)          & (1.56e-02)          & (3.01e-02)          & (5.37e-02)          & (2.73e-02) \\\hline
\multirow{2}{*}{MaF8} & \multirow{2}{*}{3}  & 3.422e-01 $+$       & 3.238e-01 $+$       & 3.326e-01 $+$       & 3.382e-01 $+$       & 2.408e-01 $+$       & 3.658e-01 $\approx$ & 3.662e-01 $\approx$ & 3.587e-01 $+$       & 3.683e-01 $\approx$ & 2.722e-01 $+$       & 3.679e-01 $\approx$ & 3.755e-01  \\
        &      & (2.01e-02)          & (1.00e-02)          & (1.03e-02)          & (7.51e-03)          & (2.28e-02)          & (7.93e-03)          & (1.07e-02)          & (6.40e-03)          & (9.76e-03)          & (1.08e-01)          & (4.17e-03)          & (5.53e-03) \\\hline
\multirow{2}{*}{MaF9} & \multirow{2}{*}{3}   & 1.098e+00 $\approx$ & 9.921e-01 $+$       & 1.105e+00 $\approx$ & 9.636e-01 $+$       & 8.683e-01 $+$       & 1.064e+00 $+$       & 8.092e-01 $+$       & 1.063e+00 $+$       & 1.097e+00 $\approx$ & 1.093e+00 $+$       & 1.101e+00 $\approx$ & 1.113e+00  \\
        &      & (1.80e-02)          & (9.71e-02)          & (1.11e-02)          & (2.29e-01)          & (4.78e-02)          & (1.43e-02)          & (4.36e-02)          & (2.60e-02)          & (9.05e-03)          & (2.02e-02)          & (2.34e-02)          & (5.33e-03) \\\hline
\multirow{2}{*}{MaF10} & \multirow{2}{*}{3}  & 1.020e+00 $+$       & 9.474e-01 $+$       & 8.912e-01 $+$       & 1.145e+00 $\approx$ & 1.014e+00 $+$       & 1.069e+00 $+$       & 9.939e-01 $+$       & 1.129e+00 $\approx$ & 1.217e+00 $\approx$ & 8.359e-01 $+$       & 1.093e+00 $+$       & 1.221e+00  \\
        &      & (3.90e-02)          & (5.66e-02)          & (8.70e-02)          & (2.76e-02)          & (3.93e-02)          & (3.12e-02)          & (3.77e-02)          & (3.30e-02)          & (2.02e-02)          & (5.54e-02)          & (3.09e-02)          & (1.93e-02) \\\hline
\multirow{2}{*}{MaF11} & \multirow{2}{*}{3}  & 1.238e+00 $+$       & 1.226e+00 $+$       & 1.233e+00 $+$       & 1.244e+00 $\approx$ & 1.190e+00 $+$       & 1.180e+00 $+$       & 1.230e+00 $+$       & 1.215e+00 $+$       & 1.238e+00 $+$       & 1.229e+00 $+$       & 1.240e+00 $\approx$ & 1.247e+00  \\
        &      & (2.20e-03)          & (3.60e-03)          & (3.61e-03)          & (1.47e-03)          & (1.56e-02)          & (1.72e-02)          & (2.22e-03)          & (1.05e-02)          & (5.01e-03)          & (4.48e-03)          & (1.81e-03)          & (1.39e-03) \\\hline
\multirow{2}{*}{MaF12} & \multirow{2}{*}{3}  & 7.031e-01 $\approx$ & 7.002e-01 $+$       & 6.905e-01 $+$       & 6.864e-01 $+$       & 6.527e-01 $+$       & 6.501e-01 $+$       & 6.953e-01 $+$       & 7.090e-01 $\approx$ & 7.080e-01 $\approx$ & 7.019e-01 $+$       & 6.958e-01 $+$       & 7.130e-01  \\
        &      & (6.09e-03)          & (4.90e-03)          & (5.28e-03)          & (3.20e-02)          & (1.47e-02)          & (9.94e-03)          & (9.03e-03)          & (3.86e-03)          & (3.05e-02)          & (5.74e-03)          & (2.96e-02)          & (6.78e-03) \\\hline
\multirow{2}{*}{MaF13} & \multirow{2}{*}{3}   & 6.572e-01 $+$       & 5.963e-01 $+$       & 6.889e-01 $\approx$ & 6.009e-01 $+$       & 6.784e-01 $+$       & 6.707e-01 $+$       & 6.364e-01 $+$       & 6.563e-01 $+$       & 7.023e-01 $\approx$ & 7.048e-01 $\approx$ & 6.544e-01 $+$       & 7.100e-01  \\
        &      & (1.82e-02)          & (4.18e-02)          & (1.70e-02)          & (3.26e-02)          & (1.50e-02)          & (1.55e-02)          & (1.81e-02)          & (1.91e-02)          & (1.89e-02)          & (1.47e-02)          & (2.14e-02)          & (1.58e-02) \\\hline
\multirow{2}{*}{MaF14} & \multirow{2}{*}{3}  & 7.134e-03 $\approx$ & 8.921e-04 $\approx$ & 1.274e-02 $\approx$ & 5.213e-03 $\approx$ & 7.648e-02 $\approx$ & 3.417e-02 $\approx$ & 1.499e-02 $\approx$ & 1.251e-02 $\approx$ & 2.041e-02 $\approx$ & 5.029e-03 $\approx$ & 9.238e-02 $\approx$ & 3.086e-02  \\
        &      & (2.24e-02)          & (3.90e-03)          & (3.27e-02)          & (1.31e-02)          & (9.46e-02)          & (3.98e-02)          & (4.16e-02)          & (3.63e-02)          & (5.39e-02)          & (1.57e-02)          & (8.15e-02)          & (5.14e-02) \\\hline
\multirow{2}{*}{MaF15} & \multirow{2}{*}{3}  & 2.352e-02 $+$       & 2.729e-02 $+$       & 3.762e-01 $\approx$ & 6.709e-03 $+$       & 4.106e-01 $-$       & 3.778e-01 $\approx$ & 2.831e-02 $+$       & 1.379e-01 $\approx$ & 4.539e-01 $-$       & 4.106e-01 $-$       & 2.516e-01 $\approx$ & 2.391e-01  \\
        &      & (4.16e-02)          & (3.47e-02)          & (3.31e-02)          & (6.49e-03)          & (4.68e-02)          & (2.88e-02)          & (1.54e-02)          & (2.85e-02)          & (2.49e-02)          & (1.48e-02)          & (5.25e-02)          & (8.66e-02)  \\ \hline

\multicolumn{2}{c}{$+$/$\approx$/$-$}                   & 11/4/0  & 13/2/0   & 10/4/1 &  11/4/0   &  12/2/1    &  11/3/1 & 9/6/0  & 9/6/0    & 1/13/1   & 11/2/2  & 5/10/0    &    
\\\bottomrule 
    \end{tabular}%
    }
    \end{center}
    \vspace{6pt}
\footnotesize `$+$', `$\approx$', and `$-$' indicate that the result of E3A is significantly better than, equivalent to, and significantly worse than that of the peer algorithm at a $0.05$ level by the statistical tests (i.e., Friedman test and posthoc Nemenyi test), respectively.
\end{table}

\begin{figure}[tbp]
	\begin{center}
		\footnotesize
		\begin{tabular}{@{}cccc}
			\includegraphics[scale=0.25]{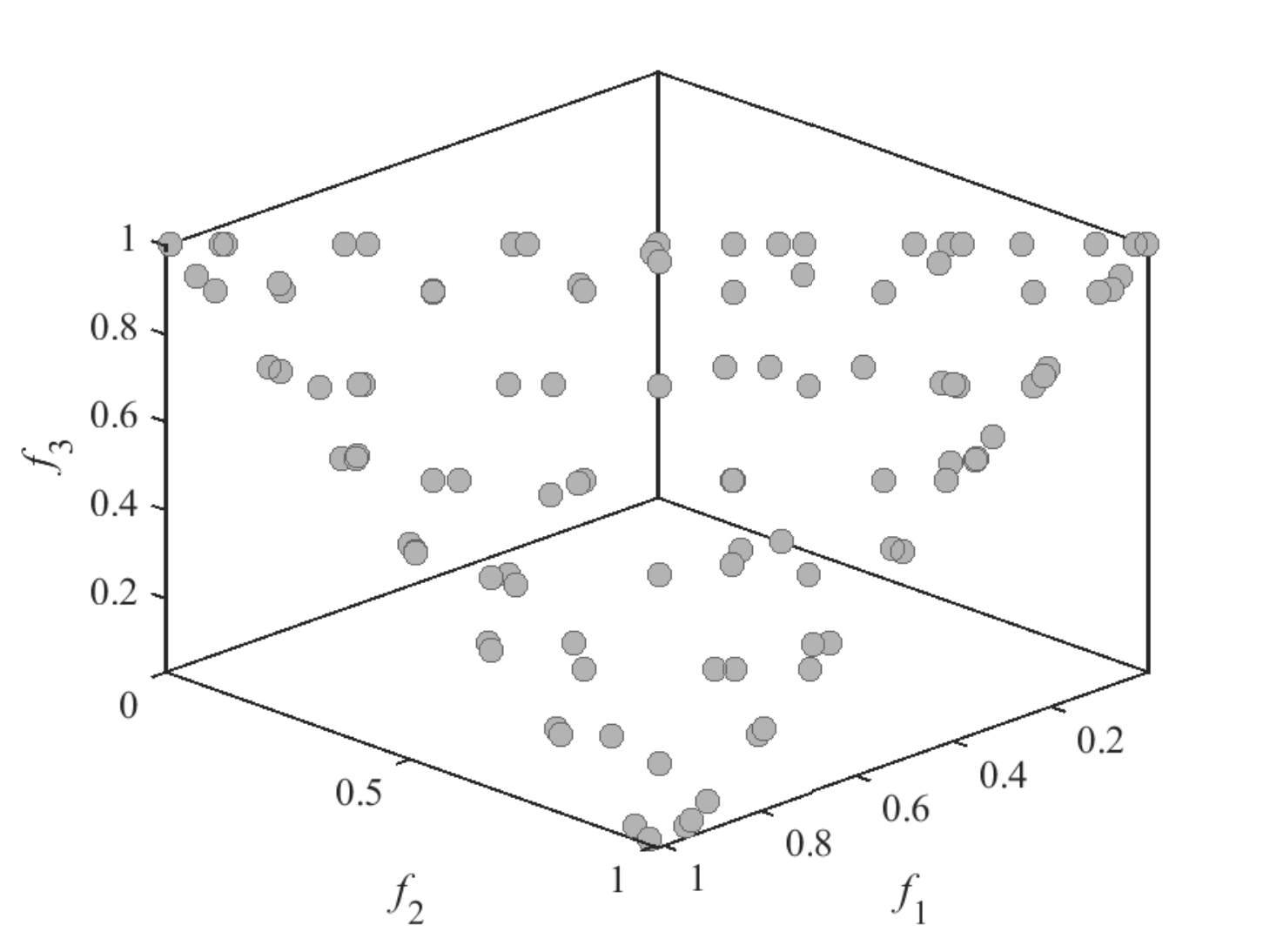}&
			\includegraphics[scale=0.25]{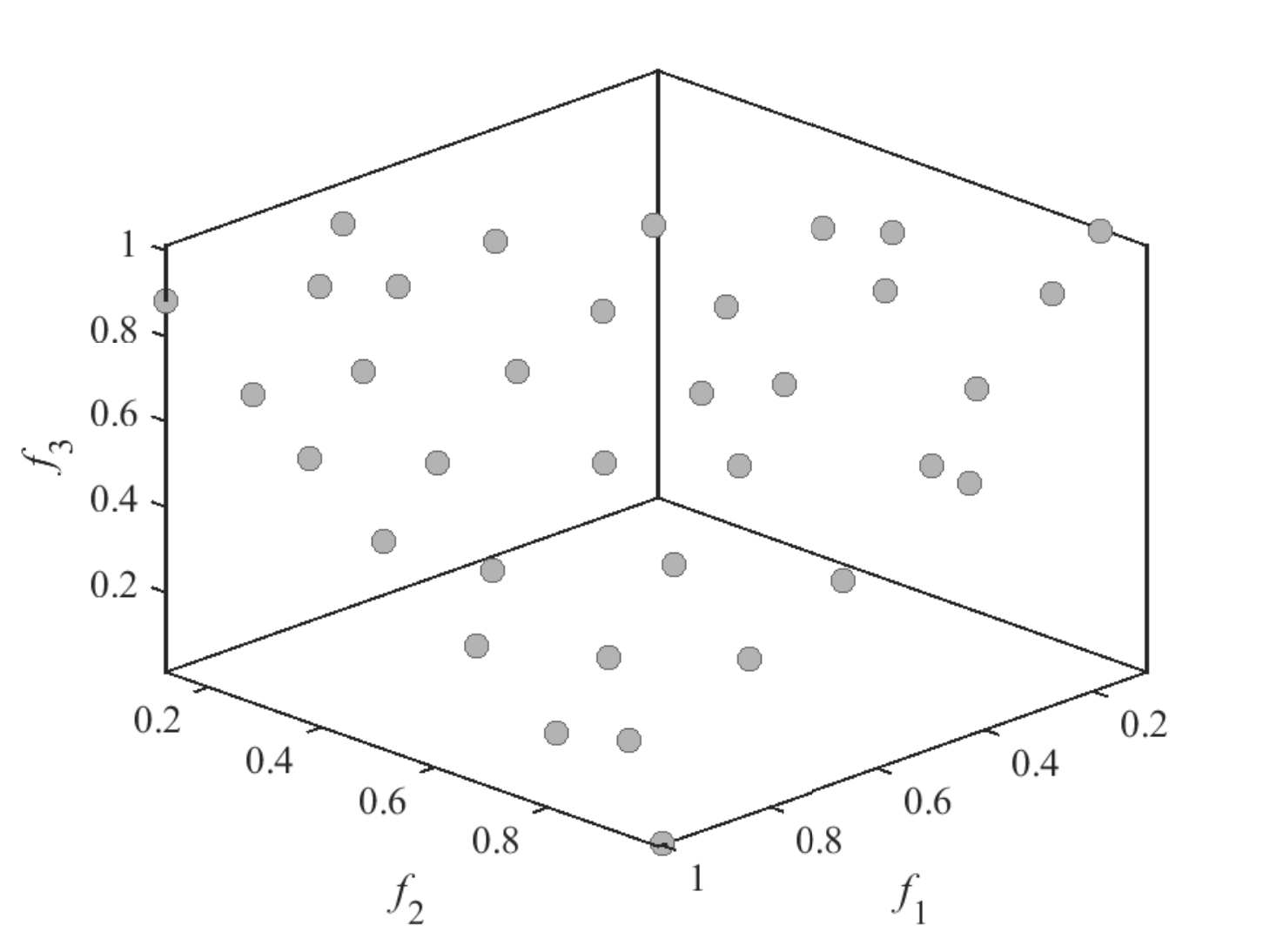}&
			\includegraphics[scale=0.25]{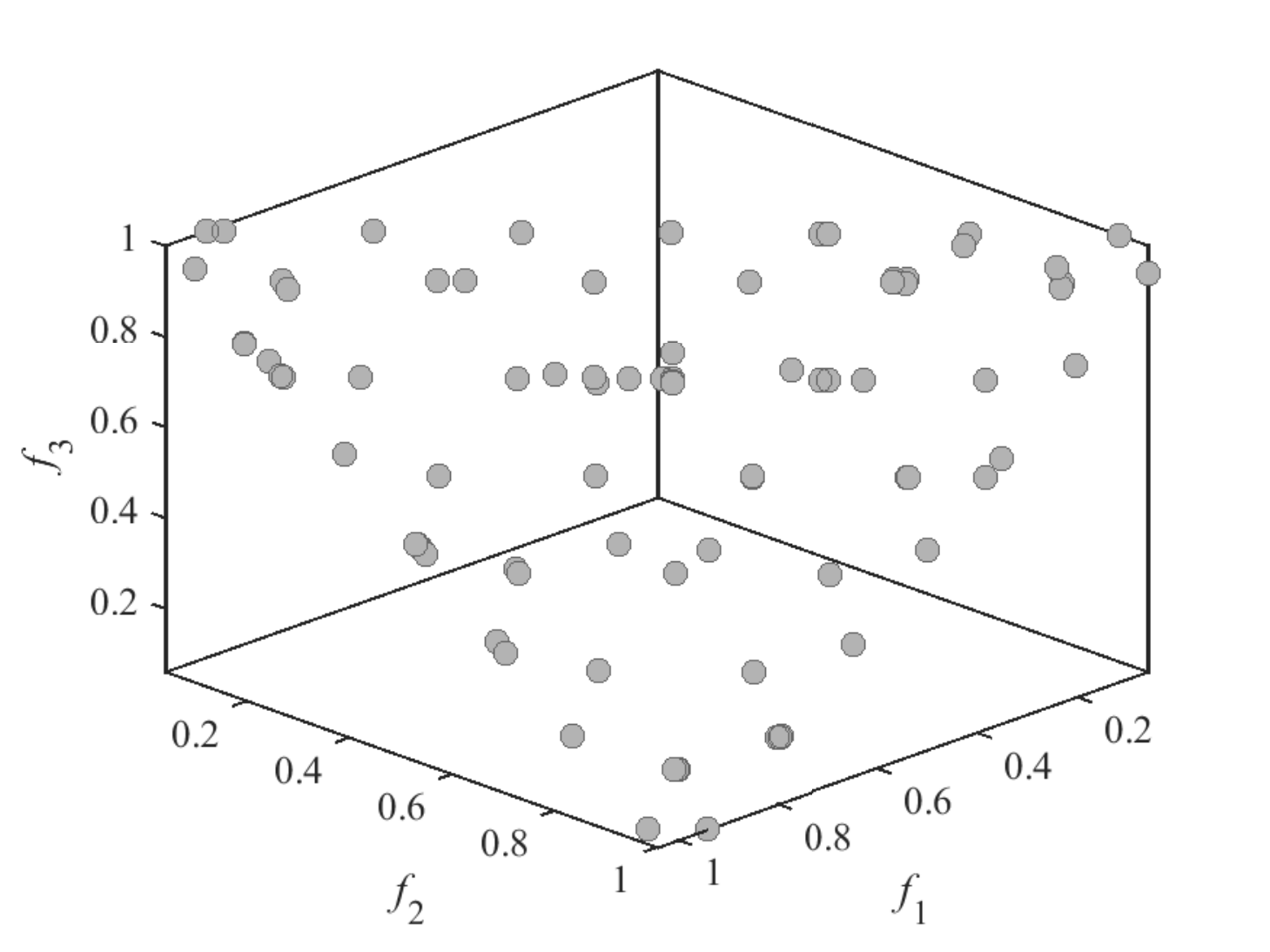}&
			\includegraphics[scale=0.25]{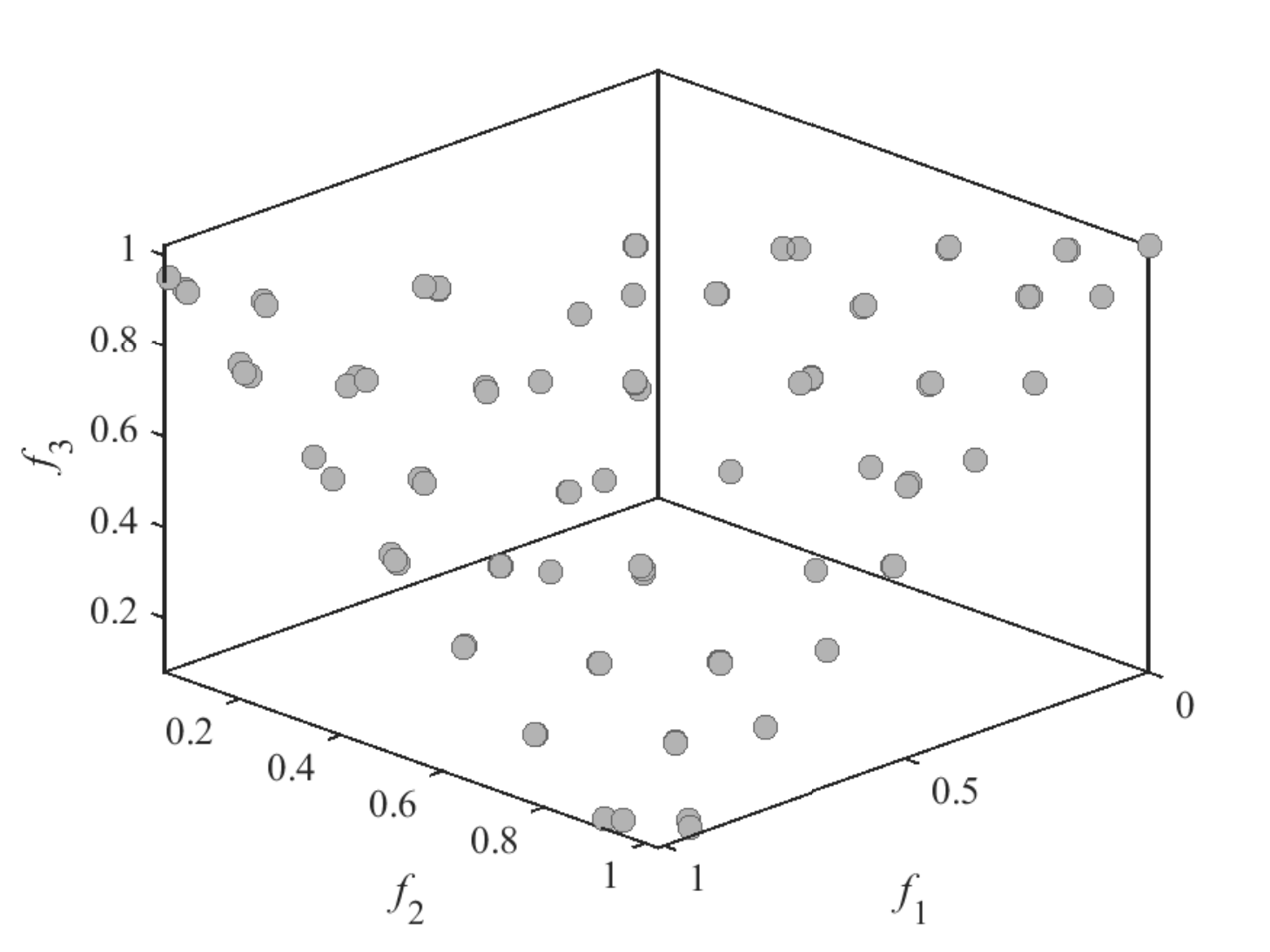}\\
			(a) NSGA-III  & (b) RVEA  & (c) MOEA/DD &  (d) RPD-NSGA-II  \\
		\end{tabular}
		\begin{tabular}{@{}cccc}
			\includegraphics[scale=0.25]{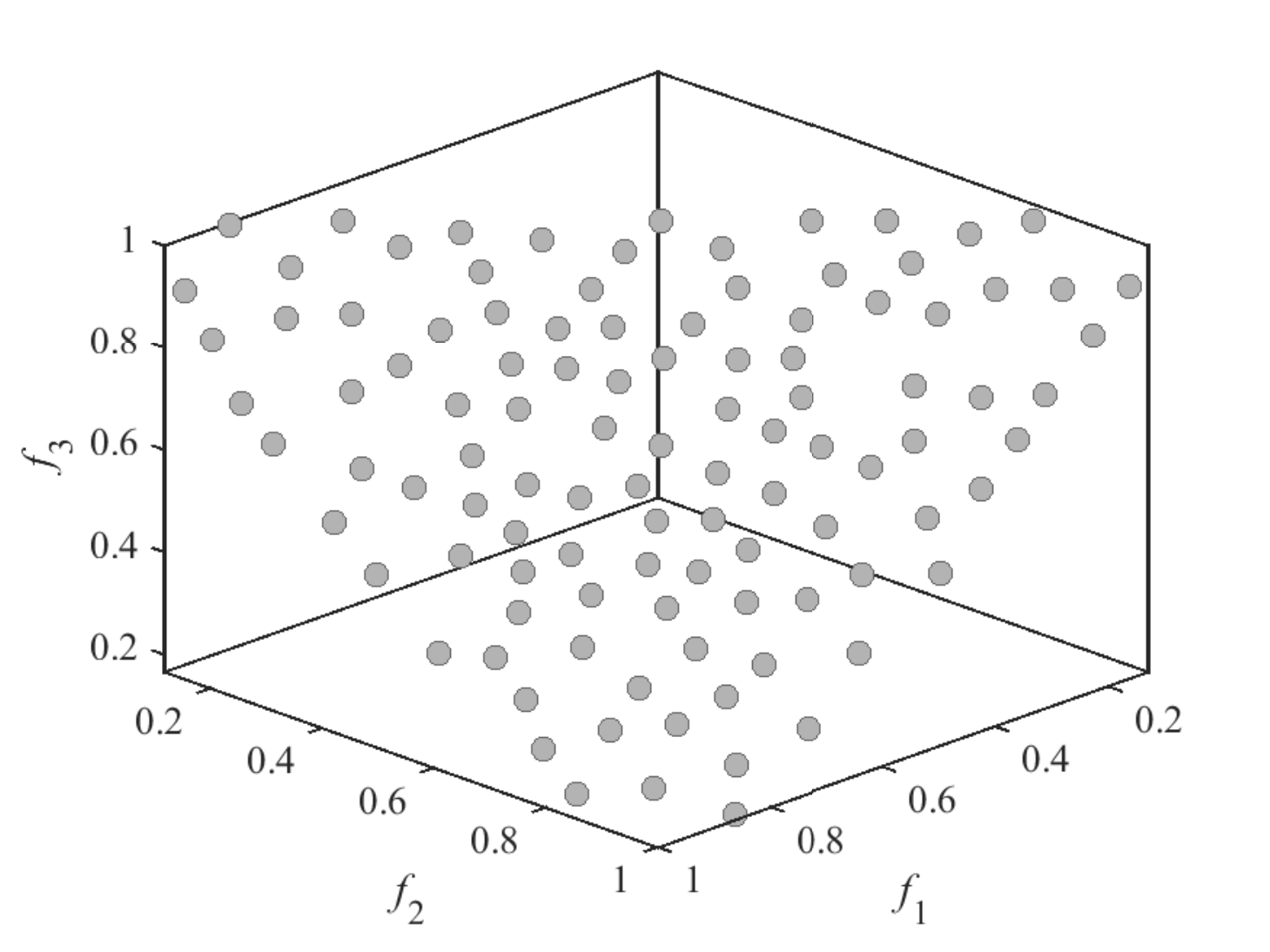}&
			\includegraphics[scale=0.25]{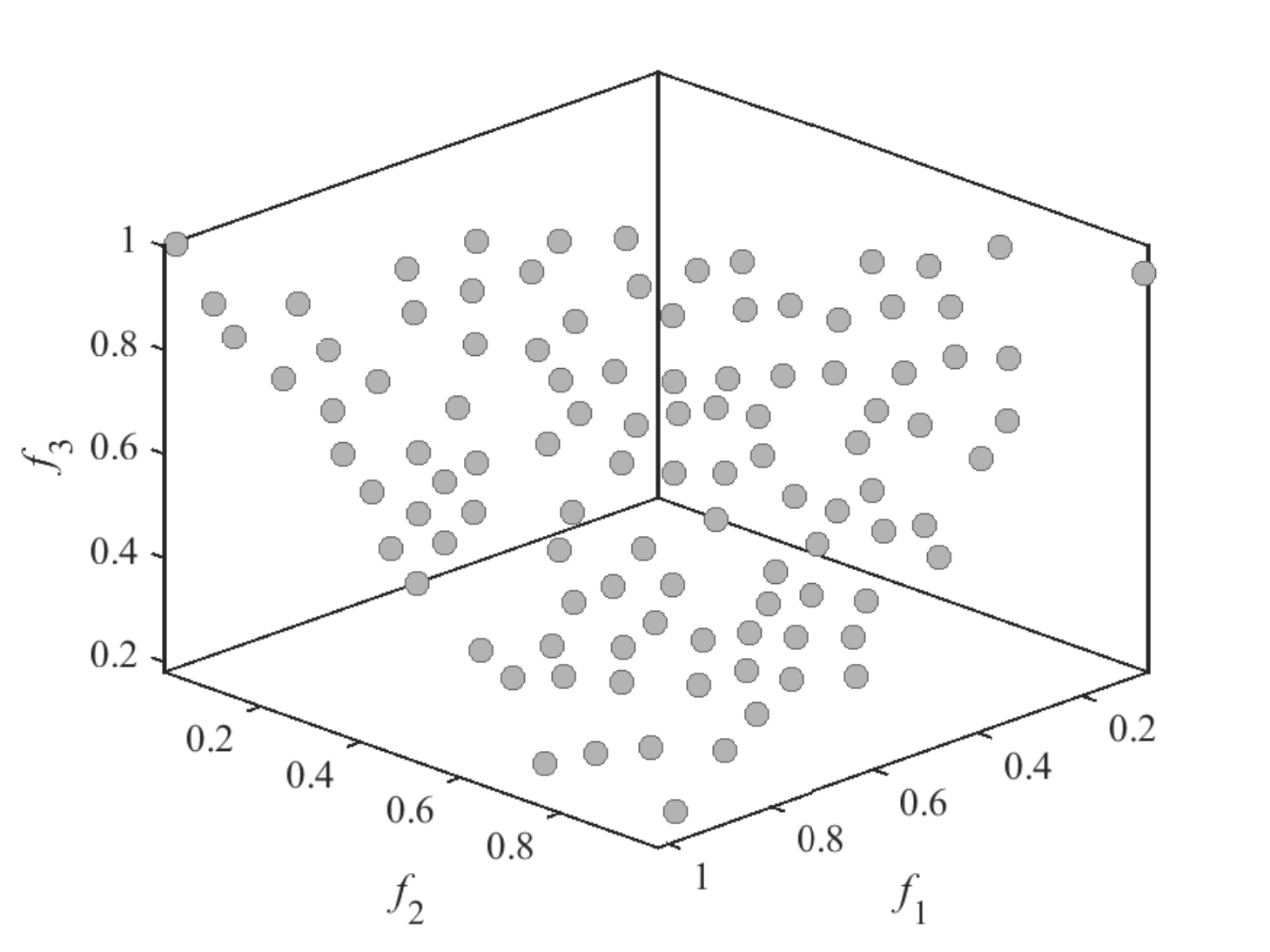}&
			\includegraphics[scale=0.25]{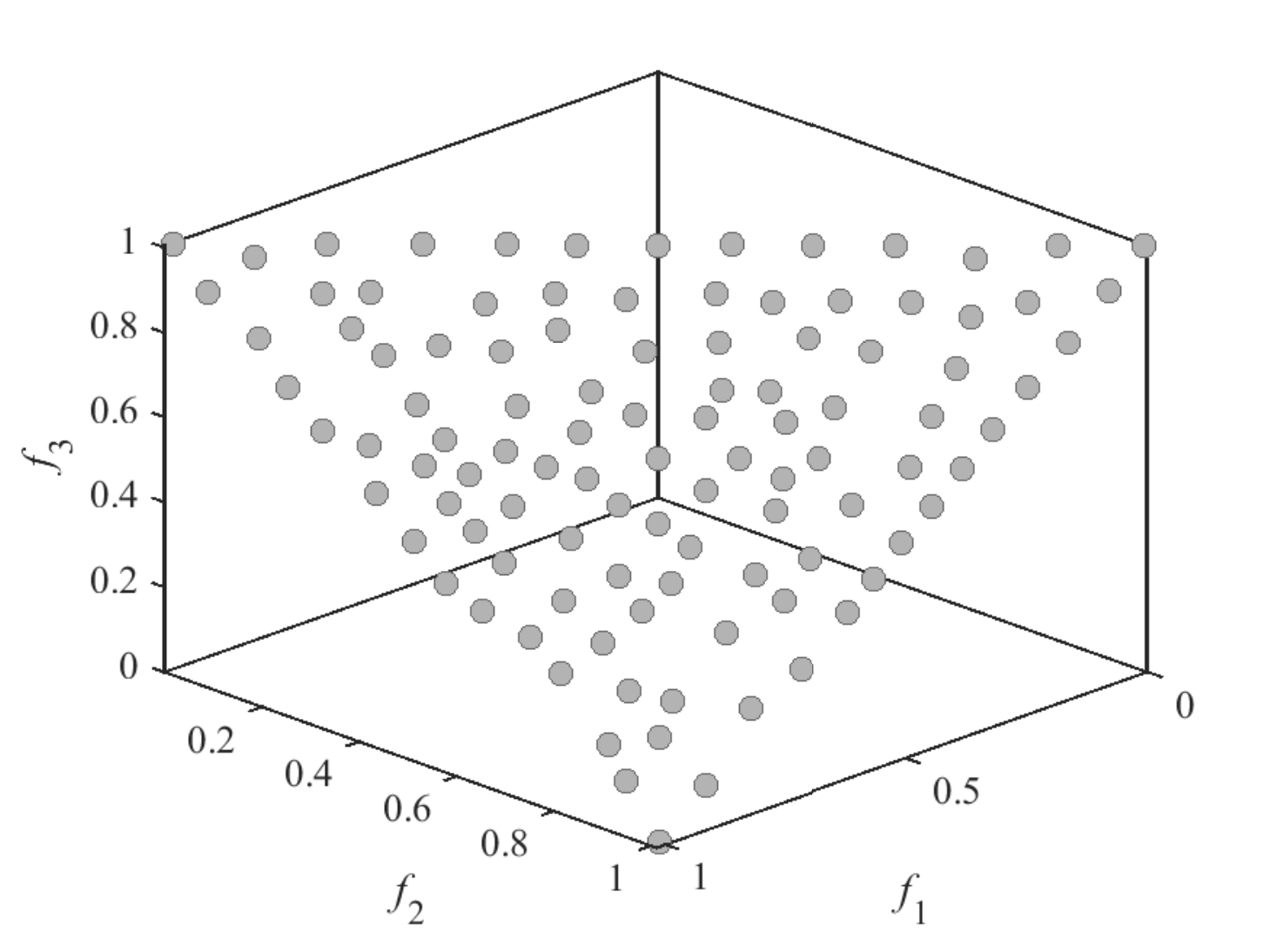}&
			\includegraphics[scale=0.25]{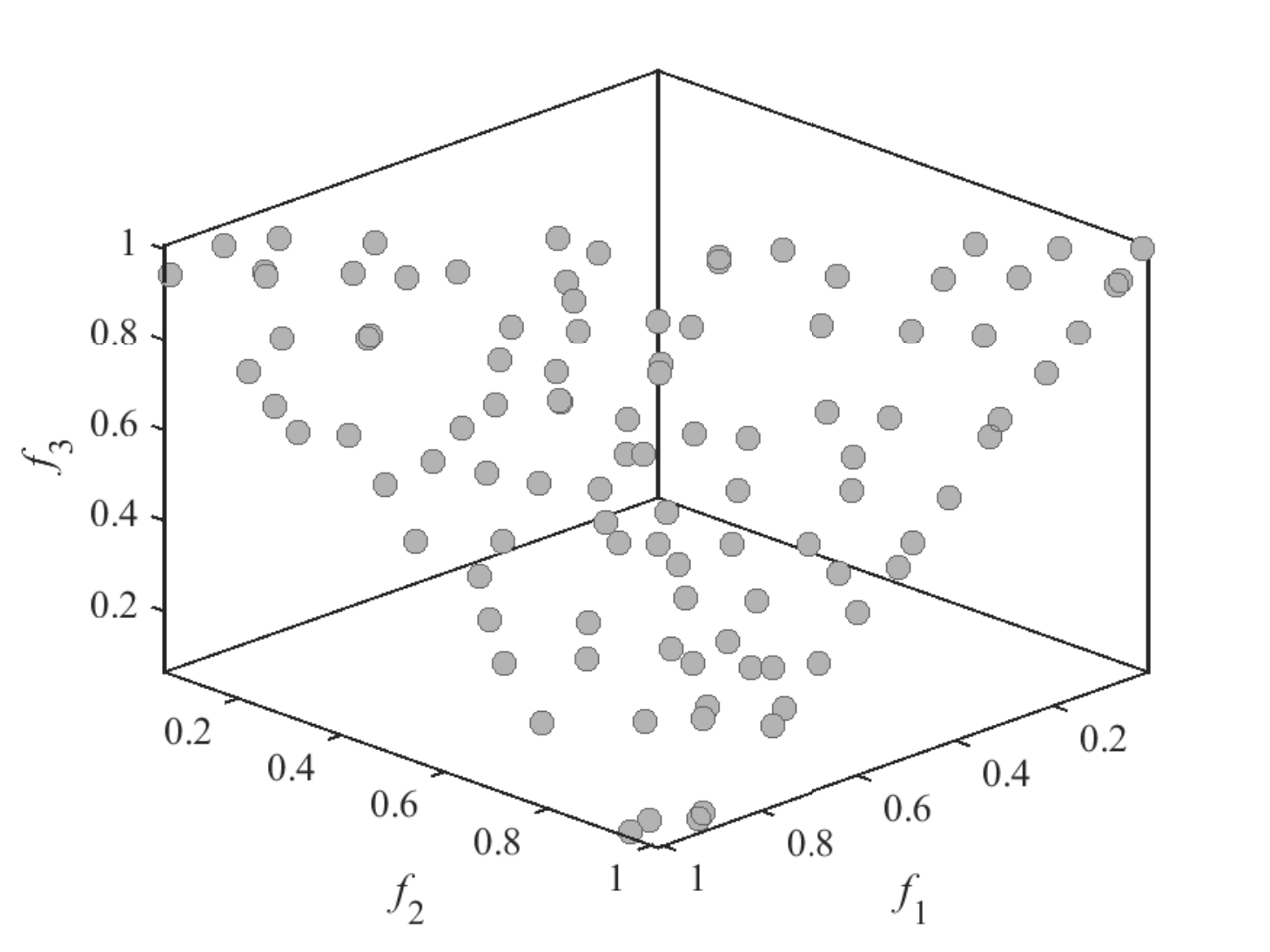}\\
			(e) 1by1EA  & (f) MaOEA-CSS &  (g) VaEA  & (h) NSGA-II/SDR  \\
		\end{tabular}
		\begin{tabular}{@{}cccc}
			\includegraphics[scale=0.25]{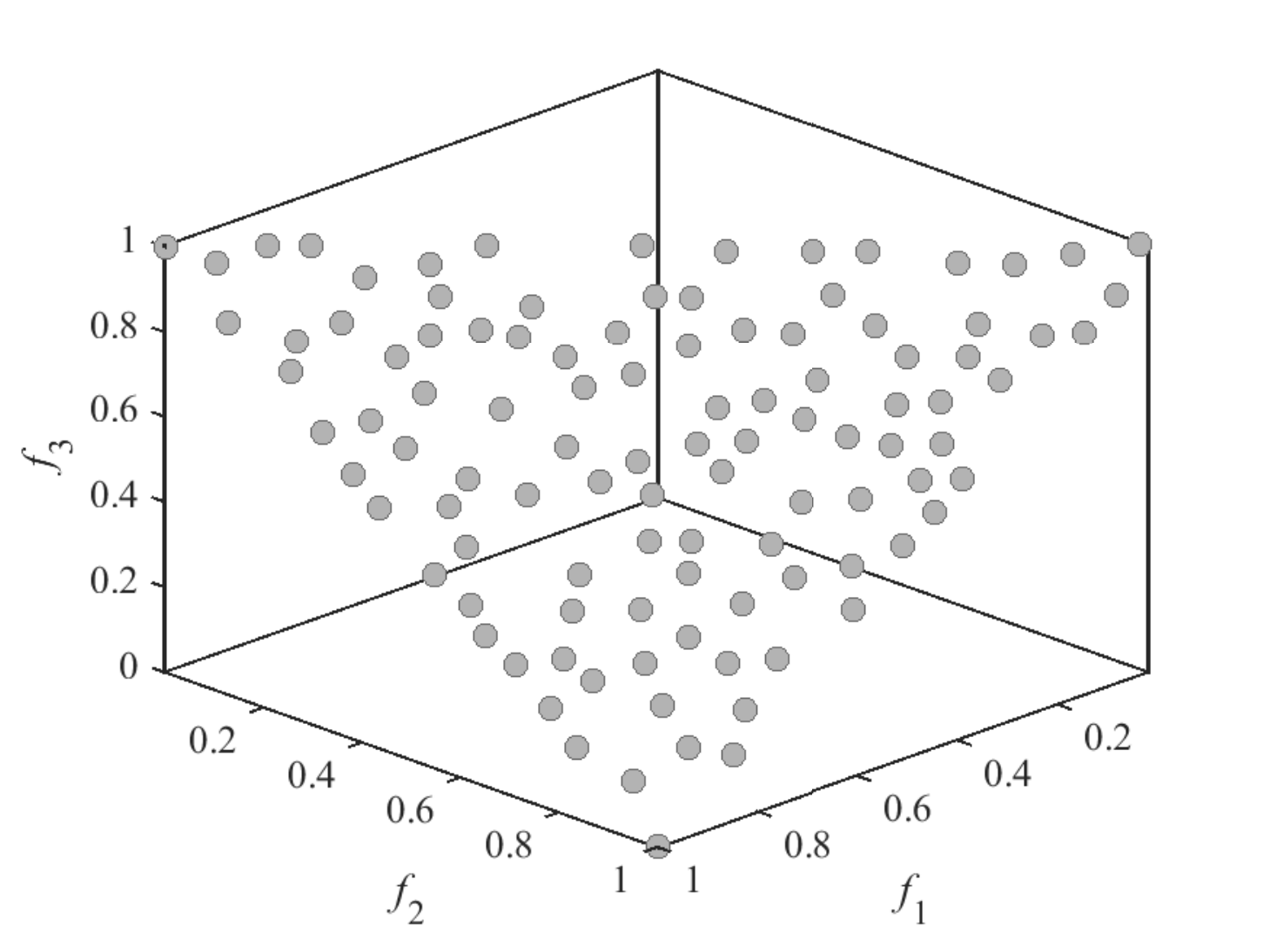}&
			\includegraphics[scale=0.25]{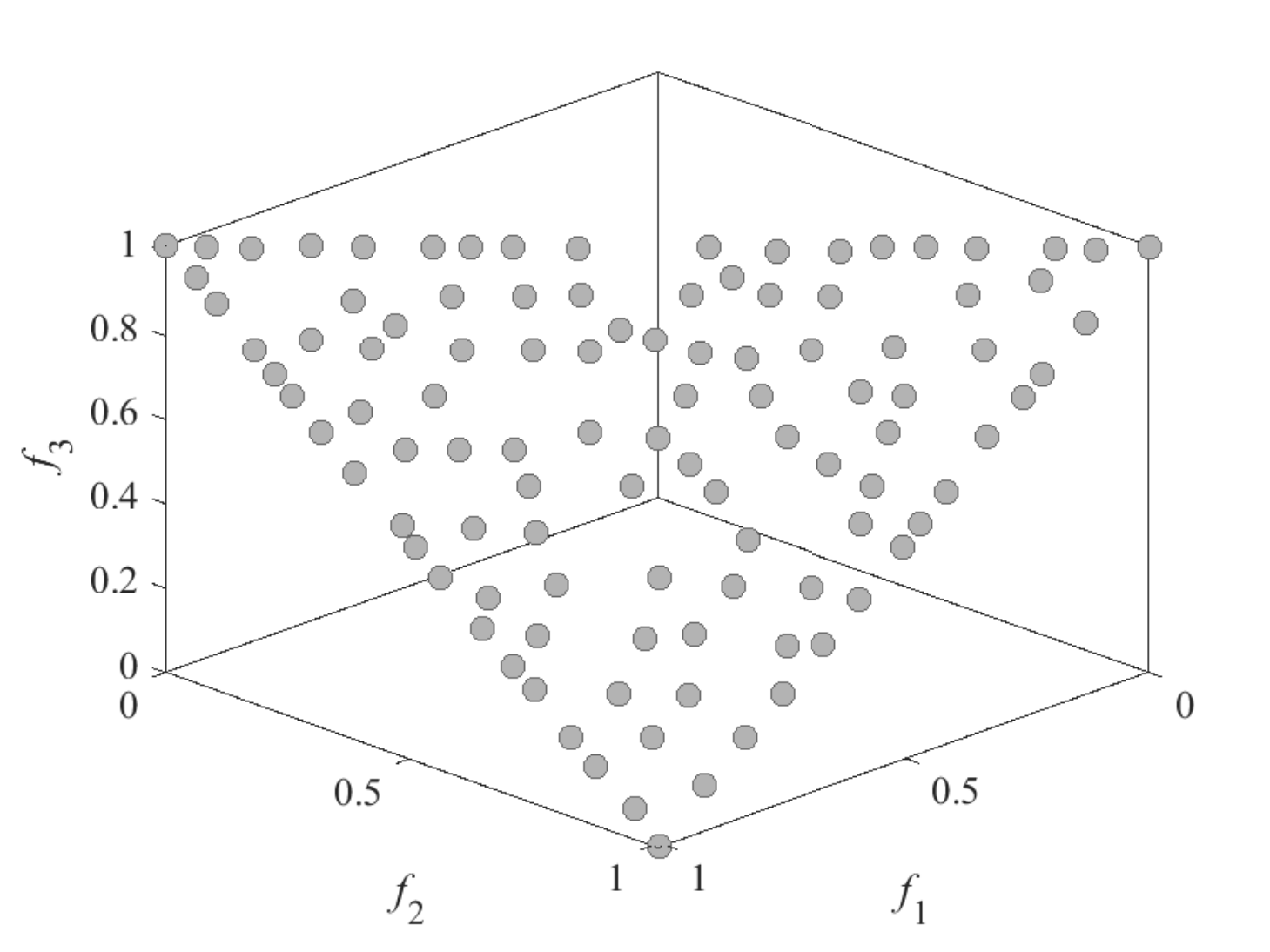}&
			\includegraphics[scale=0.25]{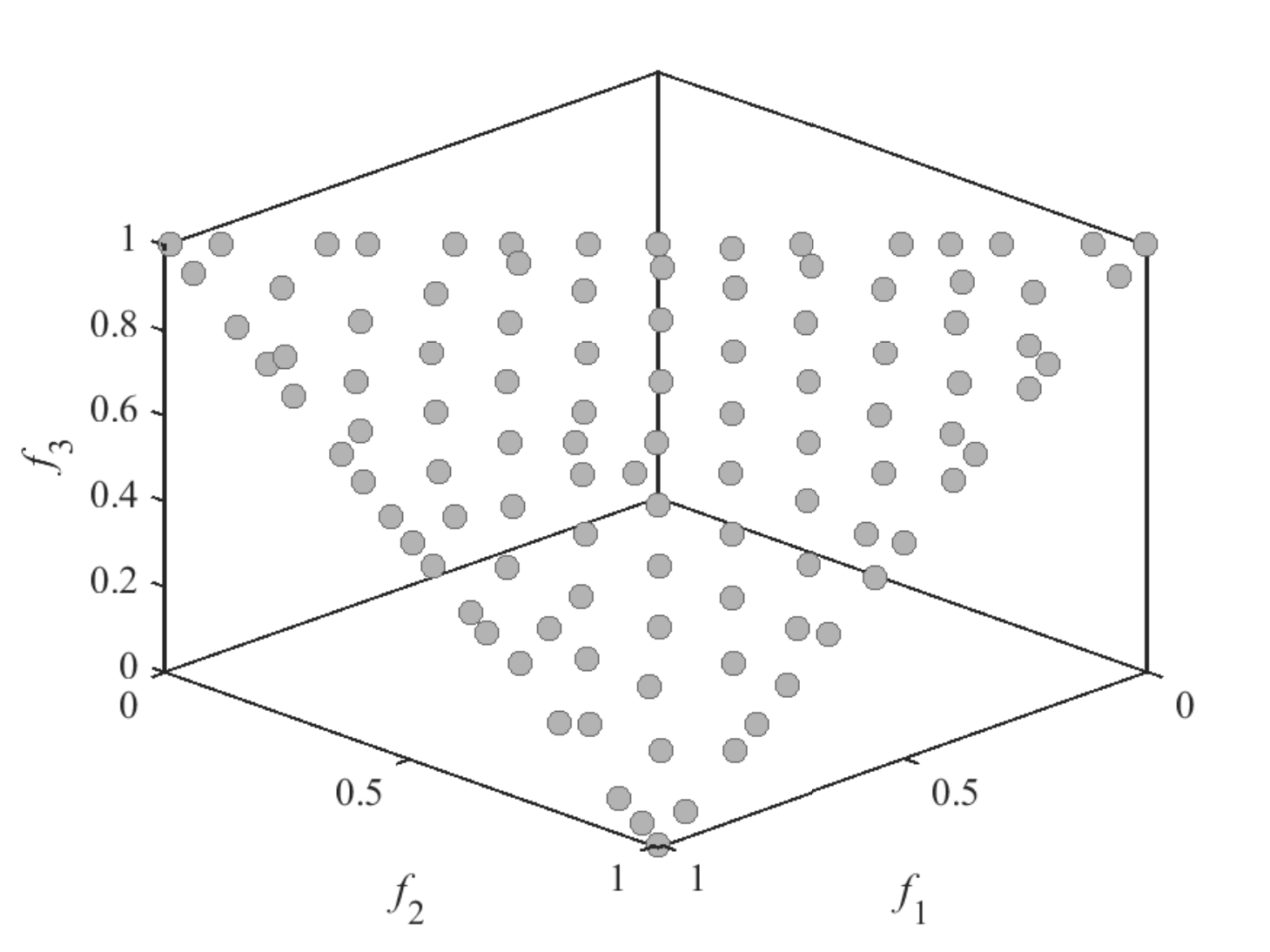}&
			\includegraphics[scale=0.25]{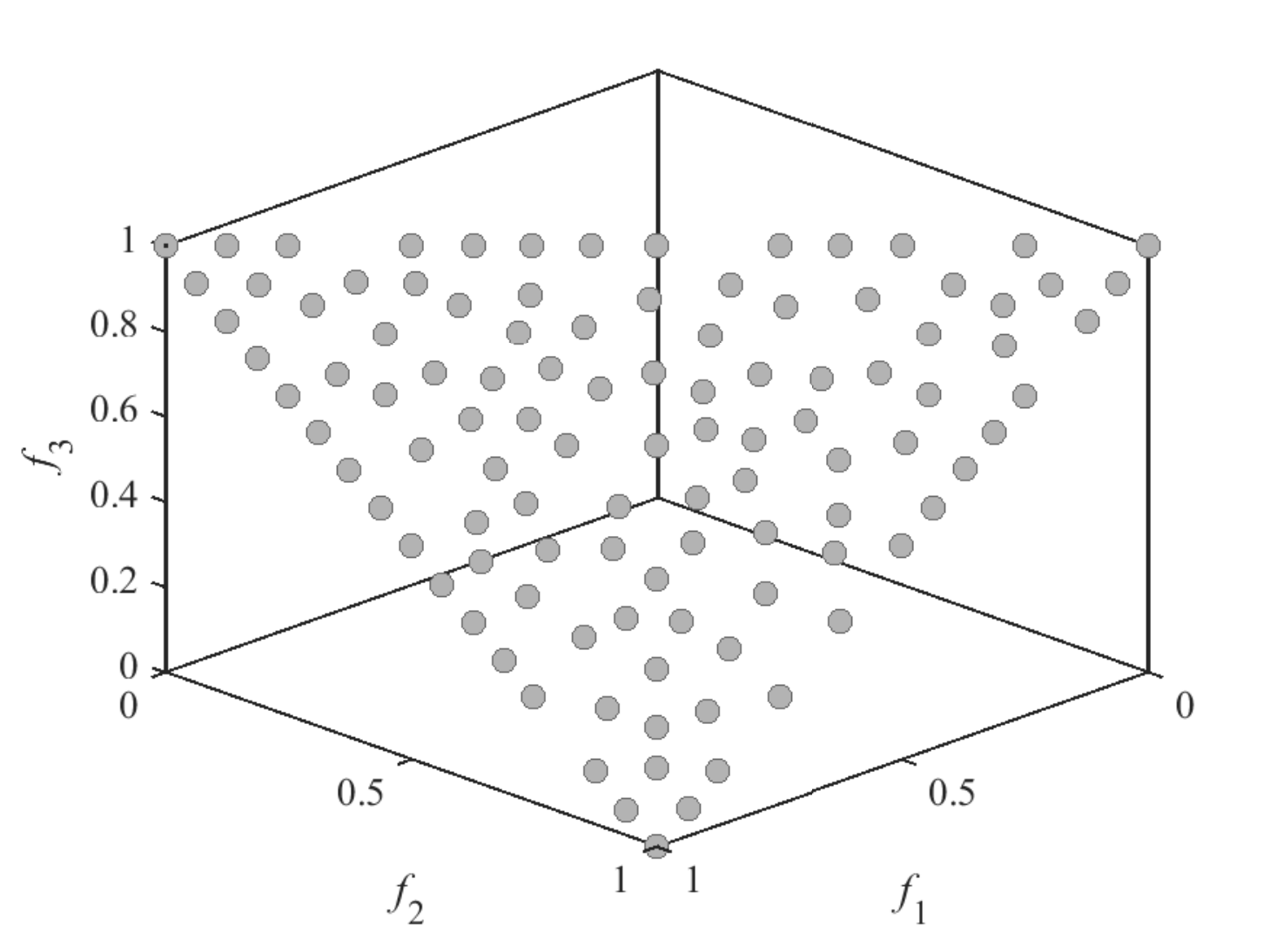}\\
			 (i) SPEA2+SDE &  (j) SRA & (k) AR-MOEA  & (l) E3A  \\
		\end{tabular}
	\end{center}
	\caption{The final solution set with the median IGD obtained by the $12$ algorithms on the tri-objective MaF1.}
	\label{Fig:MaF1-3OBJ}
\end{figure}

For a visual understanding of how solutions are distributed,
Figure~\ref{Fig:MaF1-3OBJ} plots the final solution set obtained by each algorithm on the tri-objective MaF1 with the median IGD value among all the $30$ runs. As seen in Figure~\ref{Fig:MaF1-3OBJ}, VaEA, SRA, AR-MOEA, and E3A are the only four algorithms that perform well in terms of both convergence and diversity.
Among them, E3A appears to be the best, with its solutions uniformly distributed over the Pareto front. For the other eight algorithms, maintaining a set of diverse solutions seems challenging, especially for four decomposition-based algorithms, NSGA-III, RVEA, MOEA/DD, and RPD-NSGA-II. The main reason is that decomposition-based algorithms, which work well for MaOPs with regular Pareto front shapes, typically struggle for irregular shapes, which is the case in this instance having an inverted hyperplane.
It is worth mentioning that RVEA only keeps one solution to a weight vector, leading to an evenly distributed solution set but with fewer solutions remaining (only $35$ out of $105$ solutions).
It can be observed from Table~\ref{Table:3obj_MaF_IGD} that the four algorithms (NSGA-III, RVEA, MOEA/DD, and RPD-NSGA-II) obtain poor results regarding the IGD metric with RVEA performing the worst.

\subsubsection{Algorithm Performance on Five-Objective Problems}
Tables \ref{Table:5obj_MaF_IGD} and \ref{Table:5obj_MaF_HV} show the mean and standard deviation (in parentheses) of the IGD and HV results obtained by E3A and the peer algorithms on the five-objective MaF problems. From Table \ref{Table:5obj_MaF_IGD}, it can be seen that E3A obtains the best mean, in terms of IGD, on four test problems, MaF$4$, MaF$8$, MaF$10$, and MaF$13$. 
Concerning pairwise comparison, the proportions of the test problems where E3A significantly outperforms NSGA-III, RVEA, MOEA/DD, RPD-NSGA-II, 1by1EA, MaOEA-CSS, VaEA, NSGA-II/SDR, SPEA2+SDE, SRA, and AR-MOEA are $10/15$, $11/15$, $12/15$, $9/15$, $9/15$, $10/15$, $6/15$, $5/15$, $3/15$, $7/15$, and $2/15$, respectively. Conversely, the proportions of the test problems where E3A performs significantly worse than these peer algorithms are $1/15$, $2/15$, $1/15$, $2/15$, $3/15$, $3/15$, $1/15$, $0/15$, $1/15$, $1/15$, and $2/15$, respectively. 

From Table \ref{Table:5obj_MaF_HV}, it can be seen that E3A obtains the best HV results on six test problems (i.e., MaF$2$, MaF$4$, MaF$8$, MaF$10$, MaF$11$, and MaF$13$). Concerning pairwise comparison, 
the proportions of the test problems where E3A significantly outperforms NSGA-III, RVEA, MOEA/DD, RPD-NSGA-II, 1by1EA, MaOEA-CSS, VaEA, NSGA-II/SDR, SPEA2+SDE, SRA, and AR-MOEA are $10/15$, $11/15$, $13/15$, $7/15$, $10/15$, $11/15$, $9/15$, $6/15$, $1/15$, $5/15$, and $6/15$, respectively. Conversely, the proportions of the test problems where E3A performs significantly worse than these peer algorithms are $0/15$, $1/15$, $1/15$, $0/15$, $2/15$, $1/15$, $1/15$, $1/15$, $1/15$, $1/15$, and $1/15$, respectively.

\begin{table}[tbp]
	\scriptsize
    \begin{center}
    \caption{Mean and standard deviation of the IGD values obtained by the $12$ algorithms on the five-objective MaF problems.}
    \label{Table:5obj_MaF_IGD}
    \resizebox{\textwidth}{!}{%
    \renewcommand{\arraystretch}{1.2} 
    \begin{tabular}{@{}cccccccccccccc@{}}
 \toprule
Problem                & Obj.               & NSGA-III                      & RVEA        & MOEA/DD      & RPD-NSGA-II   & 1by1EA      & MaOEA-CSS    & VaEA        & NSGA-II/SDR   & SPEA2+SDE    & SRA         & AR-MOEA      & E3A      \\
  \midrule
\multirow{2}{*}{MaF1}  & \multirow{2}{*}{5} & 2.507e-1  $+$ & 3.141e-1  $+$ & 3.992e-1  $+$ & 2.108e-1  $+$ & 1.197e-1  $-$ & 1.201e-1  $-$ & 1.268e-1  $\approx$ & 1.230e-1  $\approx$ & 1.253e-1  $\approx$ & 1.387e-1  $\approx$ & 1.355e-1  $\approx$ & 1.297e-1  \\
&                    & (8.49e-3)    & (3.84e-2)    & (1.38e-1)    & (8.58e-3)    & (6.55e-3)    & (2.44e-3)    & (1.16e-3)     & (1.70e-3)    & (1.34e-3)    & (2.57e-3)    & (1.20e-3)    & (1.35e-3)   \\\hline

\multirow{2}{*}{MaF2}  & \multirow{2}{*}{5} & 1.284e-1  $+$ & 1.263e-1  $+$ & 1.605e-1  $+$ & 1.290e-1  $+$ & 9.688e-2  $\approx$ & 9.207e-2  $-$ & 1.058e-1  $\approx$  & 1.139e-1  $\approx$ & 1.158e-1  $\approx$ & 1.181e-1  $+$ & 1.115e-1  $\approx$ & 1.101e-1  \\
&                    & (2.69e-3)    & (2.06e-3)    & (5.54e-3)    & (1.02e-3)    & (2.72e-3)    & (1.46e-3)    & (1.77e-3)     & (2.53e-3)    & (1.64e-3)    & (3.11e-3)    & (1.71e-3)    & (1.63e-3)   \\\hline

\multirow{2}{*}{MaF3}  & \multirow{2}{*}{5} & 8.795e+0  $+$ & 4.539e+1  $+$ & 2.212e+1  $+$ & 1.023e+0  $+$ & 6.951e-1  $+$ & 2.367e+0  $+$ & 5.937e+1  $+$ & 1.520e-1  $\approx$ & 1.913e-1  $\approx$ & 1.011e+0  $\approx$ & 4.182e-1  $\approx$ & 2.763e-1  \\
&                    & (1.33e+1)    & (4.54e+1)    & (2.46e+1)    & (1.35e+0)    & (1.05e+0)    & (3.25e+0)    & (8.34e+1)     & (1.27e-2)    & (3.61e-1)    & (4.33e+0)    & (7.19e-1)    & (5.61e-1)   \\\hline

\multirow{2}{*}{MaF4}  & \multirow{2}{*}{5} & 1.272e+1  $+$ & 7.091e+0  $+$ & 6.246e+0  $+$ & 4.493e+0  $+$ & 8.078e+0  $+$ & 5.019e+0  $+$ & 6.922e+0  $+$ & 3.685e+0  $\approx$ & 3.685e+0  $+$ & 4.530e+0  $+$ & 3.095e+0  $\approx$ & 2.630e+0  \\
&                    & (1.36e+1)    & (5.77e+0)    & (3.16e+0)    & (4.46e+0)    & (6.68e-1)    & (2.03e+0)    & (7.75e+0)     & (2.30e+0)    & (1.58e+0)    & (4.28e+0)    & (1.62e+0)    & (1.46e+0)   \\\hline

\multirow{2}{*}{MaF5}  & \multirow{2}{*}{5} & 2.552e+0  $+$ & 2.589e+0  $+$ & 5.037e+0  $+$ & 2.378e+0  $\approx$ & 4.591e+0  $+$ & 4.639e+0  $+$ & 2.102e+0  $\approx$  & 1.350e+1  $+$ & 2.684e+0  $+$ & 2.647e+0  $+$ & 2.522e+0  $+$ & 2.227e+0  \\
&                    & (6.64e-1)    & (4.09e-1)    & (6.75e-1)    & (5.07e-2)    & (6.37e-1)    & (7.47e-1)    & (4.03e-2)     & (2.27e+0)    & (9.26e-1)    & (9.09e-1)    & (6.27e-1)    & (3.19e-1)   \\\hline

\multirow{2}{*}{MaF6}  & \multirow{2}{*}{5} & 4.790e-2  $\approx$  & 1.491e-1  $+$ & 7.374e-2  $+$ & 1.490e-1  $+$ & 3.568e-3  $-$ & 6.069e-2  $+$ & 4.223e-3  $\approx$  & 1.747e-2  $\approx$ & 8.101e-3  $\approx$ & 7.020e-3  $\approx$ & 3.513e-3  $-$ & 7.839e-3  \\
&                    & (8.30e-3)    & (1.48e-1)    & (9.37e-3)    & (5.87e-2)    & (7.00e-5)    & (1.12e-2)    & (1.59e-4)     & (9.13e-3)    & (7.92e-4)    & (2.19e-3)    & (7.21e-5)    & (1.32e-4)   \\\hline

\multirow{2}{*}{MaF7}  & \multirow{2}{*}{5} & 3.398e-1  $\approx$ & 5.921e-1  $+$ & 3.001e+0  $+$ & 3.730e-1  $-$ & 4.086e-1  $+$ & 3.945e-1  $+$ & 3.408e-1  $\approx$ & 3.823e-1  $+$ & 3.070e-1  $\approx$ & 3.002e-1  $\approx$ & 3.259e-1  $\approx$ & 3.733e-1  \\
&                    & (2.21e-2)    & (4.14e-2)    & (1.56e-6)    & (1.17e-2)    & (4.10e-2)    & (2.42e-2)    & (1.73e-2)     & (3.63e-2)    & (4.34e-2)    & (4.47e-2)    & (9.97e-3)    & (1.87e-1)   \\\hline

\multirow{2}{*}{MaF8}  & \multirow{2}{*}{5} & 2.034e-1  $+$ & 4.033e-1  $+$ & 3.342e-1  $+$ & 3.099e-1  $+$ & 5.191e-1  $+$ & 1.137e-1  $\approx$ & 1.159e-1  $\approx$  & 1.419e-1  $+$ & 1.184e-1  $\approx$ & 5.193e-1  $+$ & 1.250e-1  $\approx$ & 1.064e-1  \\
&                    & (3.26e-2)    & (3.25e-2)    & (7.99e-2)    & (2.98e-2)    & (5.87e-2)    & (8.77e-3)    & (1.13e-2)     & (1.67e-2)    & (1.98e-2)    & (9.76e-1)    & (2.59e-2)    & (1.35e-2)   \\\hline

\multirow{2}{*}{MaF9}  & \multirow{2}{*}{5} & 5.181e-1  $+$ & 3.556e-1  $+$ & 2.669e-1  $+$ & 3.250e-1  $+$ & 2.717e-1  $+$ & 2.019e-1  $+$ & 5.023e-1  $+$  & 1.702e-1  $\approx$ & 9.954e-2  $\approx$ & 1.122e-1  $\approx$ & 1.250e-1  $\approx$ & 1.206e-1  \\
&                    & (2.12e-1)    & (4.95e-2)    & (5.64e-2)    & (6.31e-2)    & (4.93e-2)    & (1.92e-2)    & (2.66e-1)     & (1.53e-2)    & (1.25e-2)    & (1.94e-2)    & (1.94e-2)    & (2.02e-2)   \\\hline

\multirow{2}{*}{MaF10} & \multirow{2}{*}{5} & 9.263e-1  $+$ & 7.799e-1  $+$ & 1.440e+0  $+$ & 5.898e-1  $\approx$ & 9.473e-1  $+$ & 8.581e-1  $+$ & 1.218e+0  $+$ & 8.074e-1  $+$ & 5.225e-1  $\approx$ & 1.213e+0  $+$ & 8.091e-1  $+$ & 4.704e-1  \\
&                    & (9.27e-2)    & (9.08e-2)    & (1.37e-1)    & (5.77e-2)    & (6.61e-2)    & (6.31e-2)    & (1.15e-1)     & (1.05e-1)    & (2.75e-2)    & (1.26e-1)    & (7.40e-2)    & (2.46e-2)   \\\hline

\multirow{2}{*}{MaF11} & \multirow{2}{*}{5} & 4.650e-1  $-$ & 4.601e-1  $-$ & 5.433e-1  $\approx$ & 4.549e-1  $-$ & 7.613e-1  $+$ & 6.662e-1  $+$ & 4.635e-1  $-$ & 5.723e-1  $\approx$ & 5.688e-1  $\approx$ & 5.498e-1  $\approx$ & 4.718e-1  $\approx$ & 5.323e-1  \\
&                    & (3.84e-3)    & (1.93e-2)    & (1.35e-2)    & (5.62e-3)    & (6.16e-2)    & (4.50e-2)    & (8.76e-3)     & (6.98e-2)    & (2.08e-2)    & (2.15e-2)    & (4.02e-3)    & (1.50e-2)   \\\hline

\multirow{2}{*}{MaF12} & \multirow{2}{*}{5} & 1.114e+0  $\approx$ & 1.136e+0  $\approx$ & 1.207e+0  $+$ & 1.152e+0  $\approx$ & 1.614e+0  $+$ & 1.477e+0  $+$ & 1.091e+0  $\approx$ & 1.150e+0  $\approx$ & 1.256e+0  $+$ & 1.196e+0  $+$ & 1.135e+0  $\approx$ & 1.135e+0  \\
&                    & (5.38e-3)    & (3.98e-3)    & (1.04e-2)    & (3.76e-2)    & (1.37e-1)    & (7.20e-2)    & (1.69e-2)     & (1.43e-2)    & (2.74e-2)    & (2.73e-2)    & (3.71e-3)    & (9.92e-3)   \\\hline

\multirow{2}{*}{MaF13} & \multirow{2}{*}{5} & 2.396e-1  $+$ & 4.700e-1  $+$ & 2.534e-1  $+$ & 4.185e-1  $+$ & 1.197e-1  $\approx$ & 1.506e-1  $+$ & 1.732e-1  $+$ & 1.821e-1  $+$ & 1.411e-1  $\approx$ & 1.445e-1  $\approx$ & 1.328e-1  $\approx$ & 1.195e-1  \\
&                    & (3.22e-2)    & (1.27e-1)    & (7.24e-2)    & (5.97e-2)    & (1.12e-2)    & (1.35e-2)    & (2.68e-2)     & (1.89e-2)    & (1.92e-2)    & (2.53e-2)    & (1.27e-2)    & (1.47e-2)   \\\hline

\multirow{2}{*}{MaF14} & \multirow{2}{*}{5} & 4.133e+0  $+$ & 1.669e+0  $\approx$ & 1.585e+0  $\approx$ & 6.702e+0  $+$ & 1.015e+0  $\approx$ & 1.003e+0  $\approx$ & 8.138e+0  $+$ & 9.462e-1  $\approx$ & 1.537e+0  $\approx$ & 2.210e+0  $+$ & 1.318e+0  $\approx$ & 1.223e+0  \\
&                    & (1.33e+0)    & (9.68e-1)    & (4.75e-1)    & (2.02e+0)    & (3.89e-1)    & (1.73e-1)    & (2.56e+0)     & (2.37e-1)    & (3.82e-1)    & (6.28e-1)    & (3.81e-1)    & (3.34e-1)   \\\hline

\multirow{2}{*}{MaF15} & \multirow{2}{*}{5} & 2.458e+0  $\approx$ & 6.161e-1  $-$ & 6.142e-1  $-$ & 1.473e+0  $\approx$ & 6.497e-1  $-$ & 5.510e-1  $-$ & 1.473e+0  $\approx$  & 8.568e-1 $\approx$ & 4.492e-1  $-$ & 5.260e-1  $-$ & 6.262e-1  $-$ & 1.540e+0  \\
&                    & (8.30e-1)    & (6.46e-2)    & (1.18e-1)    & (2.33e-1)    & (7.31e-2)    & (5.25e-2)    & (1.41e-1)     & (3.26e-2)    & (5.34e-2)    & (5.24e-2)    & (5.16e-2)    & (1.02e+0)    \\\hline
\multicolumn{2}{c}{$+$/$\approx$/$-$}                   & 10/4/1  & 11/2/2  & 12/2/1 & 9/4/2    & 9/3/3   & 10/2/3   & 6/8/1    & 5/10/0 & 3/11/1    & 7/7/1  & 2/11/2       &           
\\\bottomrule 
    \end{tabular}%
    }
    \end{center}
    \vspace{6pt}
\footnotesize `$+$', `$\approx$', and `$-$' indicate that the result of E3A is significantly better than, equivalent to, and significantly worse than that of the peer algorithm at a $0.05$ level by the statistical tests (i.e., Friedman test and posthoc Nemenyi test), respectively.
\end{table}

\begin{table}[tbp]
	\scriptsize
    \begin{center}
    \caption{Mean and standard deviation of the HV values obtained by the $12$ algorithms on the five-objective MaF problems.}
    \label{Table:5obj_MaF_HV}
    \resizebox{\textwidth}{!}{%
    \renewcommand{\arraystretch}{1.2} 
    \begin{tabular}{@{}cccccccccccccc@{}}
 \toprule
Problem                & Obj.               & NSGA-III                      & RVEA        & MOEA/DD      & RPD-NSGA-II   & 1by1EA      & MaOEA-CSS    & VaEA        & NSGA-II/SDR   & SPEA2+SDE    & SRA         & AR-MOEA      & E3A      \\\midrule
\multirow{2}{*}{MaF1}  & \multirow{2}{*}{5} & 7.271e-03 $+$       & 3.884e-03 $+$       & 3.638e-03 $+$       & 8.698e-03 $+$       & 1.847e-02 $-$       & 1.821e-02 $\approx$ & 1.599e-02 $\approx$ & 1.779e-02 $\approx$ & 1.740e-02 $\approx$ & 1.350e-02 $\approx$ & 1.458e-02 $\approx$ & 1.677e-02  \\
                       &                    & (4.86e-04)          & (1.08e-03)          & (2.47e-03)          & (6.51e-04)          & (4.82e-04)          & (1.47e-04)          & (3.25e-04)          & (2.74e-04)          & (1.89e-04)          & (4.73e-04)          & (3.44e-04)          & (1.72e-04) \\\hline
\multirow{2}{*}{MaF2}  & \multirow{2}{*}{5} & 7.673e-01 $+$       & 7.568e-01 $+$       & 6.236e-01 $+$       & 7.513e-01 $+$       & 7.916e-01 $+$       & 7.773e-01 $+$       & 8.119e-01 $+$       & 8.605e-01 $\approx$ & 8.720e-01 $\approx$ & 8.307e-01 $\approx$ & 7.883e-01 $+$       & 8.734e-01  \\
                       &                    & (1.35e-02)          & (9.55e-03)          & (2.26e-02)          & (9.16e-03)          & (1.36e-02)          & (1.39e-02)          & (1.19e-02)          & (1.10e-02)          & (7.73e-03)          & (1.00e-02)          & (8.11e-03)          & (6.11e-03) \\\hline
\multirow{2}{*}{MaF3}  & \multirow{2}{*}{5} & 3.975e-01 $+$       & 0.000e+00 $+$       & 5.266e-02 $+$       & 1.049e+00 $+$       & 1.132e+00 $+$       & 5.362e-01 $+$       & 4.511e-02 $+$       & 1.562e+00 $\approx$ & 1.471e+00 $\approx$ & 1.437e+00 $\approx$ & 1.357e+00 $\approx$ & 1.419e+00  \\
                       &                    & (6.08e-01)          & (0.00e+00)          & (2.84e-01)          & (6.89e-01)          & (6.00e-01)          & (6.69e-01)          & (2.43e-01)          & (2.18e-02)          & (4.01e-01)          & (4.79e-01)          & (5.41e-01)          & (4.97e-01) \\\hline
\multirow{2}{*}{MaF4}  & \multirow{2}{*}{5} & 2.925e-02 $+$       & 1.631e-02 $+$       & 4.889e-02 $+$       & 1.001e-01 $\approx$ & 3.954e-02 $+$       & 8.523e-02 $+$       & 6.212e-02 $+$       & 1.496e-01 $\approx$ & 9.529e-02 $\approx$ & 7.708e-02 $+$       & 8.642e-02 $+$       & 1.514e-01  \\
                       &                    & (3.49e-02)          & (1.29e-02)          & (2.32e-02)          & (3.33e-02)          & (1.52e-02)          & (4.13e-02)          & (5.14e-02)          & (6.10e-02)          & (3.73e-02)          & (4.33e-02)          & (3.53e-02)          & (4.48e-02) \\\hline
\multirow{2}{*}{MaF5}  & \multirow{2}{*}{5} & 1.250e+00 $\approx$ & 1.225e+00 $\approx$ & 9.926e-01 $+$       & 1.265e+00 $\approx$ & 1.016e+00 $+$       & 9.709e-01 $+$       & 1.240e+00 $\approx$ & 2.424e-01 $+$       & 1.217e+00 $\approx$ & 1.184e+00 $\approx$ & 1.266e+00 $\approx$ & 1.248e+00  \\
                       &                    & (7.47e-02)          & (9.35e-02)          & (1.16e-01)          & (9.49e-03)          & (4.01e-02)          & (2.38e-02)          & (5.38e-03)          & (1.32e-01)          & (4.28e-02)          & (3.61e-02)          & (3.76e-02)          & (6.30e-02) \\\hline
\multirow{2}{*}{MaF6}  & \multirow{2}{*}{5} & 1.962e-01 $+$       & 1.819e-01 $+$       & 1.838e-01 $+$       & 1.805e-01 $+$       & 2.083e-01 $\approx$ & 1.305e-01 $+$       & 2.088e-01 $-$       & 1.962e-01 $+$       & 2.071e-01 $\approx$ & 2.052e-01 $\approx$ & 2.084e-01 $\approx$ & 2.074e-01  \\
                       &                    & (2.73e-03)          & (9.88e-03)          & (2.08e-03)          & (1.06e-02)          & (1.03e-04)          & (3.16e-02)          & (6.15e-05)          & (7.44e-03)          & (4.58e-04)          & (1.22e-03)          & (5.99e-05)          & (5.85e-05) \\\hline
\multirow{2}{*}{MaF7}  & \multirow{2}{*}{5} & 4.893e-01 $+$       & 4.030e-01 $+$       & 1.464e-01 $+$       & 5.145e-01 $\approx$ & 2.933e-01 $+$       & 2.851e-01 $+$       & 4.830e-01 $+$       & 5.195e-01 $\approx$ & 5.445e-01 $\approx$ & 5.160e-01 $\approx$ & 4.829e-01 $+$       & 5.255e-01  \\
                       &                    & (1.36e-02)          & (2.28e-02)          & (6.59e-07)          & (7.36e-03)          & (5.54e-02)          & (3.37e-02)          & (1.08e-02)          & (7.80e-03)          & (9.84e-03)          & (6.73e-03)          & (6.27e-03)          & (9.51e-02) \\\hline
\multirow{2}{*}{MaF8}  & \multirow{2}{*}{5} & 1.605e-01 $+$       & 1.273e-01 $+$       & 1.306e-01 $+$       & 1.417e-01 $+$       & 1.403e-01 $+$       & 2.010e-01 $\approx$ & 1.993e-01 $\approx$ & 1.962e-01 $+$       & 2.014e-01 $\approx$ & 1.496e-01 $+$       & 1.950e-01 $+$       & 2.018e-01  \\
                       &                    & (1.06e-02)          & (7.69e-03)          & (1.63e-02)          & (9.28e-03)          & (1.10e-02)          & (1.78e-03)          & (2.84e-03)          & (2.02e-03)          & (3.40e-03)          & (5.89e-02)          & (6.07e-03)          & (2.47e-03) \\\hline
\multirow{2}{*}{MaF9}  & \multirow{2}{*}{5} & 2.824e-01 $+$       & 3.185e-01 $+$       & 3.816e-01 $+$       & 3.434e-01 $+$       & 3.649e-01 $+$       & 4.101e-01 $+$       & 2.994e-01 $+$       & 4.445e-01 $\approx$ & 4.998e-01 $\approx$ & 4.942e-01 $\approx$ & 4.833e-01 $\approx$ & 4.909e-01  \\
                       &                    & (8.28e-02)          & (2.47e-02)          & (3.47e-02)          & (3.37e-02)          & (2.98e-02)          & (1.20e-02)          & (9.67e-02)          & (9.80e-03)          & (8.83e-03)          & (1.41e-02)          & (1.45e-02)          & (1.34e-02) \\\hline
\multirow{2}{*}{MaF10} & \multirow{2}{*}{5} & 1.062e+00 $+$       & 1.164e+00 $+$       & 7.850e-01 $+$       & 1.348e+00 $\approx$ & 1.127e+00 $+$       & 1.182e+00 $+$       & 8.463e-01 $+$       & 1.269e+00 $+$       & 1.513e+00 $\approx$ & 8.351e-01 $+$       & 1.154e+00 $+$       & 1.546e+00  \\
                       &                    & (6.80e-02)          & (7.88e-02)          & (6.36e-02)          & (6.67e-02)          & (7.23e-02)          & (5.49e-02)          & (6.38e-02)          & (8.40e-02)          & (4.73e-02)          & (7.39e-02)          & (5.56e-02)          & (5.23e-02) \\\hline
\multirow{2}{*}{MaF11} & \multirow{2}{*}{5} & 1.591e+00 $\approx$ & 1.578e+00 $+$       & 1.551e+00 $+$       & 1.593e+00 $\approx$ & 1.564e+00 $+$       & 1.514e+00 $+$       & 1.585e+00 $+$       & 1.551e+00 $+$       & 1.575e+00 $+$       & 1.558e+00 $+$       & 1.597e+00 $\approx$ & 1.606e+00  \\
                       &                    & (3.30e-03)          & (9.67e-03)          & (7.34e-03)          & (3.60e-03)          & (8.61e-03)          & (1.92e-02)          & (3.98e-03)          & (1.20e-02)          & (7.34e-03)          & (7.90e-03)          & (3.15e-03)          & (1.59e-03) \\\hline
\multirow{2}{*}{MaF12} & \multirow{2}{*}{5} & 1.149e+00 $\approx$ & 1.152e+00 $\approx$ & 1.064e+00 $+$       & 1.137e+00 $\approx$ & 9.670e-01 $+$       & 9.259e-01 $+$       & 1.079e+00 $+$       & 1.182e+00 $-$       & 1.157e+00 $\approx$ & 1.114e+00 $+$       & 1.112e+00 $\approx$ & 1.145e+00  \\
                       &                    & (2.08e-02)          & (1.83e-02)          & (2.60e-02)          & (4.36e-02)          & (2.67e-02)          & (3.20e-02)          & (5.41e-02)          & (7.57e-03)          & (9.95e-03)          & (1.37e-02)          & (2.77e-02)          & (4.19e-02) \\\hline
\multirow{2}{*}{MaF13} & \multirow{2}{*}{5} & 3.120e-01 $+$       & 2.787e-01 $+$       & 3.331e-01 $+$       & 1.304e-01 $+$       & 4.816e-01 $\approx$ & 4.455e-01 $+$       & 3.995e-01 $+$       & 4.072e-01 $+$       & 4.872e-01 $\approx$ & 4.802e-01 $\approx$ & 4.491e-01 $+$       & 5.035e-01  \\
                       &                    & (5.28e-02)          & (7.48e-02)          & (8.86e-02)          & (4.48e-02)          & (1.56e-02)          & (1.40e-02)          & (2.61e-02)          & (2.01e-02)          & (1.47e-02)          & (1.55e-02)          & (1.50e-02)          & (1.46e-02) \\\hline
\multirow{2}{*}{MaF14} & \multirow{2}{*}{5} & 0.000e+00 $\approx$ & 4.348e-02 $\approx$ & 1.747e-02 $\approx$ & 0.000e+00 $\approx$ & 1.518e-01 $\approx$ & 8.000e-02 $\approx$ & 0.000e+00 $\approx$ & 1.585e-01 $\approx$ & 1.183e-02 $\approx$ & 4.464e-04 $\approx$ & 3.441e-02 $\approx$ & 4.395e-02  \\
                       &                    & (0.00e+00)          & (6.65e-02)          & (5.06e-02)          & (0.00e+00)          & (1.34e-01)          & (9.18e-02)          & (0.00e+00)          & (1.21e-01)          & (3.81e-02)          & (2.30e-03)          & (6.46e-02)          & (7.26e-02) \\\hline
\multirow{2}{*}{MaF15} & \multirow{2}{*}{5} & 0.000e+00 $\approx$ & 1.454e-02 $-$       & 1.127e-02 $-$       & 0.000e+00 $\approx$ & 3.153e-02 $-$       & 4.106e-02 $-$       & 0.000e+00 $\approx$ & 1.229e-03 $\approx$ & 5.409e-02 $-$       & 2.745e-02 $-$       & 1.155e-02 $-$       & 8.201e-05  \\
                       &                    & (0.00e+00)          & (6.99e-03)          & (7.22e-03)          & (0.00e+00)          & (1.10e-02)          & (9.65e-03)          & (0.00e+00)          & (8.13e-04)          & (1.58e-02)          & (1.23e-02)          & (3.87e-03)          & (1.80e-04)  \\ \hline
\multicolumn{2}{c}{$+$/$\approx$/$-$}                   & 10/5/0 & 11/3/1 &  13/1/1  & 7/8/0 &  10/3/2    & 11/3/1  & 9/5/1  & 6/8/1    &  1/13/1  &   5/9/1    &   6/8/1  &

\\\bottomrule 
    \end{tabular}%
    }
    \end{center}
    \vspace{6pt}
\footnotesize `$+$', `$\approx$', and `$-$' indicate that the result of E3A is significantly better than, equivalent to, and significantly worse than that of the peer algorithm at a $0.05$ level by the statistical tests (i.e., Friedman test and posthoc Nemenyi test), respectively.
\end{table}

\subsubsection{Algorithm Performance on Ten-Objective Problems}
Tables \ref{Table:10obj_MaF_IGD} and \ref{Table:10obj_MaF_HV} show the mean and standard deviation (in parentheses) of the IGD values obtained by E3A and the peer algorithms on the ten-objective MaF problems. From Table \ref{Table:10obj_MaF_IGD}, it can be seen that E3A obtains the best results on four test problems, MaF4, MaF8, MaF10, and MaF13. Concerning pairwise comparison, E3A performs significantly better than peer algorithms except SPEA2+SDE on over half of the ten-objective test problems. Specifically, the proportions of the test problems where E3A significantly outperforms NSGA-III, RVEA, MOEA/DD, RPD-NSGA-II, 1by1EA, MaOEA-CSS, VaEA, NSGA-II/SDR, SPEA2+SDE, SRA, and AR-MOEA are $6/15$, $9/15$, $11/15$, $8/15$, $12/15$, $7/15$, $5/15$, $9/15$, $1/15$, $4/15$, and $4/15$, respectively. Conversely, the proportions of the test problems where E3A performs significantly worse than these peer algorithms are $0/15$, $1/15$, $1/15$, $2/15$, $1/15$, $2/15$, $5/15$, $1/15$, $3/15$, $3/15$, and $2/15$, respectively.

From Table \ref{Table:10obj_MaF_HV}, it can be seen that E3A obtains the best HV results on eight test problems (i.e., MaF$2$--$4$, MaF$10$--$11$, and MaF$13$--$15$). Concerning pairwise comparison, 
the proportions of the test problems where E3A significantly outperforms NSGA-III, RVEA, MOEA/DD, RPD-NSGA-II, 1by1EA, MaOEA-CSS, VaEA, NSGA-II/SDR, SPEA2+SDE, SRA, and AR-MOEA are $6/15$, $8/15$, $11/15$, $6/15$, $7/15$, $8/15$, $4/15$, $6/15$, $2/15$, $6/15$, and $5/15$, respectively. Conversely, the proportions of the test problems where E3A performs significantly worse than these peer algorithms are $2/15$, $2/15$, $1/15$, $4/15$, $1/15$, $1/15$, $0/15$, $3/15$, $1/15$, $0/15$, and $2/15$, respectively.

\begin{table}[tbp]
	\scriptsize
    \begin{center}
    \caption{Mean and standard deviation of the IGD values obtained by the $12$ algorithms on the ten-objective MaF problems.}
    \label{Table:10obj_MaF_IGD}
    \resizebox{\textwidth}{!}{%
    \renewcommand{\arraystretch}{1.2} 
    \begin{tabular}{@{}cccccccccccccc@{}}
 \toprule
Problem                & Obj.               & NSGA-III                      & RVEA        & MOEA/DD      & RPD-NSGA-II   & 1by1EA      & MaOEA-CSS    & VaEA        & NSGA-II/SDR   & SPEA2+SDE    & SRA         & AR-MOEA      & E3A      \\
  \midrule
\multirow{2}{*}{MaF1} & \multirow{2}{*}{10} & 2.960e-1  $\approx$ & 5.769e-1  $+$ & 3.805e-1  $+$ & 3.869e-1  $+$ & 3.936e-1  $+$ & 2.057e-1  $-$ & 2.217e-1  $-$ & 2.244e-1  $\approx$ & 2.311e-1  $\approx$ & 2.648e-1  $\approx$  & 2.373e-1  $\approx$ & 2.474e-1   \\
&           & (4.03e-3)  & (7.79e-2)  & (2.45e-2)  & (5.75e-2)  & (5.56e-2)  & (1.03e-3)  & (1.37e-3)  & (2.17e-3)  & (1.49e-3)  & (3.05e-3)   & (1.47e-3)  & (1.34e-2)   \\\hline

\multirow{2}{*}{MaF2} & \multirow{2}{*}{10} & 2.179e-1  $\approx$ & 3.980e-1  $+$ & 2.572e-1  $+$ & 2.095e-1  $\approx$ & 4.618e-1  $+$ & 1.968e-1  $\approx$ & 1.800e-1  $-$ & 2.365e-1  $+$ & 1.702e-1  $-$ & 1.616e-1  $-$ & 2.073e-1 $\approx$ & 2.064e-1   \\
&           & (1.95e-2)  & (1.85e-1)  & (3.10e-2)  & (3.60e-3)  & (2.03e-2)  & (9.51e-3)  & (2.26e-3)  & (1.78e-2)  & (4.17e-3)  & (3.22e-3)   & (7.29e-3)  & (7.20e-3)   \\\hline

\multirow{2}{*}{MaF3} & \multirow{2}{*}{10}  & 1.947e+3  $+$ & 6.067e+0  $+$ & 6.370e+1  $+$ & 4.772e-1  $+$ & 4.235e+1  $+$ & 2.403e+0  $\approx$ & 3.717e+4  $+$ & 1.670e-1  $\approx$ & 1.066e-1 $\approx$ & 1.862e+2  $\approx$  & 3.234e+0  $\approx$ & 1.087e-1   \\
&           & (2.98e+3)  & (6.75e+0)  & (4.47e+1)  & (1.01e+0)  & (6.52e+1)  & (6.33e+0)  & (6.77e+4)  & (9.34e-2)  & (5.51e-3)  & (1.02e+3)   & (1.11e+1)  & (1.75e-2)   \\\hline

\multirow{2}{*}{MaF4} & \multirow{2}{*}{10}  & 1.226e+2  $+$ & 1.977e+2  $+$ & 4.041e+2  $+$ & 1.403e+2  $+$ & 3.095e+2  $+$ & 1.483e+2  $+$ & 5.870e+1  $\approx$ & 1.961e+2  $+$ & 1.340e+2  $+$ & 1.216e+2  $+$  & 9.843e+1  $\approx$ & 5.416e+1   \\
&           & (8.39e+1)  & (5.49e+1)  & (2.17e+1)  & (2.54e+1)  & (2.58e+1)  & (2.01e+1)  & (4.76e+0)  & (3.17e+1)  & (1.62e+1)  & (2.74e+1)   & (5.11e+0)  & (2.35e+0)   \\\hline

\multirow{2}{*}{MaF5} & \multirow{2}{*}{10} & 8.665e+1  $\approx$ & 9.436e+1  $\approx$ & 2.875e+2  $+$ & 7.743e+1 $\approx$ & 2.052e+2  $+$ & 2.603e+2  $+$ & 4.705e+1  $-$ & 2.987e+2  $+$ & 6.704e+1  $\approx$ & 5.660e+1  $\approx$  & 1.006e+2  $+$ & 7.620e+1   \\
&           & (1.58e+0)  & (6.80e+0)  & (1.46e+1)  & (4.31e+0)  & (1.51e+1)  & (1.70e+1)  & (1.92e+0)  & (2.89e+1)  & (3.74e+0)  & (4.90e+0)   & (4.03e+0)  & (3.01e+0)   \\\hline

\multirow{2}{*}{MaF6} & \multirow{2}{*}{10} & 6.623e-1  $\approx$ & 1.468e-1  $-$ & 1.100e-1  $-$ & 2.515e-1  $-$ & 1.918e-3  $-$ & 2.242e-1  $-$ & 1.121e+0 $\approx$ & 7.803e-3  $-$ & 8.912e-1  $\approx$ & 4.159e+0  $\approx$  & 1.620e-1  $-$ & 1.304e+0   \\
&           & (2.33e-1)  & (8.93e-2)  & (1.12e-2)  & (1.64e-1)  & (3.00e-5)  & (8.62e-2)  & (4.72e-1)  & (2.17e-3)  & (4.56e-1)  & (1.48e+0)   & (2.24e-1)  & (5.71e-1)   \\\hline

\multirow{2}{*}{MaF7} & \multirow{2}{*}{10} & 1.219e+0 $\approx$ & 2.036e+0  $\approx$ & 2.130e+0  $\approx$ & 1.703e+0 $\approx$ & 2.412e+0  $+$ & 2.426e+0  $+$ & 1.078e+0  $-$ & 1.653e+0 $\approx$ & 8.581e-1  $-$ & 8.630e-1  $-$  & 1.418e+0 $\approx$ & 1.675e+0   \\
&           & (1.04e-1)  & (5.43e-1)  & (3.11e-1)  & (2.17e-1)  & (4.15e-1)  & (4.54e-1)  & (4.37e-2)  & (3.43e-1)  & (3.24e-2)  & (2.40e-2)   & (8.22e-2)  & (7.06e-1)   \\\hline

\multirow{2}{*}{MaF8} & \multirow{2}{*}{10} & 3.835e-1  $+$ & 7.819e-1  $+$ & 8.993e-1  $+$ & 5.342e-1  $+$ & 4.379e-1  $+$ & 1.222e-1  $\approx$ & 1.158e-1  $\approx$ & 1.633e-1  $+$ & 1.281e-1  $\approx$ & 7.649e-1  $+$  & 1.384e-1  $+$ & 1.135e-1 \\
&           & (7.49e-2)  & (1.07e-1)  & (1.97e-2)  & (9.71e-2)  & (6.95e-2)  & (5.56e-3)  & (2.73e-3)  & (1.77e-2)  & (6.73e-3)  & (1.97e+0)   & (4.46e-3)  & (1.86e-3)   \\\hline

\multirow{2}{*}{MaF9} & \multirow{2}{*}{10} & 5.916e-1  $+$ & 8.262e-1  $+$ & 4.948e-1  $+$ & 4.154e-1  $+$ & 1.238e-1  $\approx$ & 1.861e-1  $\approx$ & 1.639e-1  $\approx$ & 1.880e-1  $\approx$ & 1.063e-1  $-$ & 1.113e-1  $-$  & 1.739e-1 $\approx$ & 1.762e-1 \\
&           & (1.98e-1)  & (2.53e-1)  & (9.87e-2)  & (4.84e-2)  & (1.44e-2)  & (7.91e-3)  & (3.65e-2)  & (9.22e-3)  & (1.25e-3)  & (3.20e-3)   & (7.17e-3)  & (1.87e-2)   \\\hline

\multirow{2}{*}{MaF10} & \multirow{2}{*}{10} & 1.646e+0  $+$ & 1.292e+0  $+$ & 1.712e+0  $+$ & 1.052e+0  $\approx$ & 1.760e+0  $+$ & 1.438e+0  $+$ & 2.453e+0  $+$ & 1.703e+0  $+$ & 1.143e+0  $\approx$ & 1.938e+0  $+$  & 1.489e+0  $+$ & 9.818e-1 \\
&           & (1.29e-1)  & (6.00e-2)  & (1.05e-1)  & (2.99e-2)  & (9.65e-2)  & (6.38e-2)  & (1.70e-1)  & (1.12e-1)  & (4.02e-2)  & (1.46e-1)   & (5.49e-2)  & (2.16e-2)   \\\hline

\multirow{2}{*}{MaF11} & \multirow{2}{*}{10} & 1.262e+0 $\approx$ & 1.148e+0  $\approx$ & 1.483e+0  $\approx$ & 1.129e+0  $-$ & 1.776e+0  $+$ & 1.680e+0  $+$ & 1.022e+0  $-$ & 1.668e+0  $+$ & 1.179e+0  $\approx$ & 1.182e+0  $\approx$  & 1.10e+0  $-$  & 1.284e+0 \\
&           & (1.31e-1)  & (4.48e-2)  & (2.98e-2)  & (1.64e-2)  & (6.44e-2)  & (7.25e-2)  & (1.33e-2)  & (1.10e-1)  & (3.97e-2)  & (3.62e-2)   & (3.22e-2)  & (1.30e-1)   \\\hline

\multirow{2}{*}{MaF12} & \multirow{2}{*}{10} & 4.481e+0  $+$ & 4.488e+0  $+$ & 6.195e+0  $+$ & 4.632e+0  $+$ & 5.584e+0  $+$ & 5.504e+0  $+$ & 3.994e+0  $\approx$ & 4.500e+0  $+$ & 4.471e+0  $\approx$ & 4.438e+0  $\approx$  & 4.619e+0  $+$ & 4.214e+0 \\
&           & (4.07e-2)  & (7.37e-2)  & (2.09e-1)  & (6.21e-2)  & (2.07e-1)  & (1.80e-1)  & (2.48e-2)  & (4.85e-2)  & (4.53e-2)  & (7.41e-2)   & (3.47e-2)  & (4.85e-2)   \\\hline

\multirow{2}{*}{MaF13} & \multirow{2}{*}{10} & 2.580e-1  $+$ & 9.350e-1  $+$ & 2.780e-1  $+$ & 5.802e-1  $+$ & 1.946e-1  $+$ & 1.506e-1  $+$ & 1.387e-1  $+$ & 1.856e-1  $+$ & 1.211e-1  $\approx$ & 1.276e-1  $\approx$  & 1.281e-1  $\approx$ & 1.027e-1 \\
&           & (2.71e-2)  & (2.17e-1)  & (2.73e-2)  & (6.02e-2)  & (3.22e-2)  & (1.01e-2)  & (1.33e-2)  & (1.93e-2)  & (2.58e-2)  & (1.95e-2)   & (7.20e-3)  & (1.08e-2)   \\\hline

\multirow{2}{*}{MaF14} & \multirow{2}{*}{10} & 1.364e+1  $+$ & 1.491e+0  $\approx$ & 1.777e+0 $\approx$ & 6.032e+0  $\approx$ & 1.627e+0  $\approx$ & 1.613e+0  $\approx$ & 1.349e+1  $+$ & 1.465e+0  $\approx$ & 2.656e+0 $\approx$ & 5.655e+0  $+$  & 1.487e+0  $\approx$ & 2.281e+0 \\
&           & (5.88e+0)  & (3.11e-1)  & (4.19e-1)  & (5.79e+0)  & (3.83e-1)  & (1.53e-1)  & (4.08e+0)  & (1.60e-1)  & (8.41e-1)  & (1.68e+0)   & (4.46e-1)  & (1.00e+0)   \\\hline

\multirow{2}{*}{MaF15} & \multirow{2}{*}{10} & 3.841e+0  $+$ & 1.122e+0  $\approx$ & 1.354e+0  $+$ & 1.384e+0  $+$ & 1.196e+0  $+$ & 9.876e-1  $\approx$ & 1.918e+0  $+$ & 1.196e+0  $+$ & 8.632e-1 $\approx$ & 1.006e+0  $\approx$  & 1.103e+0  $\approx$ & 9.521e-1 \\
&           & (1.75e+0)  & (1.26e-1)  & (2.06e-1)  & (3.40e-1)  & (6.48e-2)  & (5.72e-2)  & (2.90e-1)  & (7.23e-2)  & (1.80e-1)  & (1.42e-1)   & (8.71e-2)  & (4.18e-1)
\\\hline

\multicolumn{2}{c}{$+$/$\approx$/$-$} & 6/9/0 & 9/5/1 &  11/3/1 &   8/5/2   & 12/2/1    &   7/6/2   &   5/5/5  & 9/5/1  & 1/11/3   &  4/8/3    & 4/9/2        &           
\\\bottomrule 
    \end{tabular}%
    }
    \end{center}
    \vspace{6pt}
\footnotesize `$+$', `$\approx$', and `$-$' indicate that the result of E3A is significantly better than, equivalent to, and significantly worse than that of the peer algorithm at a $0.05$ level by the statistical tests (i.e., Friedman test and posthoc Nemenyi test), respectively.
\end{table}

\begin{table}[tbp]
	\scriptsize
    \begin{center}
    \caption{Mean and standard deviation of the HV values obtained by the $12$ algorithms on the ten-objective MaF problems.}
    \label{Table:10obj_MaF_HV}
    \resizebox{\textwidth}{!}{%
    \renewcommand{\arraystretch}{1.2} 
    \begin{tabular}{@{}cccccccccccccc@{}}
 \toprule
Problem                & Obj.               & NSGA-III                      & RVEA        & MOEA/DD      & RPD-NSGA-II   & 1by1EA      & MaOEA-CSS    & VaEA        & NSGA-II/SDR   & SPEA2+SDE    & SRA         & AR-MOEA      & E3A      \\\midrule
\multirow{2}{*}{MaF1} & \multirow{2}{*}{10} & 1.210e-06 $\approx$ & 0.000e+00 $\approx$ & 1.729e-07 $\approx$ & 2.594e-07 $\approx$ & 5.187e-07 $\approx$ & 2.075e-06 $\approx$ & 6.917e-07 $\approx$ & 1.556e-06 $\approx$ & 7.781e-07 $\approx$ & 3.458e-07 $\approx$ & 1.124e-06 $\approx$ & 4.323e-07  \\ 
        &      & (1.74e-06)          & (0.00e+00)          & (6.47e-07)          & (7.78e-07)          & (1.04e-06)          & (2.54e-06)          & (1.33e-06)          & (2.07e-06)          & (1.66e-06)          & (8.82e-07)          & (2.19e-06)          & (9.67e-07) \\ \hline
\multirow{2}{*}{MaF2} & \multirow{2}{*}{10} & 1.519e+00 $+$       & 1.079e+00 $+$       & 1.237e+00 $+$       & 1.468e+00 $+$       & 1.111e+00 $+$       & 1.501e+00 $+$       & 1.606e+00 $\approx$ & 1.569e+00 $+$       & 1.583e+00 $\approx$ & 1.573e+00 $+$       & 1.573e+00 $+$       & 1.678e+00  \\
        &      & (2.37e-02)          & (4.36e-01)          & (3.92e-02)          & (6.93e-03)          & (5.72e-02)          & (1.93e-02)          & (1.01e-02)          & (1.46e-02)          & (1.29e-02)          & (7.56e-03)          & (1.01e-02)          & (5.44e-03) \\ \hline
\multirow{2}{*}{MaF3} & \multirow{2}{*}{10} & 0.000e+00 $+$       & 1.527e-01 $+$       & 2.545e-03 $+$       & 2.054e+00 $+$       & 1.124e-01 $+$       & 1.758e+00 $+$       & 0.000e+00 $+$       & 2.538e+00 $+$       & 2.589e+00 $\approx$ & 2.324e+00 $\approx$ & 2.122e+00 $\approx$ & 2.594e+00  \\
        &      & (0.00e+00)          & (4.99e-01)          & (1.37e-02)          & (5.26e-01)          & (4.74e-01)          & (1.16e+00)          & (0.00e+00)          & (1.92e-01)          & (1.91e-03)          & (7.14e-01)          & (9.58e-01)          & (1.55e-04) \\ \hline
\multirow{2}{*}{MaF4} & \multirow{2}{*}{10} & 4.470e-04 $\approx$ & 6.917e-07 $+$       & 6.917e-07 $+$       & 2.907e-04 $\approx$ & 3.977e-06 $+$       & 1.107e-05 $+$       & 1.660e-04 $\approx$ & 2.585e-05 $+$       & 5.360e-06 $+$       & 2.075e-06 $+$       & 5.620e-06 $+$       & 7.454e-04  \\
        &      & (1.81e-04)          & (2.00e-06)          & (1.49e-06)          & (1.24e-04)          & (5.71e-06)          & (8.81e-06)          & (5.42e-05)          & (1.29e-05)          & (4.23e-06)          & (3.02e-06)          & (7.46e-06)          & (7.66e-05) \\ \hline
\multirow{2}{*}{MaF5} & \multirow{2}{*}{10}  & 2.507e+00 $-$       & 2.458e+00 $\approx$ & 1.473e+00 $+$       & 2.510e+00 $-$       & 2.062e+00 $+$       & 1.727e+00 $+$       & 2.374e+00 $\approx$ & 2.886e-01 $+$       & 2.292e+00 $\approx$ & 2.125e+00 $+$       & 2.496e+00 $\approx$ & 2.433e+00  \\
        &      & (6.88e-04)          & (1.37e-02)          & (6.94e-02)          & (2.10e-03)          & (3.31e-02)          & (5.85e-02)          & (3.10e-02)          & (1.28e-01)          & (1.38e-02)          & (5.78e-02)          & (4.96e-03)          & (4.39e-03) \\ \hline
\multirow{2}{*}{MaF6} & \multirow{2}{*}{10} & 5.100e-03 $\approx$ & 1.838e-01 $-$       & 2.319e-01 $-$       & 2.410e-01 $-$       & 2.622e-01 $-$       & 5.480e-03 $\approx$ & 8.716e-03 $\approx$ & 2.595e-01 $-$       & 2.599e-02 $\approx$ & 0.000e+00 $\approx$ & 1.671e-01 $-$       & 8.639e-03  \\
        &      & (1.91e-02)          & (8.72e-02)          & (1.57e-02)          & (7.98e-03)          & (5.77e-04)          & (1.01e-02)          & (4.69e-02)          & (1.19e-03)          & (7.80e-02)          & (0.00e+00)          & (1.25e-01)          & (4.65e-02) \\ \hline
\multirow{2}{*}{MaF7} & \multirow{2}{*}{10} & 4.202e-01 $-$       & 3.327e-01 $-$       & 4.242e-04 $+$       & 4.615e-01 $-$       & 9.377e-02 $\approx$ & 8.201e-04 $+$       & 2.876e-01 $\approx$ & 3.987e-01 $-$       & 1.816e-01 $\approx$ & 2.012e-01 $\approx$ & 3.278e-01 $-$       & 1.677e-01  \\
        &      & (3.18e-02)          & (5.20e-02)          & (4.03e-04)          & (2.96e-02)          & (4.65e-02)          & (6.69e-04)          & (3.58e-02)          & (1.78e-01)          & (7.08e-02)          & (6.21e-02)          & (1.70e-02)          & (1.34e-01) \\ \hline
\multirow{2}{*}{MaF8} & \multirow{2}{*}{10}  & 2.480e-02 $+$       & 1.362e-02 $+$       & 1.710e-02 $+$       & 1.868e-02 $+$       & 2.920e-02 $\approx$ & 3.085e-02 $-$       & 3.057e-02 $\approx$ & 2.970e-02 $\approx$ & 3.021e-02 $\approx$ & 2.300e-02 $+$       & 2.944e-02 $\approx$ & 2.978e-02  \\
        &      & (1.08e-03)          & (1.50e-03)          & (4.99e-04)          & (1.20e-03)          & (9.29e-04)          & (3.02e-04)          & (3.06e-04)          & (3.30e-04)          & (3.16e-04)          & (9.20e-03)          & (3.12e-04)          & (3.12e-04) \\ \hline
\multirow{2}{*}{MaF9} & \multirow{2}{*}{10} & 2.219e-02 $+$       & 1.214e-02 $+$       & 1.808e-02 $+$       & 2.726e-02 $+$       & 4.805e-02 $\approx$ & 4.147e-02 $\approx$ & 4.263e-02 $\approx$ & 4.074e-02 $\approx$ & 4.814e-02 $-$       & 4.766e-02 $\approx$ & 4.156e-02 $\approx$ & 4.302e-02  \\
        &      & (6.37e-03)          & (2.88e-03)          & (3.58e-03)          & (2.68e-03)          & (5.13e-04)          & (5.23e-04)          & (3.40e-03)          & (7.63e-04)          & (3.43e-04)          & (3.83e-04)          & (7.35e-04)          & (1.24e-03) \\ \hline
\multirow{2}{*}{MaF10} & \multirow{2}{*}{10} & 1.559e+00 $+$       & 2.259e+00 $+$       & 1.640e+00 $+$       & 2.397e+00 $\approx$ & 1.812e+00 $+$       & 2.150e+00 $+$       & 8.373e-01 $+$       & 2.512e+00 $\approx$ & 2.514e+00 $\approx$ & 1.223e+00 $+$       & 1.807e+00 $+$       & 2.591e+00  \\
        &      & (1.35e-01)          & (1.81e-01)          & (1.76e-01)          & (1.30e-01)          & (1.29e-01)          & (1.16e-01)          & (7.62e-02)          & (8.37e-02)          & (9.53e-02)          & (1.39e-01)          & (1.17e-01)          & (1.66e-03) \\ \hline
\multirow{2}{*}{MaF11} & \multirow{2}{*}{10} & 2.578e+00 $\approx$ & 2.536e+00 $+$       & 2.482e+00 $+$       & 2.557e+00 $+$       & 2.561e+00 $+$       & 2.537e+00 $+$       & 2.560e+00 $+$       & 2.547e+00 $+$       & 2.562e+00 $+$       & 2.552e+00 $+$       & 2.565e+00 $+$       & 2.589e+00  \\
        &      & (5.55e-03)          & (1.11e-02)          & (2.20e-02)          & (7.37e-03)          & (6.21e-03)          & (1.32e-02)          & (6.38e-03)          & (8.96e-03)          & (6.19e-03)          & (8.31e-03)          & (7.74e-03)          & (2.30e-03) \\ \hline
\multirow{2}{*}{MaF12} & \multirow{2}{*}{10} & 2.190e+00 $\approx$ & 2.122e+00 $\approx$ & 1.644e+00 $+$       & 2.240e+00 $-$       & 1.867e+00 $+$       & 1.780e+00 $+$       & 2.106e+00 $\approx$ & 2.326e+00 $-$       & 2.186e+00 $\approx$ & 2.107e+00 $\approx$ & 2.090e+00 $\approx$ & 2.140e+00  \\
        &      & (7.71e-02)          & (7.14e-02)          & (1.06e-01)          & (6.92e-02)          & (6.61e-02)          & (4.88e-02)          & (6.95e-02)          & (6.56e-03)          & (2.70e-02)          & (2.74e-02)          & (5.84e-02)          & (7.53e-02) \\ \hline
\multirow{2}{*}{MaF13} & \multirow{2}{*}{10} & 3.346e-01 $+$       & 3.188e-01 $+$       & 3.317e-01 $+$       & 2.118e-02 $+$       & 5.582e-01 $\approx$ & 5.351e-01 $\approx$ & 4.633e-01 $+$       & 4.383e-01 $+$       & 5.446e-01 $\approx$ & 5.362e-01 $\approx$ & 5.153e-01 $+$       & 5.505e-01  \\
        &      & (6.94e-02)          & (9.41e-02)          & (7.87e-02)          & (5.11e-02)          & (6.53e-03)          & (7.07e-03)          & (3.69e-02)          & (1.75e-02)          & (8.50e-03)          & (8.34e-03)          & (1.38e-02)          & (8.99e-03) \\ \hline
\multirow{2}{*}{MaF14} & \multirow{2}{*}{10} & 0.000e+00 $\approx$ & 0.000e+00 $\approx$ & 5.473e-05 $\approx$ & 0.000e+00 $\approx$ & 1.628e-02 $\approx$ & 2.616e-04 $\approx$ & 0.000e+00 $\approx$ & 0.000e+00 $\approx$ & 0.000e+00 $\approx$ & 0.000e+00 $\approx$ & 5.470e-02 $\approx$ & 8.338e-03  \\
        &      & (0.00e+00)          & (0.00e+00)          & (2.95e-04)          & (0.00e+00)          & (7.02e-02)          & (1.41e-03)          & (0.00e+00)          & (0.00e+00)          & (0.00e+00)          & (0.00e+00)          & (1.17e-01)          & (4.10e-02) \\ \hline
\multirow{2}{*}{MaF15} & \multirow{2}{*}{10} & 0.000e+00 $\approx$ & 8.646e-08 $\approx$ & 0.000e+00 $\approx$ & 0.000e+00 $\approx$ & 0.000e+00 $\approx$ & 0.000e+00 $\approx$ & 0.000e+00 $\approx$ & 0.000e+00 $\approx$ & 5.533e-06 $\approx$ & 0.000e+00 $\approx$ & 0.000e+00 $\approx$ & 1.037e-06  \\
        &      & (0.00e+00)          & (4.66e-07)          & (0.00e+00)          & (0.00e+00)          & (0.00e+00)          & (0.00e+00)          & (0.00e+00)          & (0.00e+00)          & (9.84e-06)          & (0.00e+00)          & (0.00e+00)          & (2.73e-06) \\ \hline
\multicolumn{2}{c}{$+$/$\approx$/$-$}                   &  6/7/2 & 8/5/2  & 11/3/1 &  6/5/4   &   7/7/1   & 8/6/1  & 4/11/0  &  6/6/3   &  2/12/1  & 6/9/0  &   5/8/2    &    
\\\bottomrule 
    \end{tabular}%
    }
    \end{center}
    \vspace{6pt}
\footnotesize `$+$', `$\approx$', and `$-$' indicate that the result of E3A is significantly better than, equivalent to, and significantly worse than that of the peer algorithm at a $0.05$ level by the statistical tests (i.e., Friedman test and posthoc Nemenyi test), respectively.
\end{table}

\begin{figure}[tbp]
	\begin{center}
		\footnotesize
		\begin{tabular}{@{}cccc}
			\includegraphics[scale=0.257]{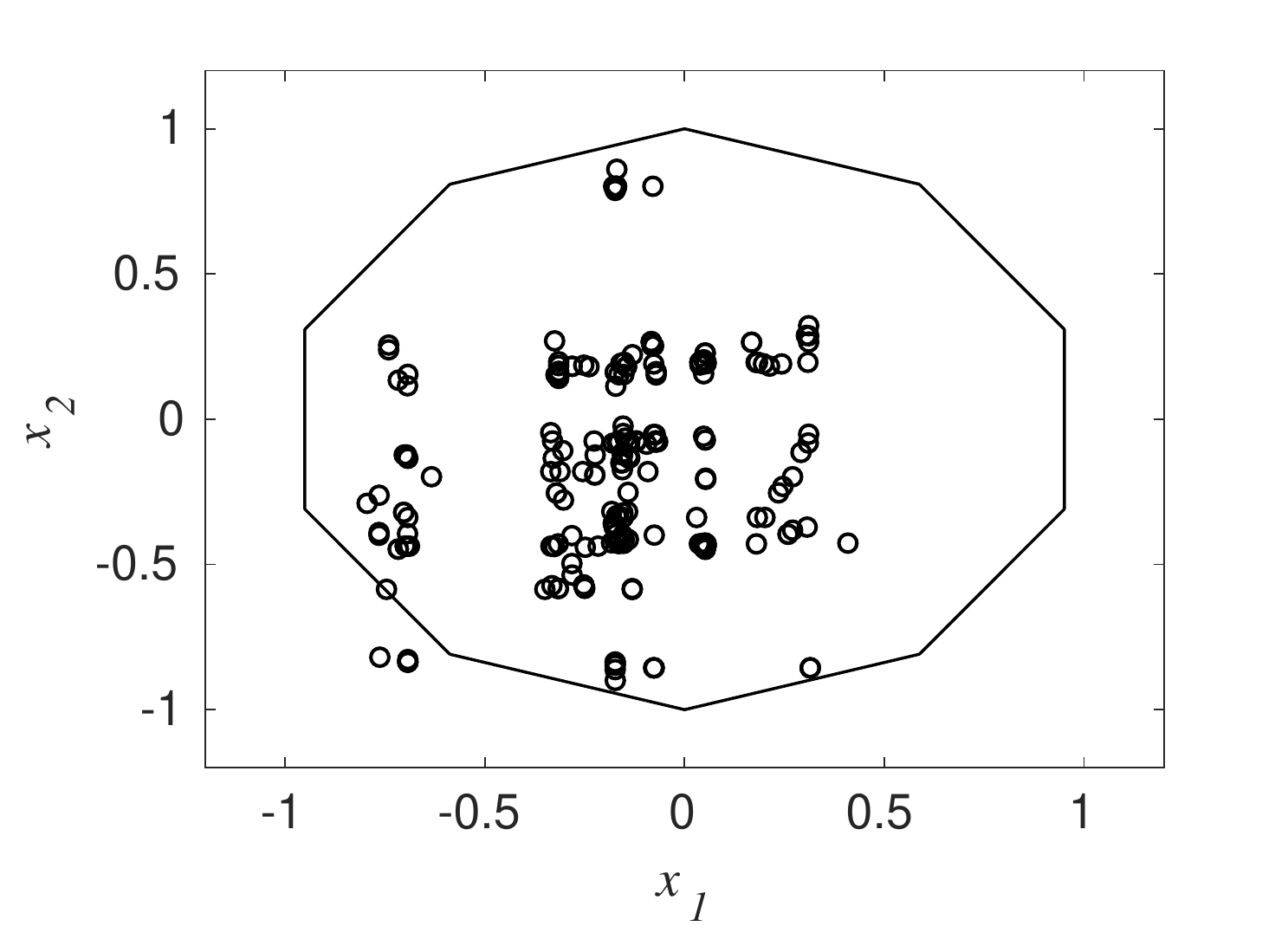}&
			\includegraphics[scale=0.257]{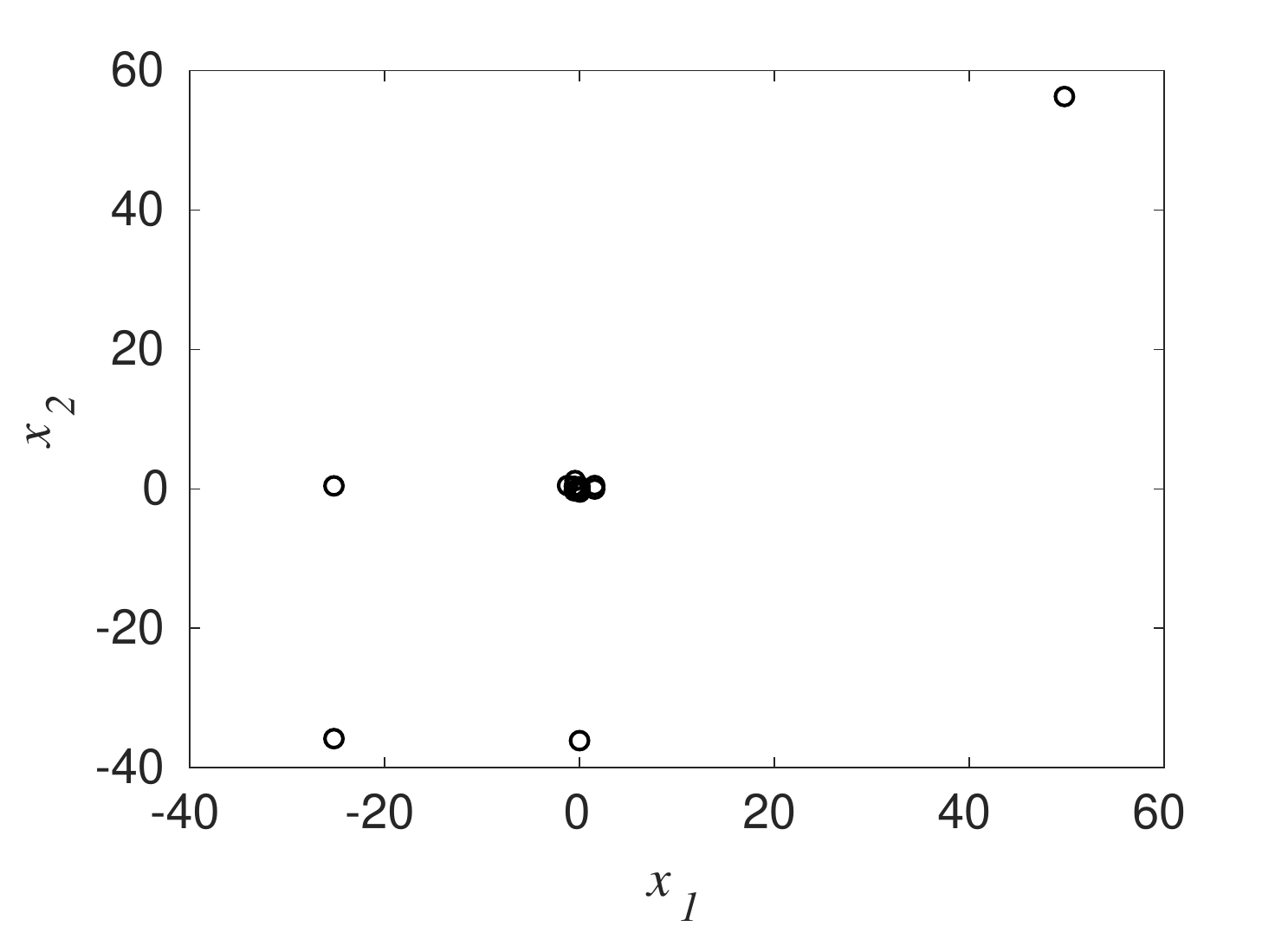}&
			\includegraphics[scale=0.257]{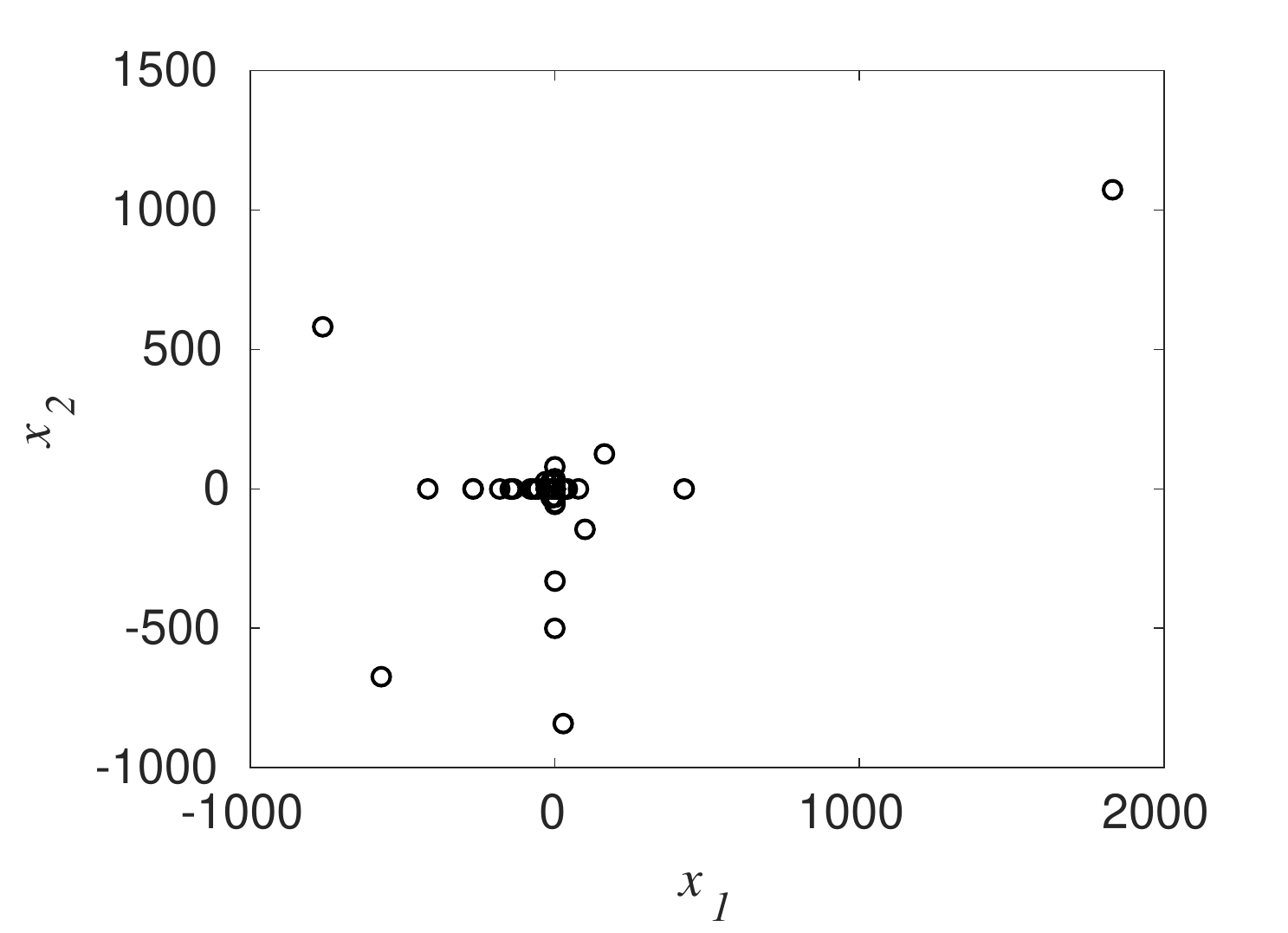}&
			\includegraphics[scale=0.257]{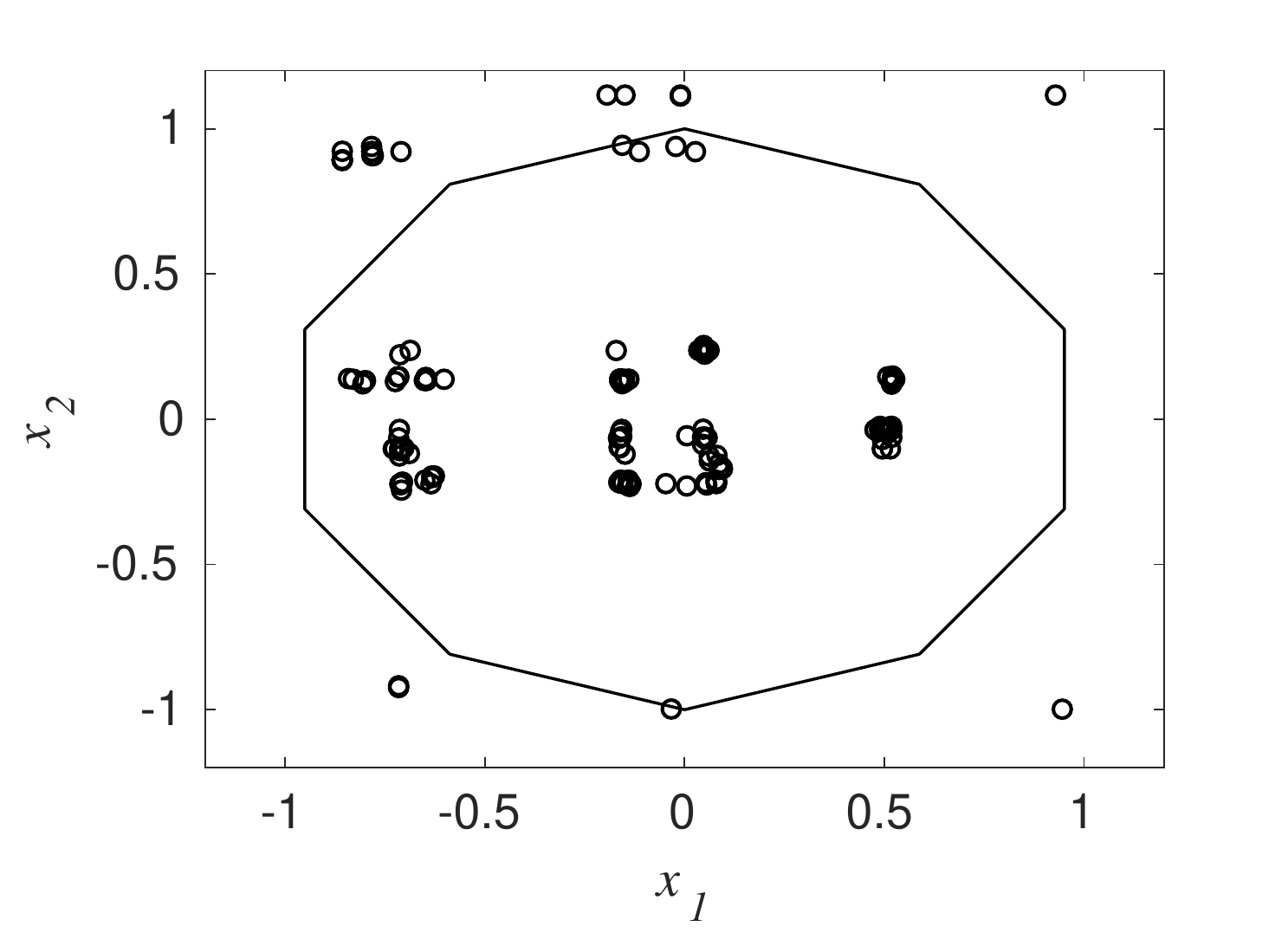}\\
			(a) NSGA-III  & (b) RVEA  & (c) MOEA/DD &  (d) RPD-NSGA-II \\
		\end{tabular}
		\begin{tabular}{@{}cccc}
			\includegraphics[scale=0.257]{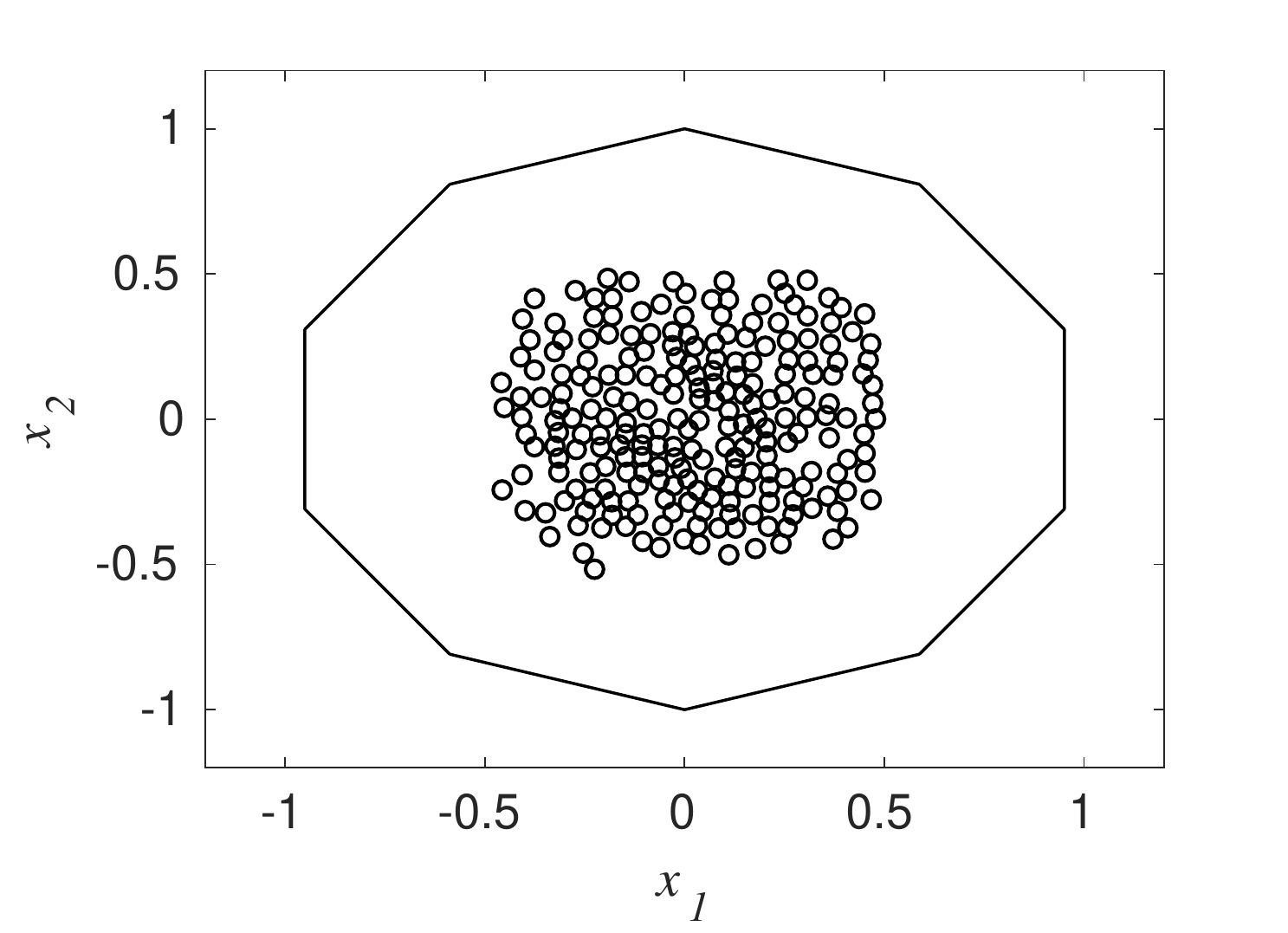}&
			\includegraphics[scale=0.257]{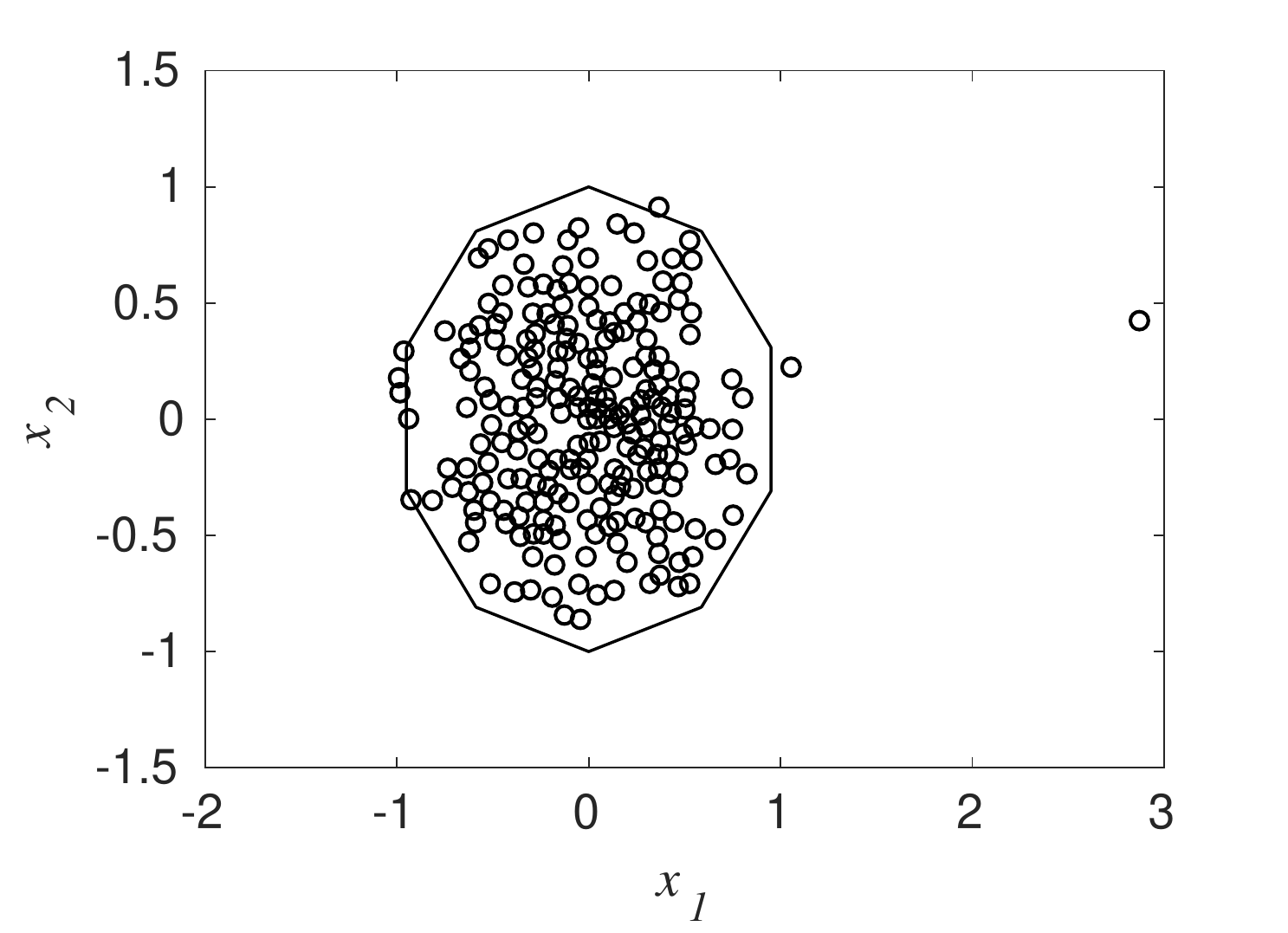}&
			\includegraphics[scale=0.257]{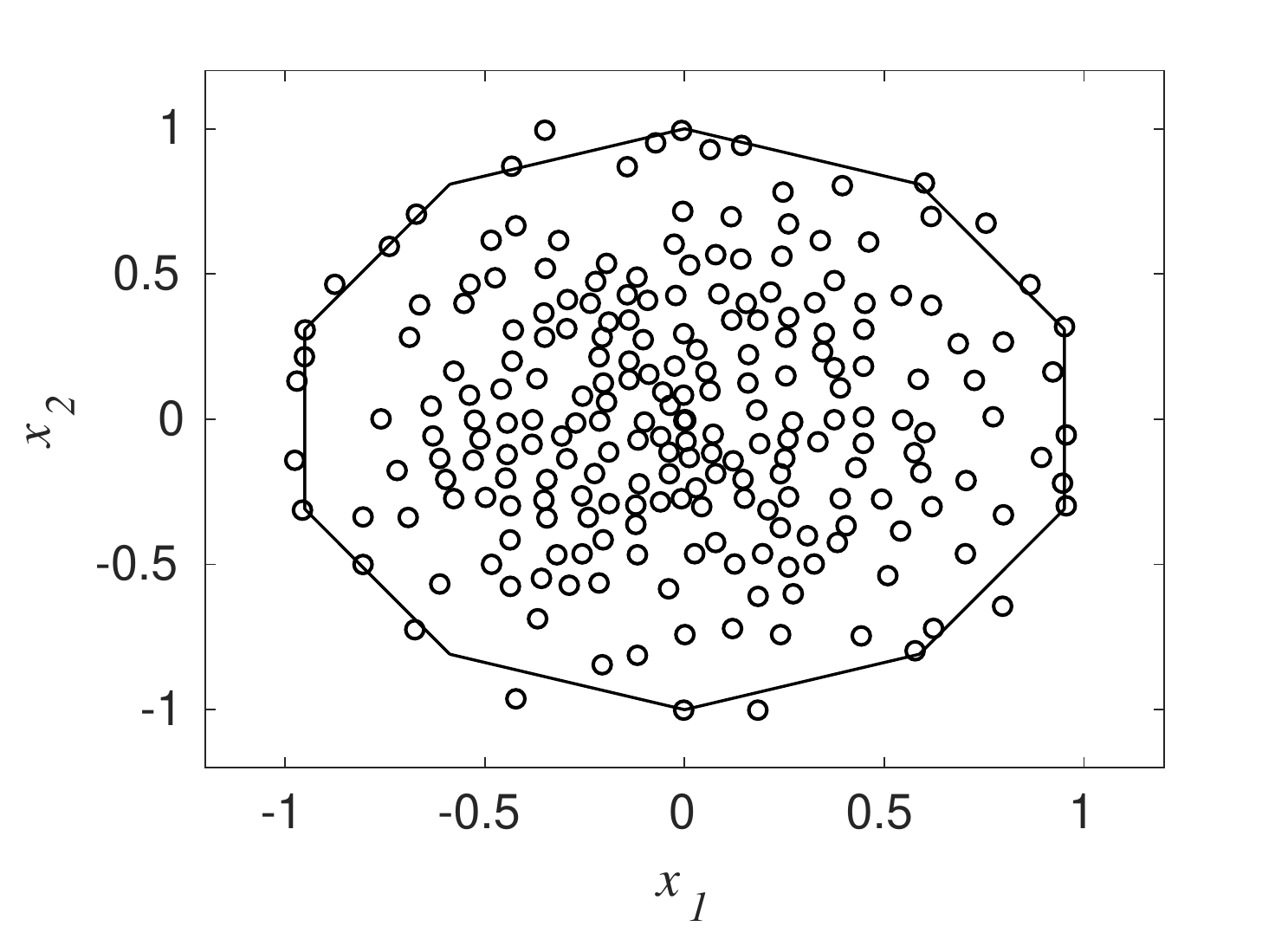}&
			\includegraphics[scale=0.257]{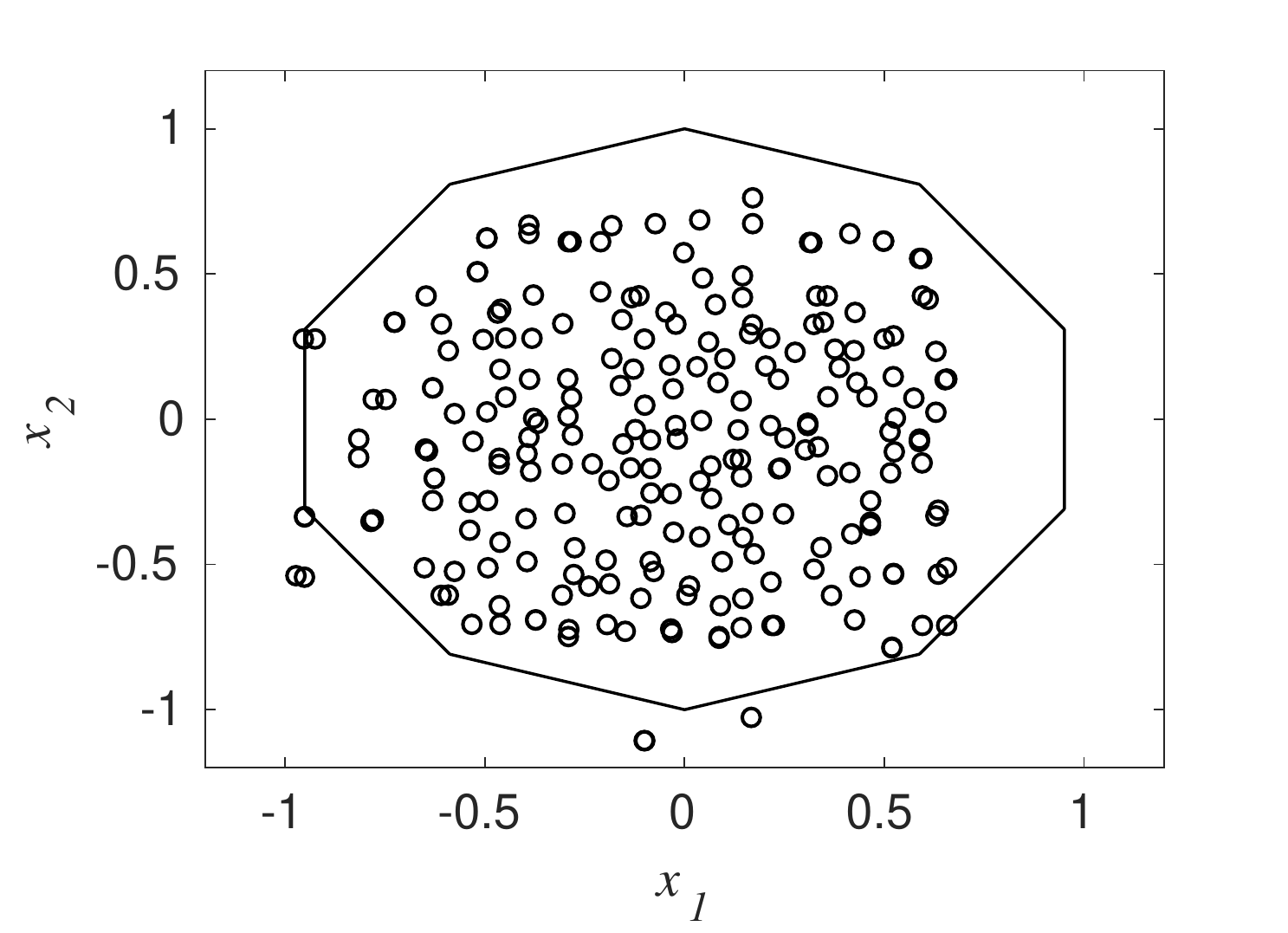}\\
            (e) 1by1EA  & (f) MaOEA-CSS  &  (g) VaEA  & (h) NSGA-II/SDR  \\
		\end{tabular}
		\begin{tabular}{@{}cccc}
			\includegraphics[scale=0.257]{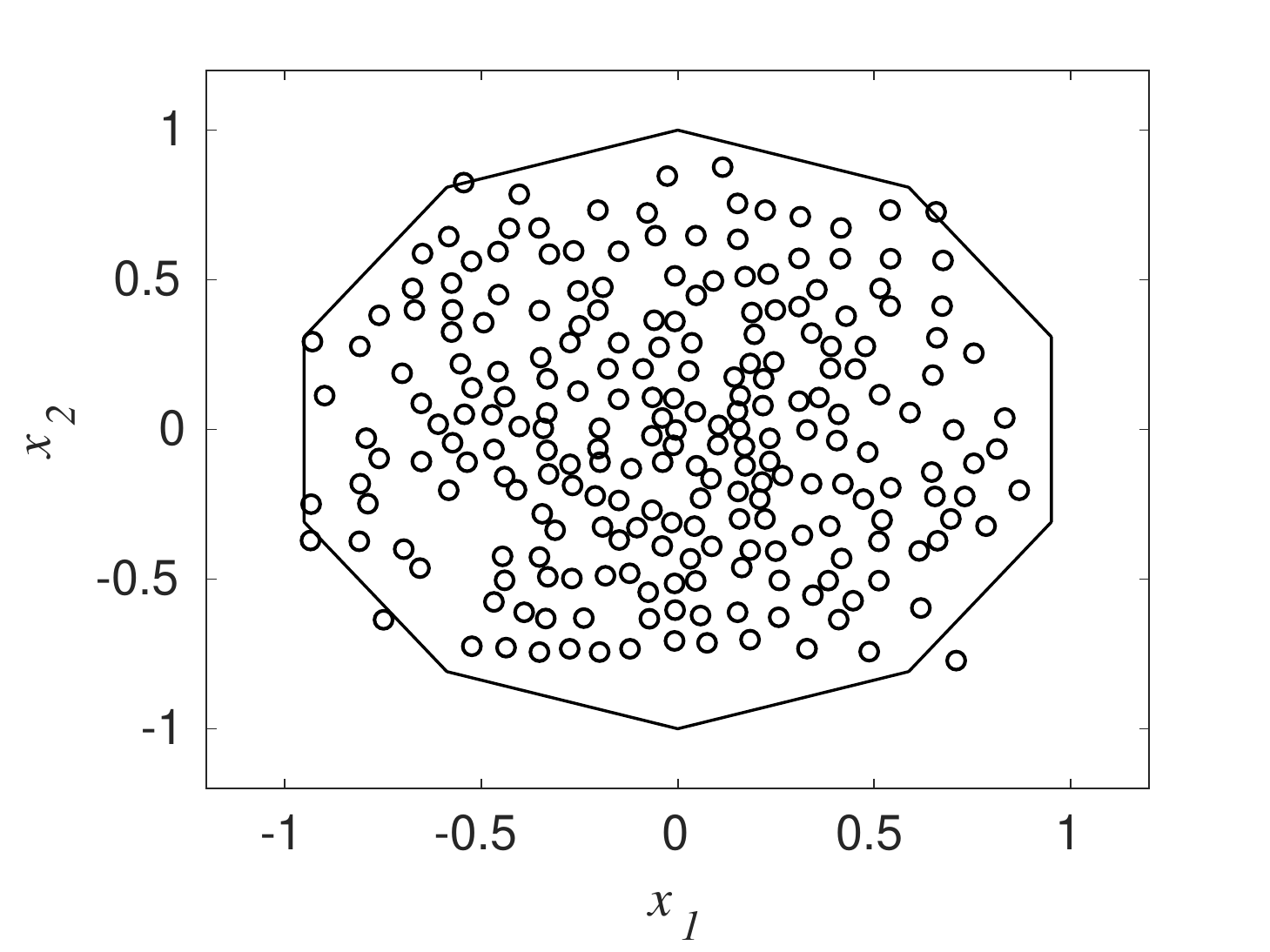}&
			\includegraphics[scale=0.257]{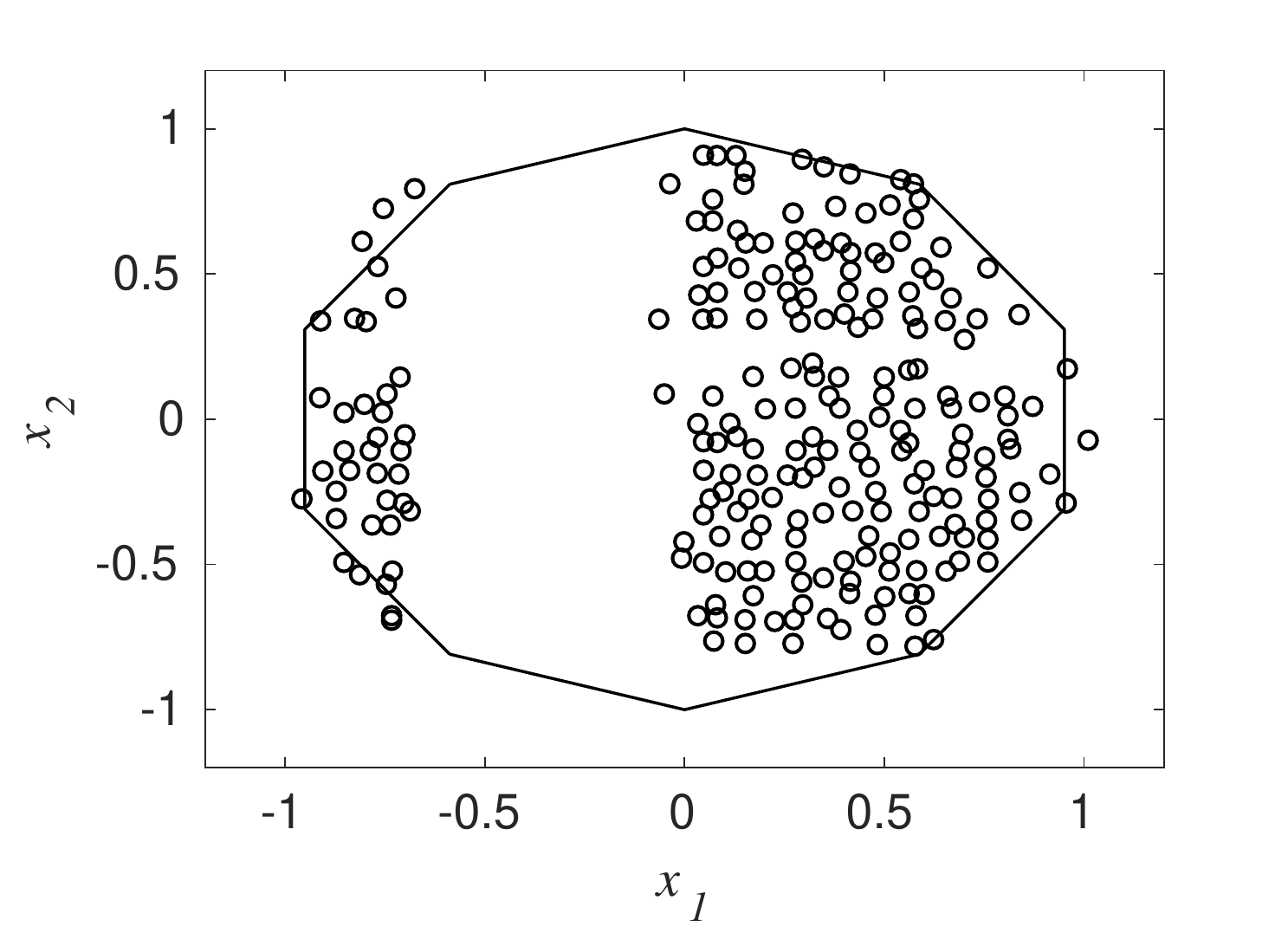}&
			\includegraphics[scale=0.257]{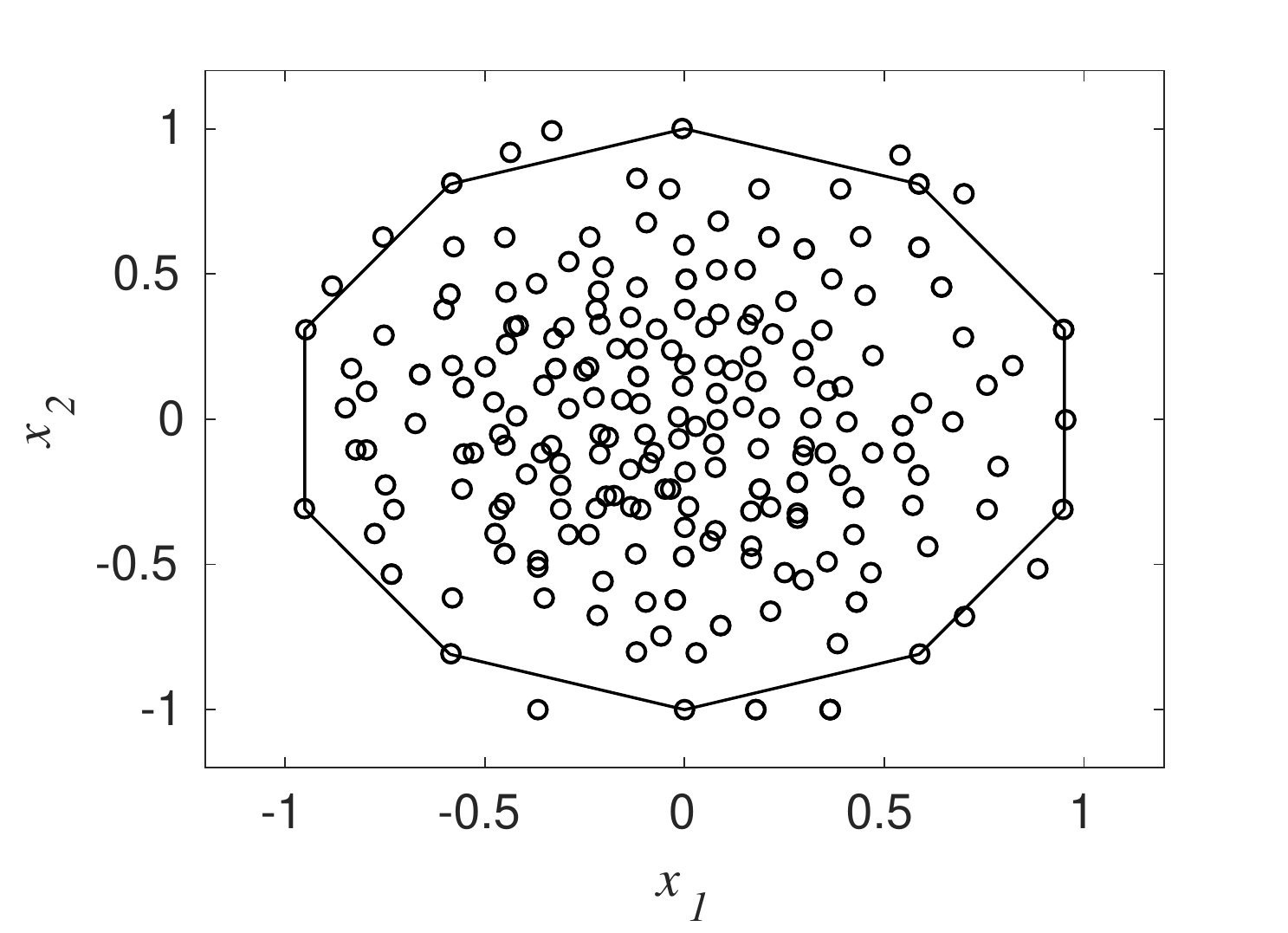}&
			\includegraphics[scale=0.257]{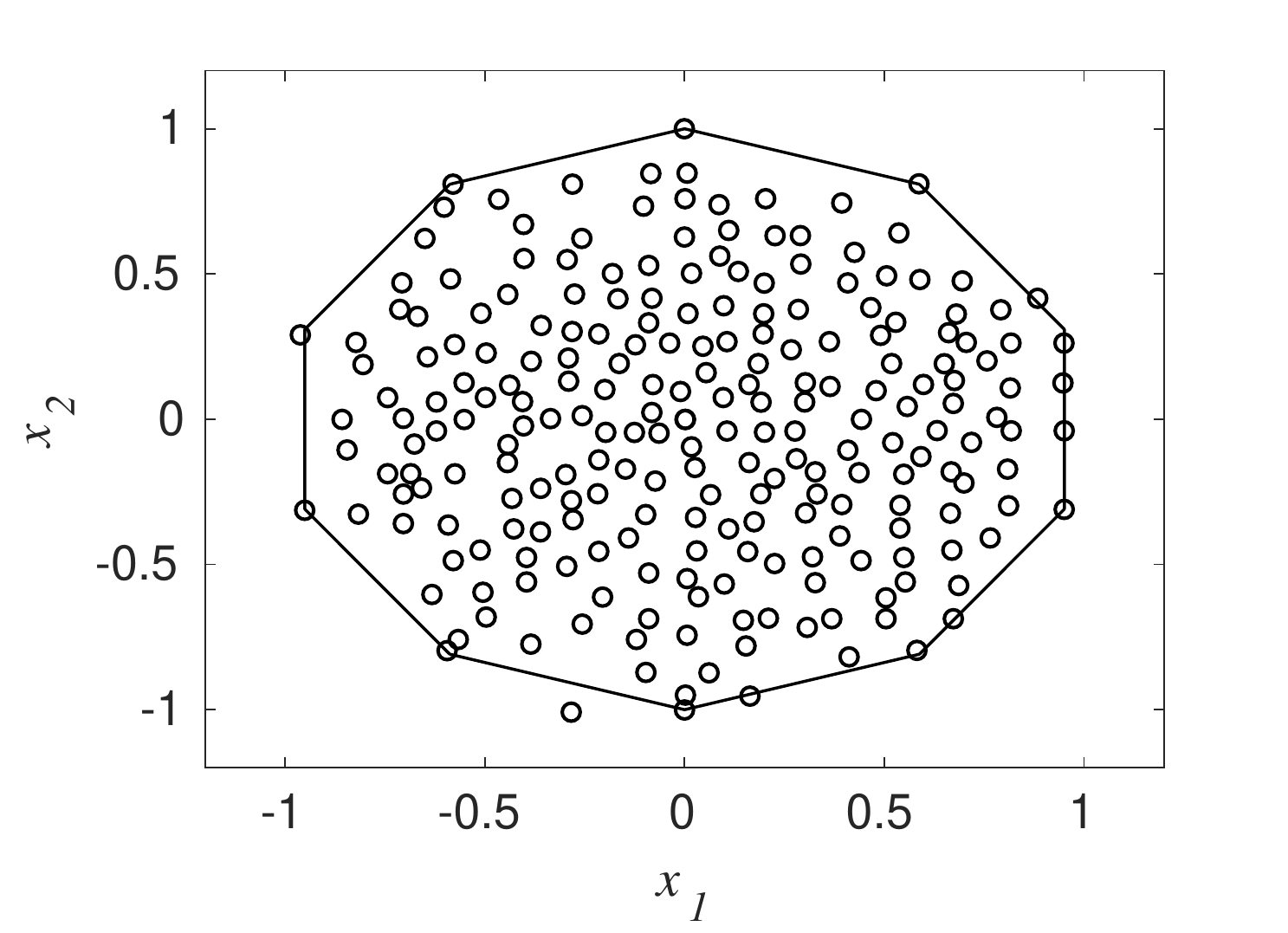}\\
			(i) SPEA2+SDE &  (j) SRA & (k) AR-MOEA  & (l) E3A  \\
		\end{tabular}
	\end{center}
	\caption{The final solution set with the median IGD obtained by the $12$ algorithms on the ten-objective MaF8.}
	\label{Fig:MaF8-10OBJ}
\end{figure}

For a visual understanding of how the solutions are distributed, Figure~\ref{Fig:MaF8-10OBJ} plots the final solution set obtained by each algorithm with the median IGD value among all the $30$ runs on the ten-objective MaF8, where the optimal region is a decagon in the decision space. From Figure~\ref{Fig:MaF8-10OBJ}, it can be observed that all the algorithms except RVEA, MOEA/DD, and MaOEA-CSS perform well regarding convergence (solutions located inside or very close to the polygon). Among these algorithms, NSGA-III, RPD-NSGA-II, 1by1EA, and SRA perform poorly regarding diversity, with their solutions crowded in some small sub-regions, leading to a large proportion of sparse regions. One interesting observation is that 1by1EA obtains a solution set that are evenly distributed but only cover the central region of the polygon. One possible explanation is that the boundary maintenance mechanism in 1by1EA may fail to find boundary solutions when the Pareto front is irregular. This can also be seen in Figure~\ref{Fig:MaF1-3OBJ} (e), where 1by1EA is unable to obtain boundary solutions even for the tri-objective MaF1 which has an inverted simplex-like Pareto front.
Although VaEA, NSGA-II/SDR, AR-MOEA, and SPEA2+SDE can provide a better balance between convergence and diversity, they have their own disadvantages. The solutions of VaEA and AR-MOEA are not uniformly distributed over the decagon, with some solutions crowded or even overlapping in some regions. NSGA-II/SDR and SPEA2+SDE struggle to find boundary solutions of the decagon. Finally, among these algorithms, E3A obtains very promising results on the ten-objective MaF8, with a set of evenly distributed solutions covering the whole decagon.

\subsubsection{Algorithm Performance on Fifteen-Objective Problems}
Tables \ref{Table:15obj_MaF_IGD} and \ref{Table:15obj_MaF_HV} show the mean and standard deviation (in parentheses) of the IGD values obtained by E3A and the peer algorithms on the fifteen-objective MaF problems. From Table \ref{Table:15obj_MaF_IGD}, it can be seen that E3A obtains the best IGD mean on five test problems, MaF$2$, MaF$4$, MaF$8$, MaF$10$, and MaF$13$. Concerning pairwise comparison, the proportions of the test problems where E3A significantly outperforms NSGA-III, RVEA, MOEA/DD, RPD-NSGA-II, 1by1EA, MaOEA-CSS, VaEA, NSGA-II/SDR, SPEA2+SDE, SRA, and AR-MOEA are $9/15$, $8/15$, $11/15$, $9/15$, $9/15$, $10/15$, $9/15$, $8/15$, $3/15$, $6/15$, and $6/15$, respectively. Conversely, the proportions of the test problems where E3A performs significantly worse than these peer algorithms are $0/15$, $1/15$, $1/15$, $2/15$, $1/15$, $2/15$, $1/15$, $3/15$, $2/15$, $2/15$, and $1/15$, respectively.

From Table \ref{Table:15obj_MaF_HV}, it can be seen that E3A obtains the best HV results on six test problems (i.e., MaF$2$, MaF$4$, MaF$8$, MaF$10$, MaF$11$, and MaF$13$). Concerning pairwise comparison, 
the proportions of the test problems where E3A significantly outperforms NSGA-III, RVEA, MOEA/DD, RPD-NSGA-II, 1by1EA, MaOEA-CSS, VaEA, NSGA-II/SDR, SPEA2+SDE, SRA, and AR-MOEA are $10/15$, $11/15$, $13/15$, $7/15$, $10/15$, $11/15$, $9/15$, $6/15$, $1/15$, $5/15$, and $6/15$, respectively. Conversely, the proportions of the test problems where E3A performs significantly worse than these peer algorithms are $0/15$, $1/15$, $1/15$, $0/15$, $2/15$, $1/15$, $1/15$, $1/15$, $1/15$, $1/15$, and $1/15$, respectively.


\begin{table}[tbp]
	\scriptsize
    \begin{center}
    \caption{Mean and standard deviation of the IGD values obtained by the $12$ algorithms on the fifteen-objective MaF problems.}
    \label{Table:15obj_MaF_IGD}
    \resizebox{\textwidth}{!}{%
    \renewcommand{\arraystretch}{1.2} 
    \begin{tabular}{@{}cccccccccccccc@{}}
 \toprule
Problem                & Obj.               & NSGA-III                      & RVEA        & MOEA/DD      & RPD-NSGA-II   & 1by1EA      & MaOEA-CSS    & VaEA        & NSGA-II/SDR   & SPEA2+SDE    & SRA         & AR-MOEA      & E3A      \\
  \midrule
\multirow{2}{*}{MaF1} & \multirow{2}{*}{15} & 3.196e-1  $\approx$ & 6.171e-1  $+$ & 5.388e-1  $+$ & 5.129e-1  $+$ & 4.846e-1  $\approx$ & 2.786e-1  $-$ & 2.767e-1  $-$ & 2.818e-1  $-$ & 3.034e-1  $\approx$  & 3.256e-1 $\approx$ & 3.228e-1  $\approx$ & 3.296e-1 \\
&           & (6.86e-3)    & (8.12e-2)    & (2.97e-2)    & (4.24e-2)    & (3.65e-2)    & (2.94e-2)    & (2.32e-3)    & (3.45e-3)    & (1.59e-2)     & (9.80e-3)    & (8.10e-3)    & (1.37e-2)   \\\hline

\multirow{2}{*}{MaF2} & \multirow{2}{*}{15} & 2.128e-1  $\approx$ & 4.640e-1  $+$ & 4.263e-1  $+$ & 2.233e-1  $+$ & 5.556e-1  $+$ & 2.402e-1  $+$ & 1.999e-1  $\approx$ & 3.492e-1  $+$ & 1.993e-1  $\approx$  & 1.988e-1  $\approx$ & 2.279e-1  $+$ & 1.930e-1 \\
&           & (1.82e-2)    & (1.27e-1)    & (4.64e-2)    & (2.93e-3)    & (2.27e-2)    & (1.15e-2)    & (2.27e-3)    & (3.58e-2)    & (1.24e-2)     & (6.25e-3)    & (9.56e-3)    & (1.13e-2)   \\\hline

\multirow{2}{*}{MaF3} & \multirow{2}{*}{15} & 2.211e+3  $+$ & 6.271e+0  $+$ & 4.388e+1  $+$ & 3.094e-1  $\approx$ & 7.445e+1  $+$ & 1.540e+1  $+$ & 6.712e+4  $+$ & 1.407e-1  $\approx$ & 1.126e-1 $\approx$  & 3.992e+4  $+$ & 5.024e+1  $+$ & 2.700e+0 \\
&           & (6.90e+3)    & (9.44e+0)    & (4.21e+1)    & (3.31e-1)    & (1.05e+2)    & (2.21e+1)    & (1.69e+5)    & (1.44e-3)    & (2.96e-3)     & (1.84e+5)    & (7.45e+1)    & (1.10e+1)   \\\hline

\multirow{2}{*}{MaF4} & \multirow{2}{*}{15} & 3.358e+3  $\approx$ & 8.806e+3  $+$ & 1.531e+4  $+$ & 4.614e+3  $+$ & 1.058e+4  $+$ & 5.451e+3  $+$ & 1.951e+3  $\approx$ & 7.050e+3  $+$ & 5.513e+3  $+$  & 4.568e+3  $+$ & 3.901e+3  $+$ & 1.476e+3 \\
&           & (4.18e+2)    & (1.93e+3)    & (2.37e+3)    & (9.87e+2)    & (9.96e+2)    & (9.96e+2)    & (9.38e+2)    & (1.26e+3)    & (7.65e+2)     & (1.07e+3)    & (4.26e+2)    & (9.55e+1)   \\\hline

\multirow{2}{*}{MaF5} & \multirow{2}{*}{15} & 2.372e+3  $+$ & 2.960e+3  $+$ & 7.302e+3  $+$ & 1.658e+3  $\approx$ & 6.019e+3  $+$ & 7.157e+3  $+$ & 1.385e+3 $\approx$ & 7.304e+3  $+$ & 2.131e+3  $\approx$  & 1.333e+3  $\approx$ & 3.607e+3  $+$ & 1.413e+3 \\
&           & (1.68e+2)    & (2.96e+2)    & (4.73e+1)    & (1.28e+2)    & (7.16e+1)    & (1.11e+2)    & (7.86e+1)    & (1.23e+2)    & (1.44e+2)     & (2.28e+2)    & (1.59e+2)    & (9.01e+1)   \\\hline

\multirow{2}{*}{MaF6} & \multirow{2}{*}{15} & 8.378e-1  $\approx$ & 4.023e-1  $-$ & 1.192e-1  $-$ & 3.427e-1  $-$ & 1.838e-3  $-$ & 3.294e-1  $-$ & 1.269e+0 $\approx$ & 2.378e-2  $-$ & 1.413e+0 $\approx$  & 4.589e+0  $\approx$ & 4.108e-1  $-$ & 1.342e+0 \\
&           & (3.71e-1)    & (2.74e-1)    & (1.45e-2)    & (1.16e-1)    & (3.97e-5)    & (1.69e-1)    & (4.97e-1)    & (7.15e-2)    & (8.48e-1)     & (1.61e+0)    & (1.43e-1)    & (5.41e-1)   \\\hline

\multirow{2}{*}{MaF7} & \multirow{2}{*}{15} & 5.992e+0  $+$ & 2.521e+0 $\approx$ & 3.430e+0  $\approx$ & 7.150e+0  $+$ & 3.074e+0 $\approx$ & 5.057e+0  $+$ & 2.493e+0 $\approx$ & 4.399e+0  $+$ & 1.452e+0  $-$  & 1.498e+0  $-$ & 3.361e+0  $\approx$ & 2.975e+0 \\
&           & (1.42e+0)    & (4.21e-1)    & (3.55e-2)    & (6.91e-1)    & (3.52e-1)    & (5.22e-1)    & (1.99e-1)    & (9.35e-1)    & (3.26e-2)     & (1.57e-2)    & (8.03e-1)    & (9.71e-1)   \\\hline

\multirow{2}{*}{MaF8} & \multirow{2}{*}{15} & 3.924e-1  $+$ & 1.268e+0  $+$ & 1.266e+0  $+$ & 7.496e-1  $+$ & 4.472e-1  $+$ & 1.667e-1  $\approx$ & 1.579e-1  $\approx$ & 2.018e-1  $+$ & 1.560e-1  $\approx$ & 9.902e-1  $+$ & 1.735e-1  $+$ & 1.378e-1 \\
&           & (6.82e-2)    & (1.88e-1)    & (9.66e-2)    & (1.31e-1)    & (1.03e-1)    & (8.05e-3)    & (3.12e-3)    & (2.04e-2)    & (7.37e-3)     & (2.17e+0)    & (5.57e-3)    & (1.99e-3)   \\\hline

\multirow{2}{*}{MaF9} & \multirow{2}{*}{15} & 2.320e+0  $+$ & 1.690e+0  $+$ & 9.510e-1  $+$ & 8.189e-1  $+$ & 2.727e-1  $\approx$ & 2.108e-1  $+$ & 1.749e-1 $\approx$ & 1.910e-1  $\approx$ & 1.316e-1  $\approx$  & 1.386e-1  $\approx$ & 1.558e-1 $\approx$ & 1.583e-1 \\
&           & (4.25e+0)    & (1.91e+0)    & (1.52e-2)    & (1.19e-1)    & (2.29e-1)    & (7.80e-3)    & (5.26e-2)    & (6.42e-3)    & (9.09e-4)     & (2.96e-2)    & (7.57e-3)    & (3.51e-2)   \\\hline

\multirow{2}{*}{MaF10} & \multirow{2}{*}{15} & 2.197e+0  $+$ & 1.853e+0  $\approx$ & 2.302e+0  $+$ & 1.588e+0  $\approx$ & 2.319e+0  $+$ & 2.090e+0  $+$ & 3.358e+0  $+$ & 2.418e+0  $+$ & 1.745e+0  $\approx$  & 2.904e+0  $+$ & 2.085e+0  $+$ & 1.570e+0 \\
&           & (9.53e-2)    & (8.24e-2)    & (1.21e-1)    & (4.24e-2)    & (1.29e-1)    & (7.61e-2)    & (1.92e-1)    & (5.95e-2)    & (4.12e-2)     & (4.28e-1)    & (7.12e-2)    & (1.69e-1)   \\\hline

\multirow{2}{*}{MaF11} & \multirow{2}{*}{15} & 1.566e+0 $\approx$ & 1.711e+0  $\approx$ & 1.913e+0  $+$ & 1.495e+0  $\approx$ & 2.276e+0  $+$ & 2.370e+0  $+$ & 1.665e+0  $\approx$ & 2.341e+0  $+$ & 1.770e+0  $+$  & 1.754e+0  $+$ & 1.549e+0 $\approx$ & 1.562e+0 \\
&           & (8.00e-2)    & (1.28e-1)    & (5.89e-2)    & (4.45e-2)    & (6.50e-2)    & (6.60e-2)    & (5.35e-2)    & (9.17e-2)    & (4.48e-2)     & (6.30e-2)    & (4.37e-2)    & (9.50e-2)   \\\hline

\multirow{2}{*}{MaF12} & \multirow{2}{*}{15} & 8.034e+0  $+$ & 7.665e+0 $\approx$ & 8.765e+0  $+$ & 8.410e+0  $+$ & 9.742e+0  $+$ & 9.817e+0  $+$ & 7.026e+0  $\approx$ & 7.657e+0 $\approx$ & 8.199e+0  $+$  & 7.997e+0  $\approx$ & 7.991e+0  $\approx$ & 7.624e+0 \\
&           & (1.45e-1)    & (2.51e-1)    & (3.02e-1)    & (6.20e-2)    & (3.29e-1)    & (2.44e-1)    & (7.60e-2)    & (1.70e-1)    & (4.01e-1)     & (2.34e-1)    & (2.10e-1)    & (1.90e-1)   \\\hline

\multirow{2}{*}{MaF13} & \multirow{2}{*}{15} & 2.785e-1  $+$ & 1.198e+0  $+$ & 3.427e-1  $+$ & 6.548e-1  $+$ & 3.353e-1  $+$ & 1.978e-1  $+$ & 1.593e-1  $+$ & 1.797e-1  $+$ & 1.224e-1  $\approx$  & 1.325e-1  $\approx$ & 1.490e-1  $\approx$ & 1.076e-1 \\
&           & (5.05e-2)    & (4.44e-1)    & (3.03e-2)    & (4.34e-2)    & (6.67e-2)    & (1.76e-2)    & (2.22e-2)    & (1.23e-2)    & (1.53e-2)     & (1.76e-2)    & (9.48e-3)    & (1.06e-2)   \\\hline

\multirow{2}{*}{MaF14} & \multirow{2}{*}{15} & 5.701e+0 $\approx$ & 3.577e+0 $\approx$ & 2.542e+0 $\approx$ & 1.622e+0  $-$ & 4.412e+0  $\approx$ & 4.153e+0  $\approx$ & 1.454e+1  $+$ & 1.539e+0  $-$ & 2.683e+0 $\approx$  & 5.558e+0  $+$ & 1.848e+0  $\approx$ & 2.961e+0 \\
&           & (4.99e+0)    & (1.75e+0)    & (9.36e-1)    & (4.41e-1)    & (1.99e+0)    & (1.23e+0)    & (5.92e+0)    & (2.99e-1)    & (9.00e-1)     & (1.64e+0)    & (6.14e-1)    & (1.29e+0)   \\\hline

\multirow{2}{*}{MaF15} & \multirow{2}{*}{15} & 1.530e+1  $+$ & 1.391e+0  $\approx$ & 3.401e+0  $+$ & 4.488e+0  $\approx$ & 1.764e+0  $\approx$ & 1.520e+0 $\approx$ & 4.231e+0  $+$ & 1.347e+0  $\approx$ & 1.168e+0  $-$  & 1.262e+0  $-$ & 1.956e+0  $\approx$ & 2.225e+0 \\
&           & (4.44e+0)    & (6.35e-2)    & (9.21e-1)    & (5.85e-1)    & (1.12e-1)    & (9.81e-2)    & (6.98e-1)    & (4.00e-2)    & (5.47e-2)     & (4.51e-2)    & (2.07e-1)    & (1.78e+0)
\\\hline

\multicolumn{2}{c}{$+$/$\approx$/$-$} & 9/6/0  & 8/6/1  & 11/3/1    & 9/4/2    & 9/5/1     & 10/3/2   & 9/5/1   & 8/4/3     & 3/10/2    & 6/7/2  & 6/8/1       &           
\\\bottomrule 
    \end{tabular}%
    }
    \end{center}
    \vspace{6pt}
\footnotesize `$+$', `$\approx$', and `$-$' indicate that the result of E3A is significantly better than, equivalent to, and significantly worse than that of the peer algorithm at a $0.05$ level by the statistical tests (i.e., Friedman test and posthoc Nemenyi test), respectively.
\end{table}


\begin{table}[tbp]
	\scriptsize
    \begin{center}
    \caption{Mean and standard deviation of the HV values obtained by the $12$ algorithms on the fifteen-objective MaF problems.}
    \label{Table:15obj_MaF_HV}
    \resizebox{\textwidth}{!}{%
    \renewcommand{\arraystretch}{1.2} 
    \begin{tabular}{@{}cccccccccccccc@{}}
 \toprule
Problem                & Obj.               & NSGA-III                      & RVEA        & MOEA/DD      & RPD-NSGA-II   & 1by1EA      & MaOEA-CSS    & VaEA        & NSGA-II/SDR   & SPEA2+SDE    & SRA         & AR-MOEA      & E3A      \\ \midrule
\multirow{2}{*}{MaF1} & \multirow{2}{*}{15} & 7.271e-03 $+$       & 3.884e-03 $+$       & 3.638e-03 $+$       & 8.698e-03 $+$       & 1.847e-02 $-$       & 1.821e-02 $\approx$ & 1.599e-02 $\approx$ & 1.779e-02 $\approx$ & 1.740e-02 $\approx$ & 1.350e-02 $\approx$ & 1.458e-02 $\approx$ & 1.677e-02  \\
        &      & (4.86e-04)          & (1.08e-03)          & (2.47e-03)          & (6.51e-04)          & (4.82e-04)          & (1.47e-04)          & (3.25e-04)          & (2.74e-04)          & (1.89e-04)          & (4.73e-04)          & (3.44e-04)          & (1.72e-04) \\ \hline
\multirow{2}{*}{MaF2} & \multirow{2}{*}{15}  & 7.673e-01 $+$       & 7.568e-01 $+$       & 6.236e-01 $+$       & 7.513e-01 $+$       & 7.916e-01 $+$       & 7.773e-01 $+$       & 8.119e-01 $+$       & 8.605e-01 $\approx$ & 8.720e-01 $\approx$ & 8.307e-01 $\approx$ & 7.883e-01 $+$       & 8.734e-01  \\
        &      & (1.35e-02)          & (9.55e-03)          & (2.26e-02)          & (9.16e-03)          & (1.36e-02)          & (1.39e-02)          & (1.19e-02)          & (1.10e-02)          & (7.73e-03)          & (1.00e-02)          & (8.11e-03)          & (6.11e-03) \\ \hline
\multirow{2}{*}{MaF3} & \multirow{2}{*}{15} & 3.975e-01 $+$       & 0.000e+00 $+$       & 5.266e-02 $+$       & 1.049e+00 $+$       & 1.132e+00 $+$       & 5.362e-01 $+$       & 4.511e-02 $+$       & 1.562e+00 $\approx$ & 1.471e+00 $\approx$ & 1.437e+00 $\approx$ & 1.357e+00 $\approx$ & 1.419e+00  \\
        &      & (6.08e-01)          & (0.00e+00)          & (2.84e-01)          & (6.89e-01)          & (6.00e-01)          & (6.69e-01)          & (2.43e-01)          & (2.18e-02)          & (4.01e-01)          & (4.79e-01)          & (5.41e-01)          & (4.97e-01) \\ \hline
\multirow{2}{*}{MaF4} & \multirow{2}{*}{15} & 2.925e-02 $+$       & 1.631e-02 $+$       & 4.889e-02 $+$       & 1.001e-01 $\approx$ & 3.954e-02 $+$       & 8.523e-02 $+$       & 6.212e-02 $+$       & 1.496e-01 $\approx$ & 9.529e-02 $\approx$ & 7.708e-02 $+$       & 8.642e-02 $+$       & 1.514e-01  \\
        &      & (3.49e-02)          & (1.29e-02)          & (2.32e-02)          & (3.33e-02)          & (1.52e-02)          & (4.13e-02)          & (5.14e-02)          & (6.10e-02)          & (3.73e-02)          & (4.33e-02)          & (3.53e-02)          & (4.48e-02) \\ \hline
\multirow{2}{*}{MaF5} & \multirow{2}{*}{15} & 1.250e+00 $\approx$ & 1.225e+00 $\approx$ & 9.926e-01 $+$       & 1.265e+00 $\approx$ & 1.016e+00 $+$       & 9.709e-01 $+$       & 1.240e+00 $\approx$ & 2.424e-01 $+$       & 1.217e+00 $\approx$ & 1.184e+00 $\approx$ & 1.266e+00 $\approx$ & 1.248e+00  \\
        &      & (7.47e-02)          & (9.35e-02)          & (1.16e-01)          & (9.49e-03)          & (4.01e-02)          & (2.38e-02)          & (5.38e-03)          & (1.32e-01)          & (4.28e-02)          & (3.61e-02)          & (3.76e-02)          & (6.30e-02) \\ \hline
\multirow{2}{*}{MaF6} & \multirow{2}{*}{15} & 1.962e-01 $+$       & 1.819e-01 $+$       & 1.838e-01 $+$       & 1.805e-01 $+$       & 2.083e-01 $\approx$ & 1.305e-01 $+$       & 2.088e-01 $-$       & 1.962e-01 $+$       & 2.071e-01 $\approx$ & 2.052e-01 $\approx$ & 2.084e-01 $\approx$ & 2.074e-01  \\
        &      & (2.73e-03)          & (9.88e-03)          & (2.08e-03)          & (1.06e-02)          & (1.03e-04)          & (3.16e-02)          & (6.15e-05)          & (7.44e-03)          & (4.58e-04)          & (1.22e-03)          & (5.99e-05)          & (5.85e-05) \\ \hline
\multirow{2}{*}{MaF7} & \multirow{2}{*}{15} & 4.893e-01 $+$       & 4.030e-01 $+$       & 1.464e-01 $+$       & 5.145e-01 $\approx$ & 2.933e-01 $+$       & 2.851e-01 $+$       & 4.830e-01 $+$       & 5.195e-01 $\approx$ & 5.445e-01 $\approx$ & 5.160e-01 $\approx$ & 4.829e-01 $+$       & 5.255e-01  \\
        &      & (1.36e-02)          & (2.28e-02)          & (6.59e-07)          & (7.36e-03)          & (5.54e-02)          & (3.37e-02)          & (1.08e-02)          & (7.80e-03)          & (9.84e-03)          & (6.73e-03)          & (6.27e-03)          & (9.51e-02) \\ \hline
\multirow{2}{*}{MaF8} & \multirow{2}{*}{15} & 1.605e-01 $+$       & 1.273e-01 $+$       & 1.306e-01 $+$       & 1.417e-01 $+$       & 1.403e-01 $+$       & 2.010e-01 $\approx$ & 1.993e-01 $\approx$ & 1.962e-01 $+$       & 2.014e-01 $\approx$ & 1.496e-01 $+$       & 1.950e-01 $+$       & 2.018e-01  \\
        &      & (1.06e-02)          & (7.69e-03)          & (1.63e-02)          & (9.28e-03)          & (1.10e-02)          & (1.78e-03)          & (2.84e-03)          & (2.02e-03)          & (3.40e-03)          & (5.89e-02)          & (6.07e-03)          & (2.47e-03) \\ \hline
\multirow{2}{*}{MaF9} & \multirow{2}{*}{15} & 2.824e-01 $+$       & 3.185e-01 $+$       & 3.816e-01 $+$       & 3.434e-01 $+$       & 3.649e-01 $+$       & 4.101e-01 $+$       & 2.994e-01 $+$       & 4.445e-01 $\approx$ & 4.998e-01 $\approx$ & 4.942e-01 $\approx$ & 4.833e-01 $\approx$ & 4.909e-01  \\
        &      & (8.28e-02)          & (2.47e-02)          & (3.47e-02)          & (3.37e-02)          & (2.98e-02)          & (1.20e-02)          & (9.67e-02)          & (9.80e-03)          & (8.83e-03)          & (1.41e-02)          & (1.45e-02)          & (1.34e-02) \\ \hline
\multirow{2}{*}{MaF10} & \multirow{2}{*}{15}  & 1.062e+00 $+$       & 1.164e+00 $+$       & 7.850e-01 $+$       & 1.348e+00 $\approx$ & 1.127e+00 $+$       & 1.182e+00 $+$       & 8.463e-01 $+$       & 1.269e+00 $+$       & 1.513e+00 $\approx$ & 8.351e-01 $+$       & 1.154e+00 $+$       & 1.546e+00  \\
        &      & (6.80e-02)          & (7.88e-02)          & (6.36e-02)          & (6.67e-02)          & (7.23e-02)          & (5.49e-02)          & (6.38e-02)          & (8.40e-02)          & (4.73e-02)          & (7.39e-02)          & (5.56e-02)          & (5.23e-02) \\ \hline
\multirow{2}{*}{MaF11} & \multirow{2}{*}{15} & 1.591e+00 $\approx$ & 1.578e+00 $+$       & 1.551e+00 $+$       & 1.593e+00 $\approx$ & 1.564e+00 $+$       & 1.514e+00 $+$       & 1.585e+00 $+$       & 1.551e+00 $+$       & 1.575e+00 $+$       & 1.558e+00 $+$       & 1.597e+00 $\approx$ & 1.606e+00  \\
        &      & (3.30e-03)          & (9.67e-03)          & (7.34e-03)          & (3.60e-03)          & (8.61e-03)          & (1.92e-02)          & (3.98e-03)          & (1.20e-02)          & (7.34e-03)          & (7.90e-03)          & (3.15e-03)          & (1.59e-03) \\ \hline
\multirow{2}{*}{MaF12} & \multirow{2}{*}{15} & 1.149e+00 $\approx$ & 1.152e+00 $\approx$ & 1.064e+00 $+$       & 1.137e+00 $\approx$ & 9.670e-01 $+$       & 9.259e-01 $+$       & 1.079e+00 $+$       & 1.182e+00 $-$       & 1.157e+00 $\approx$ & 1.114e+00 $+$       & 1.112e+00 $\approx$ & 1.145e+00  \\
        &      & (2.08e-02)          & (1.83e-02)          & (2.60e-02)          & (4.36e-02)          & (2.67e-02)          & (3.20e-02)          & (5.41e-02)          & (7.57e-03)          & (9.95e-03)          & (1.37e-02)          & (2.77e-02)          & (4.19e-02) \\ \hline
\multirow{2}{*}{MaF13} & \multirow{2}{*}{15} & 3.120e-01 $+$       & 2.787e-01 $+$       & 3.331e-01 $+$       & 1.304e-01 $+$       & 4.816e-01 $\approx$ & 4.455e-01 $+$       & 3.995e-01 $+$       & 4.072e-01 $+$       & 4.872e-01 $\approx$ & 4.802e-01 $\approx$ & 4.491e-01 $+$       & 5.035e-01  \\
        &      & (5.28e-02)          & (7.48e-02)          & (8.86e-02)          & (4.48e-02)          & (1.56e-02)          & (1.40e-02)          & (2.61e-02)          & (2.01e-02)          & (1.47e-02)          & (1.55e-02)          & (1.50e-02)          & (1.46e-02) \\ \hline
\multirow{2}{*}{MaF14} & \multirow{2}{*}{15} & 0.000e+00 $\approx$ & 4.348e-02 $\approx$ & 1.747e-02 $\approx$ & 0.000e+00 $\approx$ & 1.518e-01 $\approx$ & 8.000e-02 $\approx$ & 0.000e+00 $\approx$ & 1.585e-01 $\approx$ & 1.183e-02 $\approx$ & 4.464e-04 $\approx$ & 3.441e-02 $\approx$ & 4.395e-02  \\
        &      & (0.00e+00)          & (6.65e-02)          & (5.06e-02)          & (0.00e+00)          & (1.34e-01)          & (9.18e-02)          & (0.00e+00)          & (1.21e-01)          & (3.81e-02)          & (2.30e-03)          & (6.46e-02)          & (7.26e-02) \\ \hline
\multirow{2}{*}{MaF15} & \multirow{2}{*}{15} & 0.000e+00 $\approx$ & 1.454e-02 $-$       & 1.127e-02 $-$       & 0.000e+00 $\approx$ & 3.153e-02 $-$       & 4.106e-02 $-$       & 0.000e+00 $\approx$ & 1.229e-03 $\approx$ & 5.409e-02 $-$       & 2.745e-02 $-$       & 1.155e-02 $-$       & 8.201e-05  \\
        &      & (0.00e+00)          & (6.99e-03)          & (7.22e-03)          & (0.00e+00)          & (1.10e-02)          & (9.65e-03)          & (0.00e+00)          & (8.13e-04)          & (1.58e-02)          & (1.23e-02)          & (3.87e-03)          & (1.80e-04) \\ \hline
\multicolumn{2}{c}{$+$/$\approx$/$-$}                   & 10/5/0  & 11/3/1  & 13/1/1 &  7/8/0   &   10/3/2   & 11/3/1  & 9/5/1  &  6/8/1   &  1/13/1  & 5/9/1  &   6/8/1    &    
\\\bottomrule 
    \end{tabular}%
    }
    \end{center}
    \vspace{6pt}
\footnotesize `$+$', `$\approx$', and `$-$' indicate that the result of E3A is significantly better than, equivalent to, and significantly worse than that of the peer algorithm at a $0.05$ level by the statistical tests (i.e., Friedman test and posthoc Nemenyi test), respectively.
\end{table}

\subsubsection{Overall Performance}
Tables \ref{tab:summary_data_IGD} and \ref{tab:summary_data_HV} summarize the comparative results in terms of IGD and HV of E3A against the $11$ peer algorithms on all $60$ MaF problem instances, respectively.
Symbols `$+$', `$\approx$', and `$-$' indicate the number of test instances where E3A performs significantly better than, equivalent to, and significantly worse than the peer algorithm at a $0.05$ level by the statistical tests (i.e., Friedman test and posthoc Nemenyi test), respectively.

\begin{table}[tbp]
	\scriptsize
    \begin{center}
\caption{Summary of the comparative results in terms of IGD between E3A and the other $11$ state-of-the-art algorithms in all $60$ MaF problem instances. }
    \label{tab:summary_data_IGD}
    \resizebox{\textwidth}{!}{%
    \renewcommand{\arraystretch}{1.2} 
    \begin{tabular}{@{}ccccccccccccc@{}}
    \toprule
                             & Obj.  & NSGA-III & RVEA     & MOEA/DD & RPD-NSGA-II & 1by1EA & MaOEA-CSS & VaEA    & NSGA-II/SDR & SPEA2+SDE & SRA     & AR-MOEA  \\   \midrule
\multirow{5}{*}{E3A ($+$/$\approx$/$-$)} & 3     & 7/5/3   & 12/1/2  & 8/6/1    & 11/2/2   & 8/5/2     & 7/7/1    & 5/6/4    & 9/6/0     & 4/10/1   & 9/6/0   & 4/8/3  \\
                                         & 5     & 10/4/1  & 11/2/2  & 12/2/1   & 9/4/2    & 9/3/3     & 10/2/3   & 6/8/1    & 5/10/0    & 3/11/1   & 7/7/1   & 2/11/2   \\
                                         & 10    & 6/9/0   & 9/5/1   & 11/3/1   & 8/5/2    & 12/2/1    & 7/6/2    & 5/5/5    & 9/5/1     & 1/11/3   & 4/8/3   & 4/9/2   \\
                                         & 15    & 9/6/0   & 8/6/1   & 11/3/1   & 9/4/2    & 9/5/1     & 10/3/2   & 9/5/1    & 8/4/3     & 3/10/2   & 6/7/2   & 6/8/1    \\
                                         & Total & 32/24/4 & 40/14/6 & 42/14/4  & 37/15/8  & 38/15/7   & 34/18/8  & 25/24/11 & 31/25/4   & 11/42/7  & 26/28/6 & 16/36/8
    \\ \bottomrule
    \end{tabular}%
    }
    \end{center}
        \vspace{6pt}
\footnotesize `$+$', `$\approx$', and `$-$' indicate that the result of E3A is significantly better than, equivalent to, and significantly worse than that of the peer algorithm at a $0.05$ level by the statistical tests (i.e., Friedman test and posthoc Nemenyi test), respectively.
\end{table}

\begin{table}[tbp]
	\scriptsize
    \begin{center}
\caption{Summary of the comparative results in terms of HV between E3A and the other $11$ state-of-the-art algorithms in all $60$ MaF problem instances. }
    \label{tab:summary_data_HV}
    \resizebox{\textwidth}{!}{%
    \renewcommand{\arraystretch}{1.2} 
    \begin{tabular}{@{}ccccccccccccc@{}}
    \toprule
                             & Obj.  & NSGA-III & RVEA     & MOEA/DD & RPD-NSGA-II & 1by1EA & MaOEA-CSS & VaEA    & NSGA-II/SDR & SPEA2+SDE & SRA     & AR-MOEA  \\   \midrule
\multirow{5}{*}{E3A ($+$/$\approx$/$-$)} & 3     &  11/4/0  & 13/2/0   & 10/4/1  &  11/4/0   &  12/2/1    &  11/3/1 & 9/6/0   & 9/6/0    & 1/13/1   & 11/2/2  &   5/10/0 \\
                                         & 5     &  10/5/0  & 11/3/1   & 13/1/1  &  7/8/0    &  10/3/2    & 11/3/1  & 9/5/1   & 6/8/1    &  1/13/1  &  5/9/1  &   6/8/1 \\
                                         & 10    &  6/7/2   & 8/5/2    & 11/3/1  &  6/5/4    &   7/7/1    & 8/6/1   & 4/11/0  &  6/6/3   &  2/12/1  & 6/9/0   &   5/8/2  \\
                                         & 15    &  10/5/0  & 11/3/1   & 13/1/1  &  7/8/0    &  10/3/2    & 11/3/1  & 9/5/1   &  6/8/1   &  1/13/1  & 5/9/1   &   6/8/1   \\
                                         & Total &  37/21/2 & 43/13/4  & 47/9/4  &  31/25/4  &  39/15/6   & 41/15/4 & 31/27/2 & 27/28/5  & 5/51/4   & 27/29/4 &   22/34/4
    \\ \bottomrule
    \end{tabular}%
    }
    \end{center}
        \vspace{6pt}
\footnotesize `$+$', `$\approx$', and `$-$' indicate that the result of E3A is significantly better than, equivalent to, and significantly worse than that of the peer algorithm at a $0.05$ level by the statistical tests (i.e., Friedman test and posthoc Nemenyi test), respectively.
\end{table}

As shown in Table \ref{tab:summary_data_IGD}, in terms of IGD, E3A significantly outperforms the peer algorithms (NSGA-III, RVEA, MOEA/DD, RPD-NSGA-II, 1by1EA, MaOEA-CSS, VaEA, NSGA-II/SDR, SPEA2+SDE, SRA, and AR-MOEA) on $32$, $40$, $42$, $37$, $38$, $34$, $25$, $31$, $11$, $26$, and $16$ out of all $60$ test instances, respectively. Conversely, E3A significantly performs worse than the $11$ algorithms on $4$, $6$, $4$, $8$, $7$, $8$, $11$, $4$, $7$, $6$, and $8$ out of all $60$ test instances, respectively.

As shown in Table \ref{tab:summary_data_HV}, in terms of HV, E3A significantly outperforms the peer algorithms (NSGA-III, RVEA, MOEA/DD, RPD-NSGA-II, 1by1EA, MaOEA-CSS, VaEA, NSGA-II/SDR, SPEA2+SDE, SRA, and AR-MOEA) on $37$, $43$, $47$, $31$, $39$, $41$, $31$, $27$, $5$, $27$, and $22$ out of all $60$ test instances, respectively. Conversely, E3A significantly performs worse than the $11$ algorithms on $2$, $4$, $4$, $4$, $6$, $4$, $2$, $5$, $4$, $4$, and $4$ out of all $60$ test instances, respectively. The results demonstrate the effectiveness of the proposed algorithm.

Next, we investigate whether E3A performs well in terms of both effectiveness and efficiency by comparing it with all the peer algorithms based on the average performance score \cite{Bader2011hype} and the average computational time, respectively. For a specific problem instance, the performance score of an algorithm is the number of the peer algorithms that perform significantly better than the algorithm on the problem instance according to the statistical results. The smaller the score, the better the performance of the algorithm on the problem instance.

\begin{figure}[tbp]
	\begin{center}
		\footnotesize
		\includegraphics[scale=0.55]{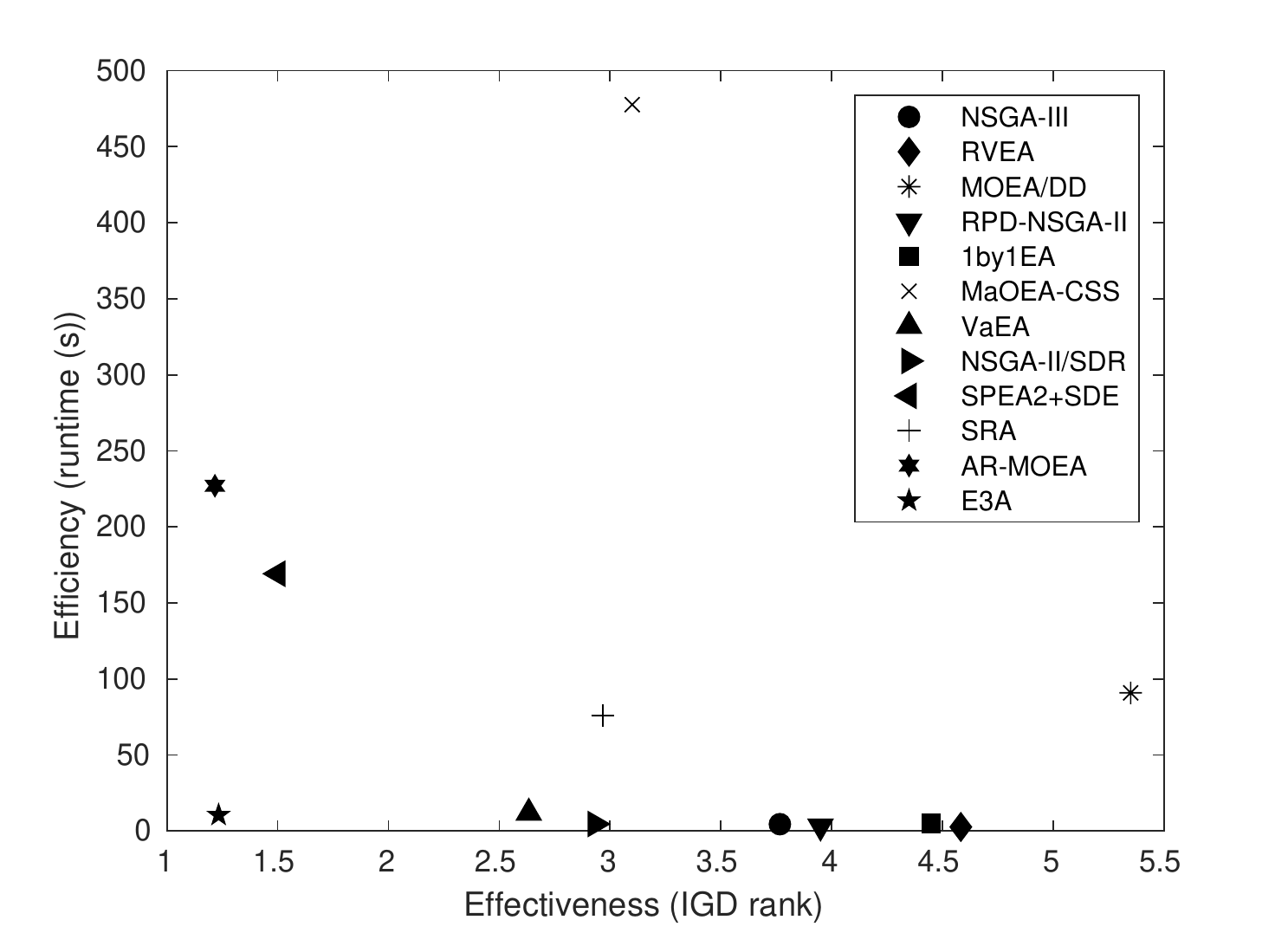}\\
	\end{center}
	\caption{Plots of effectiveness and efficiency of the $12$ algorithms on all $60$ problem instances. The effectiveness is the average performance scores of these algorithms on all $60$ problem instances in terms of IGD, and the efficiency is the average computational time (in seconds) of these algorithms on all $60$ problem instances. The smaller the score, the better the overall performance of the algorithm in terms of IGD.}
	\label{fig:Effectiveness_Efficiency_PF}
\end{figure}

Figure~\ref{fig:Effectiveness_Efficiency_PF} shows the average performance scores in terms of IGD and the average computational time (in seconds) of the $12$ algorithms on all $60$ problem instances. As can be seen, the proposed E3A strikes in general the best balance between effectiveness and efficiency out of the $12$ algorithms. E3A and AR-MOEA obtain almost the same best average IGD score ($1.23$ and $1.22$, respectively), and the computational cost of E3A is significantly lower than other well-performed algorithms such as SPEA2+SDE and AR-MOEA.

Note that the main goal of MaOEAs is to assist the decision maker to choose the preferred solution(s). Since the Pareto front of a real-world MaOP is often unknown, a set of well-distributed and well-converged solutions that approximate the Pareto front can be used to investigate the characteristics of the problem (e.g., the shape of the approximate Pareto front may be convex or degenerate). Based on the approximate Pareto front, the decision maker can determine preferences. For example, the decision maker may only be interested in solutions that the objective value is within a specific range in certain objectives. In this case, we can incorporate the decision maker's preferences into the evaluation of the obtained solution set to select those preferred solutions.

\section{Conclusion} \label{sec:E3A_Conclusion}
Achieving both high effectiveness and efficiency can be a challenging task in many-objective optimization, particularly in a real-world application, where the problem's Pareto front is typically irregular. 
In this paper, we have proposed a many-objective evolutionary algorithm (called E3A) to tackle problems with various Pareto fronts. We have compared our algorithm with $11$ state-of-the-art algorithms on $60$ problem instances with up to $15$ objectives. 
The results have shown that our algorithm achieves a better balance between effectiveness and efficiency than its peers.
Lastly, it is worth mentioning that like most many-objective optimization algorithms, our algorithm focuses on the environmental selection, one key component in an evolutionary algorithm, leaving mating selection flexible to be implemented. In the future, we would like to explore the mating selection component in many-objective optimization further, particularly on how to select solutions based on their shifted position for the crossover operation. Moreover, we would like to apply our approach to the real-world MaOPs, such as the many-objective optimal software product selection problem in engineering \cite{Hierons2016sip} and the many-objective task scheduling problem in cloud computing \cite{Zhang2022efficient}. 

\bibliographystyle{IEEEtran}
\bibliography{IEEEabrv,bibfile}

\end{document}